\documentclass{article} 
\usepackage{iclr2025,times}

\usepackage[backref]{hyperref} 
\hypersetup{
    colorlinks,
    linkcolor={purple},
    urlcolor={purple!66!black},
    citecolor={blue!50!black}
}
\usepackage{url}            
\usepackage{booktabs}       
\usepackage{multirow}
\usepackage{esvect}
\usepackage{amsfonts}       
\usepackage{nicefrac}       
\usepackage{microtype}      
\usepackage{xcolor}         
\definecolor{mygreen}{rgb}{0.05882,0.61569,0.34510}
\definecolor{myred}{rgb}{0.8554,0.2656,0.2148}
\definecolor{myblue}{rgb}{0.2578,0.5195,0.9531}
\usepackage{graphicx}       
\usepackage{wrapfig}        
\usepackage{amsmath,amsthm} 
\usepackage{amssymb,bbm,mathtools} 
\usepackage{titletoc}       
\usepackage[labelfont=bf]{caption}
\usepackage{xspace}         
\usepackage{titletoc}       
\usepackage[ruled]{algorithm2e} 
\def\cte{\textsc{cte}\xspace}
\def\ctetitle{\textsc{CTE}\xspace}
\def\kt{\textsc{kt}\xspace}
\def\compresspp{\textsc{compress}++\xspace}
\def\sage{\textsc{sage}\xspace}
\def\kernsage{\textsc{kernel-sage}\xspace}
\def\permsage{\textsc{permutation-sage}\xspace}

\def\shap{\textsc{shap}\xspace}
\def\kernshap{\textsc{kernel-shap}\xspace}
\def\permshap{\textsc{permutation-shap}\xspace}

\def\lime{\textsc{lime}\xspace}
\def\glime{\textsc{g-lime}\xspace}

\def\featureeffects{\textsc{feature-effects}\xspace}
\def\expectedgradients{\textsc{expected-gradients}\xspace}
\def\dtv{\mathrm{TV}\xspace}
\def\dmmd{\mathrm{MMD}\xspace}

\def\nvalid{n_{\text{valid}}\xspace}
\def\ntrain{n_{\text{train}}\xspace}
\def\iid{i.i.d.\xspace}

\newcommand{\norm}[1]{\left\lVert#1\right\rVert}
\newcommand{\normbig}[1]{\big\lVert#1\big\rVert}

\def\bbX{\mathbb{X}}
\def\bbXtilde{\widetilde{\mathbb{X}}}
\def\ntilde{\tilde{n}}

\def\mX{\mathcal{X}}
\def\mY{\mathcal{Y}}

\def\bbR{\mathbb{R}}
\def\bbE{\mathbb{E}}

\def\sbar{\bar{s}}

\def\bx{\mathbf{x}}
\def\bxs{\mathbf{x}_s}
\def\bxsb{\mathbf{x}_{\sbar}}

\def\bk{\mathbf{k}}

\def\bX{\mathbf{X}}
\def\bXs{\mathbf{X}_s}
\def\bXsb{\mathbf{X}_{\sbar}}
\def\pX{p_{\bX}}
\def\pXx{p_{\bX}(\bx)}
\def\pXsx{p_{\bXs}(\bxs)}
\def\pXsb{p_{\bX_{\sbar}}}
\def\pXsbx{p_{\bX_{\sbar}}(\bxsb)}
\def\pXsXt{p_{\bX_s | \bX_t}}
\def\pXsbXs{p_{\bXsb | \bXs = \bxs}}

\def\qXsb{q_{\bX_{\sbar}}}

\newtheorem{proposition}{Proposition}
\newtheorem{definition}{Definition}
\theoremstyle{remark}

\usepackage[frozencache,cachedir=minted-cache]{minted}
\definecolor{codebackground}{RGB}{240, 240, 235}
\definecolor{highlightcolor}{RGB}{218, 218, 213}
\newsavebox{\mintedbox}
\newenvironment{boxminted}[1]
 {%
  \VerbatimEnvironment
  \RecustomVerbatimEnvironment{Verbatim}{BVerbatim}{}%
  \begin{lrbox}{\mintedbox}
  \begin{minted}[highlightlines={#1},highlightcolor=highlightcolor]%
 }
 {%
  \end{minted}%
  \end{lrbox}%
  \noindent\colorbox{codebackground}{\makebox[0.99\textwidth][c]{\usebox{\mintedbox}}}%
 }

\title{Efficient and Accurate Explanation Estimation with Distribution Compression}

\author{Hubert Baniecki\\
  University of Warsaw \\
  \texttt{h.baniecki@uw.edu.pl} \\
  \And
  Giuseppe Casalicchio \\
  LMU Munich \\
  Munich Center for Machine Learning \\
  \And
  Bernd Bischl \\
  LMU Munich \\
  Munich Center for Machine Learning \\
  \And
  Przemyslaw Biecek \\
  University of Warsaw\\
  Warsaw University of Technology\\
}

\iclrfinalcopy 
\begin{document}

\maketitle

\vspace{-1em}
\begin{abstract}
    We discover a theoretical connection between explanation estimation and distribution compression that significantly improves the approximation of feature attributions, importance, and effects.
    While the exact computation of various machine learning explanations requires numerous model inferences and becomes impractical, 
    the computational cost of approximation increases with an ever-increasing size of data and model parameters. 
    We show that the standard \iid sampling used in a broad spectrum of algorithms for post-hoc explanation leads to an approximation error worthy of improvement. 
    To this end, we introduce \emph{compress then explain}~(\cte), a new paradigm of sample-efficient explainability. 
    It relies on distribution compression through kernel thinning to obtain a data sample that best approximates its marginal distribution. 
    \cte significantly improves the accuracy and stability of explanation estimation with negligible computational overhead. 
    It often achieves an on-par explanation approximation error 2--3$\times$ faster by using fewer samples, i.e. requiring 2--3$\times$ fewer model evaluations. 
    \cte is a simple, yet powerful, plug-in for any explanation method that now relies on \iid sampling.
\end{abstract}

\section{Introduction}

Computationally efficient estimation of post-hoc explanations is at the forefront of current research on explainable machine learning~\citep{strumbelj2010efficient,slack2021reliable,jethani2022fastshap,chen2023algorithms,donnelly2023rashomon,muschalik2024beyond}. 
The majority of the work focuses on improving efficiency with respect to the dimension of features~\citep{covert2020understanding,jethani2022fastshap,chen2023algorithms,fumagalli2023shapiq}, specific model classes like neural networks~\citep{erion2021improving,chen2024less} and decision trees~\citep{muschalik2024beyond}, or approximating the conditional feature distribution~\citep{chen2018learning,aas2021explaining,olsen2022using,olsen2024comparative}.

However, in many practical settings, a marginal feature distribution is used instead to estimate explanations, and \iid samples from the data typically form the so-called \emph{background data} samples, also known as \emph{reference points} or \emph{baselines}, which plays a crucial role in the estimation process~\citep{lundberg2017unified,scholbeck2020sampling,erion2021improving,ghalebikesabi2021locality,lundstrom2022rigorous}. 
For example, \citet{covert2020understanding} mention ``\emph{[O]ur sampling approximation for SAGE was run using draws from the marginal distribution. 
We used a fixed set of 512 background samples [...]}'' and we provide more such references in 
Appendix~\ref{app:motivation-quotes} to motivate our research question: 
\emph{Can we reliably improve on standard \iid sampling in explanation estimation?}

\begin{figure*}[t]
    \centering
    \includegraphics[width=0.88\textwidth]{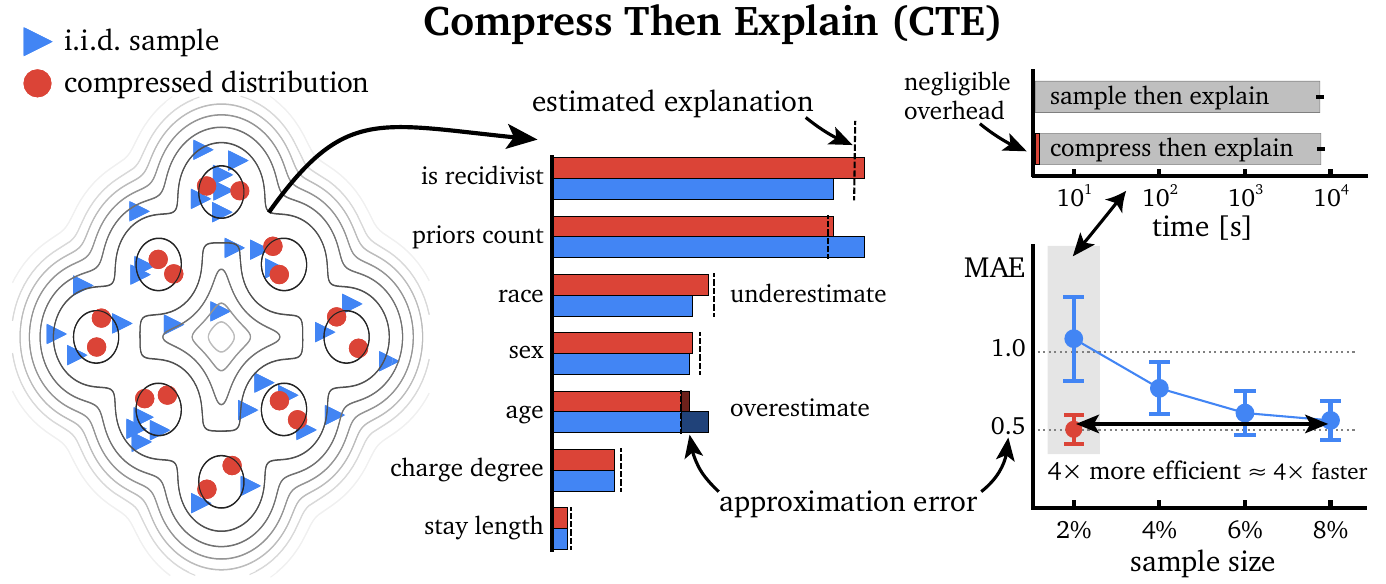}
    \caption{
    \textbf{Garbage sample in, garbage explanation out.} \emph{Sample then explain} is a conventional approach to decrease the computational cost of explanation estimation. Although fast, sampling is inefficient and prone to error, which may even lead to changes in feature importance rankings. We propose \emph{compress then explain} (\cte), a new paradigm for accurate, yet efficient, estimation of explanations based on a marginal distribution that is compressed, e.g. with kernel thinning.
    }
    \label{fig:abstract}
\end{figure*}

We make a connection to research on statistical theory, where kernel thinning \citep[\kt,][]{dwivedi2021kernel,dwivedi2022generalized} was introduced to compress a distribution more effectively than with \iid sampling. 
\kt has an efficient implementation in the \compresspp algorithm \citep{shetty2022distribution} and was applied to improve statistical kernel testing~\citep{domingo2023compress}. 
Building on this line of work, this paper aims to thoroughly quantify the error introduced by the current \emph{sample then explain} paradigm in feature marginalization, which is involved in the estimation of both local and global removal-based explanations~\citep{covert2021explaining}. 
We propose an efficient way to reduce this approximation error based on distribution compression~(Figure~\ref{fig:abstract}). 

\textbf{Contribution.} 
In summary, our work advances literature in multiple ways:
\textbf{(1)~Quantifying the error of standard \iid sampling:} We bring to attention and measure the approximation error introduced by using \iid sampling of background and foreground data in various explanation methods.
It may even lead to changes in feature importance rankings.
\textbf{(2)~Compress then explain~(\cte):} We introduce a new paradigm of sample-efficient explainability where post-hoc explanations, like feature attributions and effects, are estimated based on a marginal distribution compressed more efficiently than with \iid sampling. \cte is theoretically justified as we discover a connection between explanation estimation and distribution compression.
\textbf{(3)~Kernel thinning for (explainable) machine learning:} We show empirically that \kt outperforms \iid sampling in compressing the distribution of popular datasets used in research on explainable machine learning. 
In fact, this is the first work to evaluate distribution compression via \kt on datasets for supervised learning, which itself is valuable.  
\textbf{(4)~Decreasing the computational cost of explanation estimation:} We benchmark \emph{compress then explain} (\cte) with popular explanation methods and show it results in more accurate explanations \underline{of smaller variance}. 
\cte often achieves on-par error using 2--3$\times$ fewer samples, i.e. requiring 2--3$\times$ fewer model inferences. \cte is a simple, yet powerful, plug-in for a broad class of methods that sample from a dataset, e.g. removal-based and global explanations.

\textbf{Related work.} 
Our work is the first to empirically evaluate \kt on datasets for supervised learning, and one of the first to reliably improve on \iid sampling for multiple post-hoc explanation methods at once. 
\citet{laberge2023fooling} propose a biased sampling algorithm to attack the estimation of feature attributions, which further motivates finding robust improvements for \iid sampling. 
Our research question is orthogonal to that of \emph{how to sample perturbations around an input}~\citep{petsiuk2018rise,slack2021reliable,li2021learning,ghalebikesabi2021locality,li2023glime}, or \emph{how to efficiently sample feature coalitions}~\citep{chen2018learning,covert2021improving,fumagalli2023shapiq}. 
Instead of generating samples from the conditional distribution itself, which is challenging~\citep{olsen2022using}, we explore how to efficiently select an appropriate subset of background data for explanations~\citep{hase2021outofdistribution,lundstrom2022rigorous}. 
Specifically for Shapley-based explanations, \citet{jethani2022fastshap} propose to predict them with a learned surrogate model, while \citet{kolpaczki2024approximating} propose their representation detached from the notion of marginal contribution. 
We aim to propose a general paradigm shift that benefits a broader class of explanation methods including feature effects~\citep{apley2020visualizing,moosbauer2021explaining} and expected gradients~\citep{erion2021improving,zhang2024path}. 

Concerning distribution compression, the method most related to \kt \citep{dwivedi2021kernel} is the inferior standard thinning approach~\citep{owen2017statistically}. 
\citet{cooper2023cdgrab} use insights from \kt to accelerate distributed training, while \citet{zimmerman2024resourceaware} apply \kt in robotics. 
In the context of data-centric machine learning, we broadly relate to finding coresets to improve the efficiency of clustering~\citep{agarwal2004approximating,harpeled2004coresets} and active learning~\citep{sener2018active}, as well as dataset distillation~\citep{wang2018dataset} and dataset condensation \citep{zhao2021dataset,kim2022dataset} that create synthetic samples to improve the efficiency of model training.

\section{Preliminaries}

We aim to explain a prediction model trained on labeled data and denoted by $f \colon \mX \mapsto \bbR$ where $\mX$ is the feature space; it predicts an output using an input feature vector $\bx$. 
Usually, we assume $\mX \subseteq \bbR^{d}$.
Without loss of generality, in the case of classification, we explain the output of a single class as a posterior probability from $[0,1]$. 
We can assume a given dataset $\{(\bx^{(1)},y^{(1)}), \ldots, (\bx^{(n)}, y^{(n)})\}$, where every element comes from $\mX \times \mY$, the underlying feature and label space, on which the explanations are computed. 
Depending on the explanation method and scenario, the dataset could be provided without labels.
We denote such $n \times d$ dimensional dataset by $\bbX$ where $\bx^{(i)}$ appears in the $i$-th row of $\bbX$, which is assumed to be sampled in an \iid fashion from an underlying distribution $p(\bx, y)$ defined on $\mX \times \mY$.
We denote a random vector as $\bX \in \mX$. 
Further, let $s \subset \{1, \dots, d\}$ be a feature index set of interest with its complement $\sbar = \{1, \dots, d\} \setminus s$.
We index feature vectors $\bx$ and random variables $\bX$ by index set $s$ to restrict them to these index sets. 
We write $\pXx$ and $\pXsx$ for marginal distributions on $\bX$ and $\bX_s$, respectively, and $\pXsXt(\bx_s | \bx_t)$ for conditional distribution on $\bX_s|\bX_t$. 
We use $q_{\bbX}$ to denote an empirical distribution approximating $\pX$ based on a dataset $\bbX$.

\subsection{Sampling from the dataset is prevalent in explanation estimation}

Various estimators of post-hoc explanations sample from the dataset to efficiently approximate the explanation estimate (Appendix~\ref{app:motivation-quotes}). 
For example, many removal-based explanations~\citep{covert2021explaining} like \shap~\citep{lundberg2017unified} and \sage~\citep{covert2020understanding} rely on marginalizing features out of the model function $f$ using their joint conditional distribution $\bbE_{\bXsb \sim \pXsbXs} \left[f(\bxs, \bXsb)\right] = \int f(\bxs, \bxsb) \pXsbXs(\bxsb | \bxs) d\bxsb.$
Note that the practical approximation of the conditional distribution $\pXsbXs(\bxsb | \bxs)$ itself is challenging~\citep{chen2018learning,aas2021explaining,olsen2022using} and there is no ideal solution to this problem~\citep[see a recent benchmark by][]{olsen2024comparative}. 
For example, the default for \sage is to assume feature independence and use the marginal distribution $\pXsbXs(\bxsb | \bxs) \coloneq \pXsbx$~\citep[][Appendix D]{covert2020understanding}; so does the \kernshap estimator, i.e. a practical implementation of \shap~\citep{lundberg2017unified}. 
This trend continues in more recent work sampling from marginal distribution~\citep{fumagalli2023shapiq,krzyzinski2023survshap}.

\begin{definition}[Feature marginalization] Given a set of observed values $\bxs$, we define a model function with marginalized features from the set $\sbar$ as $f(\bxs; p_{\bX}) \coloneq \bbE_{\bXsb \sim p_{\bXsb}} \left[f(\bxs, \bXsb)\right]$.
\end{definition}

In practice, the expectation $\bbE_{\bXsb \sim \pXsb} \left[f(\bxs, \bXsb)\right]$ is estimated by \iid sampling from the dataset $\bbX$ that approximates the distribution~$\pXsbx$. 
This sampled set of points forms the so-called \emph{background data}, aka \emph{reference points}, or \emph{baselines} as specifically in case of the expected gradients~\citep{erion2021improving} explanation method, which can be defined as $\expectedgradients(\bx) \coloneq \bbE_{\bX \sim p_{\bX}, \alpha \sim U(0, 1)} \left[(\bx - \bX) \cdot \frac{\partial f(\bX + \alpha \cdot (\bx - \bX))}{\partial \bx} \right].$
\citet{klein2024navigating} benchmark feature attribution methods showing that \expectedgradients is among the most faithful and robust ones.

Furthermore, \iid sampling is used in global explanation methods, many of which are an \emph{aggregation} of local explanations. 
To improve the computational efficiency of these approximations, often only a subset of $\bbX$ is considered, called \emph{foreground data}. 
Examples include: \featureeffects explanations~\citep{apley2020visualizing}, an aggregation of \lime~\citep{ribeiro2016should} into \glime~\citep{li2023glime}, and again \sage, for which points from $\bbX$ require to have their corresponding labels $y$.

\subsection{Background on distribution compression}

Standard sampling strategies can be inefficient~\citep{dwivedi2021kernel}. 
For example, the Monte Carlo estimate $\frac{1}{n} \sum^{n}_{i=1} h(\bx^{(i)})$ of an unknown expectation $\bbE_{\bX \sim \pX} h(\bX)$ based on $n$ \iid points has $\Theta(1/\sqrt{n})$ integration error $\big|\bbE_{\bX \sim \pX} h(\bX) - \frac{1}{n} \sum^{n}_{i=1} h(\bx^{(i)})\big|$ requiring $10^2$ points for 10$\%$ relative error and $10^4$ points for 1$\%$ error~\citep{shetty2022distribution}. 
To improve on \iid sampling, given a sequence $\bbX$ of $n$ input points summarizing a target distribution $p_{\bX}$, the goal of distribution compression is to identify a high quality \emph{coreset} $\bbXtilde$ of size $\ntilde \ll n$. 
This quality is measured with the coreset's integration error $\big|\frac{1}{n} \sum^{n}_{i=1} h(\bx^{(i)}) - \frac{1}{\ntilde} \sum^{\ntilde}_{i=1} h(\tilde{\bx}^{(i)}) \big|$ for functions $h$ in the reproducing kernel Hilbert space induced by a given kernel function~$\bk$~\citep{muandet2017kernel}. 
The recently introduced \kt algorithm~\citep{dwivedi2021kernel,dwivedi2022generalized} returns such a coreset that minimizes the kernel maximum mean discrepancy~\citep[$\dmmd_{\bk}$,][]{gretton2012kernel}. 
\begin{definition}[Kernel maximum mean discrepancy \citep{gretton2012kernel,dwivedi2021kernel}] Let $\bk: \bbR^d \times \bbR^d \mapsto \bbR$ be a bounded kernel function with $\bk(\bx, \cdot)$ measurable for all $\bx \in \bbR^d$, e.g. a Gaussian kernel. Kernel maximum mean discrepancy between probability distributions $p, q$ on $\bbR^d$ is defined as $\dmmd_{\bk}(p, q) \coloneq \sup_{h \in \mathcal{H}_{\bk} : \norm{h}_{\bk} \leq 1} \big| \bbE_{\bX \sim \pX} h(\bX) - \bbE_{\bX \sim q_{\bX}} h(\bX) \big|,$ where $\mathcal{H}_{\bk}$ is a reproducing kernel Hilbert induced~by~$\bk$. 
\end{definition}
To formulate Propositions~\ref{prop:bound-removal}~\&~\ref{prop:bound-global} in the next Section, we recall a biased estimate of maximum mean discrepancy, which is discussed in \citep[][remark~3.2]{Ch_rief_Abdellatif_2022} and \citep[][corollary~4]{sriperumbudur2010hilbert}.
\begin{definition}[Biased estimator of $\dmmd_{\bk}$]\label{def:mmdhat}
From \citep[][section 5.1]{muandet2017kernel}, as shown in \citep{gretton2012kernel}, the $L_2$ distance between kernel density estimates $p_{\bbX}, q_{\bbX}$ is a special case of the biased $\dmmd_{\bk}$ estimator, i.e. we have $\widehat{\dmmd}^{2}_{\bk}(p_{\bbX}, q_{\bbX}) \coloneq \norm{p_{\bbX} - q_{\bbX}}^2_2 = \int \big(p_{\bX}(\bx) - q_{\bX}(\bx)\big)^2 d\bx.$
\end{definition}

An unbiased empirical estimate of $\dmmd_{\bk}$ can be relatively easily computed given a kernel function~$\bk$~\citep{gretton2012kernel}. 
\compresspp~\citep{shetty2022distribution} is an efficient algorithm for \kt that returns a coreset of size $\sqrt{n}$ in $\mathcal{O}(n\log^3{n})$ time and $\mathcal{O}(\sqrt{n}\log^2{n})$ space, making \kt viable for large datasets. 
It was adapted to improve the kernel two-sample test~\citep{domingo2023compress}.

\section{Compress Then Explain (\ctetitle)}\label{sec:cte}

We propose distribution compression as a substitute to \iid sampling for feature marginalization in removal-based explanations and for aggregating global explanations. 
We now formalize the problem and provide theoretical intuition as to why methods for distribution compression can lead to more accurate explanation estimates. 
We defer the proofs to Appendix~\ref{app:proofs}.
\begin{definition}[Local explanation based on feature marginalization] A local explanation is a function $g(\bx; f, p_{\bX}) \colon \mX \mapsto \bbR^p$ of input $\bx$ given model $f$ that relies on a distribution $\pX$ for feature marginalization. 
For estimation, it uses an empirical distribution $q_{\bbX}$ in place of $p_{\bX}$. 
\end{definition}
Examples of such local explanations include \shap~\citep{lundberg2017unified} and \expectedgradients~\citep{erion2021improving}. We aim to provide high-quality explanations stemming from compressed samples as measured with a given approximation error, e.g. mean absolute error.

\textbf{Problem formulation.} To optimize the approximation error, we propose a novel formulation of the sample selection problem:
\begin{equation}\label{eq:problem-formulation}
\begin{aligned}
\min_{\bbXtilde} &\quad \normbig{g(\bx; f, q_{\bbX}) - g(\bx; f, q_{\bbXtilde})} \\ 
\textrm{s.t.} &\quad |\bbXtilde| = \ntilde \ll n\\
\end{aligned}
\end{equation}
for a given $\ntilde$, where \iid sampling, distribution compression, or for example clustering, are the potential methods to find $\bbXtilde$ in an unsupervised manner. 
We discover a connection between distribution compression and explanation estimation in Propositions~\ref{prop:bound-removal}~\&~\ref{prop:bound-global}.

\begin{proposition}[Feature marginalization is bounded by the maximum mean discrepancy between data samples]\label{prop:bound-removal}
For two empirical distributions $q_{\bbX}, q_{\bbXtilde}$ approximated with a kernel density estimator $\bk$, we have $\big|f(\mathbf{x}_s; q_{\bbX}) - f(\mathbf{x}_s; q_{\bbXtilde})\big| \leq C_f \cdot \widehat{\mathrm{MMD}}_{\bk}(q_{\bbX}, q_{\bbXtilde})$, where $C_f$ denotes a constant that bounds the model function $f$, i.e. $\forall_{\bx \in \bbR^p} \big|f(\bx)\big| \leq C_f$.
\end{proposition}

Proposition~\ref{prop:bound-removal} provides a worst-case bound for feature marginalization, the backbone of local explanations, in terms of the $\dmmd_{\bk}$ distance between the (often compressed) empirical data distributions. 
It complements the results for input and model perturbations obtained in \citep[][Lemmas~1~\&~4]{lin2023robustness}, which also shows how such a bound propagates to the local explanation function~$g$. 
Effectively, Proposition~\ref{prop:bound-removal} states that an algorithm minimizing $\dmmd_{\bk}$, e.g. \kt, restricts the approximation error of explanation estimation. 
This makes \cte a natural contender to improve on \iid sampling, given it was efficient and stable, which we evaluate empirically in extensive experiments.

\textbf{Compress then explain globally.} Local explanations are often aggregated into global explanations based on a representative sample from data, resulting in estimates of feature importance and effects. 

\begin{definition}[Global explanation] A global explanation is a function that aggregates local explanations $g$ of model $f$ over input samples from distribution $p_{\bX}$, i.e. $G(p_{\bX}; f, g) \coloneq \bbE_{\bX \sim p_{\bX}} \left[g(\bX; f, \cdot)\right]$.
For estimation, it uses an empirical distribution $q_{\bbX}$ in place of $p_{\bX}$.
\end{definition} 
Examples of such global explanations include \featureeffects like partial dependence plots and accumulated local effects~\citep{apley2020visualizing}, and~\sage~\citep{covert2020understanding}, which additionally requires as an input labels $y$ of the samples drawn from $p_{\bX}$. 
Notably, the local explanation function $g$ in \sage itself relies on feature marginalization leading to using $p_{\bX}$ twice, i.e. $\bbE_{\bX \sim p_{\bX}} \left[g(\bX; f, p_{\bX})\right]$~(see Listing~\ref{listing:sage} for a practical implementation). 

\textbf{Problem formulation revisited.} For global explanations, following Equation~\ref{eq:problem-formulation}, we have:
\begin{equation}
\begin{aligned}
\min_{\bbXtilde} &\quad  \normbig{G(q_{\bbX}; f, g) - G(q_{\bbXtilde}; f, g)}\\
\textrm{s.t.} &\quad |\bbXtilde| = \ntilde \ll n.\\
\end{aligned}
\end{equation}
In Proposition~\ref{prop:bound-global}, we conduct a worst-case analysis for global aggregated explanations.

\begin{proposition}[Global explanation is bounded by the maximum mean discrepancy between data samples]\label{prop:bound-global}
For two empirical distributions $q_{\bbX}, q_{\bbXtilde}$ approximated with a kernel density estimator $\bk$, we have $\normbig{G(q_{\bbX}; f, g) - G(q_{\bbXtilde}; f, g)}_2 \leq C_g\cdot\widehat{\dmmd}_{\bk}(q_{\bbX}, q_{\bbXtilde})$, where $C_g$ denotes a constant that bounds the local explanation function $g$, i.e. $\forall_{\bx \in \bbR^p} \normbig{g(\bx; \cdot)}_2 \leq C_g$.
\end{proposition}

Analogously to Proposition~\ref{prop:bound-removal}, Proposition~\ref{prop:bound-global} states that an algorithm minimizing $\dmmd_{\bk}$, e.g. \kt, restricts the approximation error of explanation estimation, which makes \cte a natural contender to improve on \iid sampling.
It extends the bounds for total variation distance obtained in~\citep{baniecki2024robustness}.
Moreover, in Section~\ref{sec:experiments-sanity-check}, we explore empirically the impact that minimizing $\dmmd_{\bk}$ has on decreasing alternative distribution discrepancies in practical (explainable) machine learning settings. It gives further intuition as to why clustering often leads to higher errors in explanation estimation. Our insights may guide future work on tighter theoretical guarantees for improving explanation estimation with distribution compression.

\textbf{Implementation.} 
The pivotal strength of \cte is that it is simple to plug into the current workflows for explanation estimation as shown in Listing~\ref{listing:sage} for \sage. 
We provide analogous code listings for \shap, \expectedgradients and \featureeffects in Appendix~\ref{app:code}.

\begin{listing*}[ht]
\small
\begin{boxminted}{2-4}{Python}
X, y, model = ...
from goodpoints import compress
ids = compress.compresspp_kt(X, kernel_type=b"gaussian", g=4)
X_compressed = X[ids]
import sage
imputer = sage.MarginalImputer(model.predict, X_compressed)
estimator = sage.KernelEstimator(imputer)
explanation = estimator(X, y)
# or even
y_compressed = y[ids]
explanation = estimator(X_compressed, y_compressed)
\end{boxminted}
\caption{Code snippet showing the 3-line plug-in of distribution compression for \sage estimation.}\label{listing:sage}
\end{listing*}

\section{Experiments}\label{sec:experiments}

In experiments, we empirically validate that the \cte paradigm improves explanation estimation across 4 methods, 2 model classes, and over 50 datasets.
We compare \cte to the widely adopted practice of \iid sampling (see Appendix~\ref{app:motivation-quotes} for further motivation). 
We also report sanity check results for a more deterministic baseline -- sampling with k-medoids -- where centroids from the clustering define a coreset from the dataset. 
We use the default hyperparameters of explanation algorithms (details are provided in Appendix~\ref{app:setup-explanation-hyperparameters}). 
For distribution compression, we use \compresspp implemented in the \texttt{goodpoints} Python package~\citep{dwivedi2021kernel}, where we follow~\citep{shetty2022distribution} to use a Gaussian kernel $\bk$ with $\sigma = \sqrt{2d}$. 
For all the compared methods, the subsampled set of points is of size $\sqrt{n}$ as we leave oversampling distribution compression for future work.
We repeat all experiments where we apply some form of downsampling before explanation estimation 33 times and report the mean and standard error (se.) or deviation (sd.) of metric values.

\textbf{Ground truth.} 
The goal of \cte is to improve explanation estimation over the standard \iid sampling. We measure the accuracy and effectiveness of explanation estimation with respect to a ``ground truth'' explanation  (cf. Appendix~\ref{app:visual-comparison}) that is \emph{estimated using a full validation dataset} $\bbX$, i.e. without sampling or compression. We consider settings where this is very inefficient to compute in practice ($n \coloneq \nvalid$ is between 1000 and 25000 samples).
For large datasets, we truncate the validation dataset to 20$\times$ the size of the compressed dataset. 
Since some explanation methods include a random component in the algorithm, we repeat their ground truth estimation 3 times and average the resulting explanations.

\textbf{Accuracy.} 
We are mainly interested in the accuracy of estimating a single explanation, measured by the explanation approximation error. 
Namely, mean absolute error (MAE), where we measure $\frac{1}{\nvalid \cdot d} \sum^{\nvalid}_{i=1} \normbig{g(\bx^{(i)}; f, q_{\bbX}) - g(\bx^{(i)}; f, q_{\bbXtilde})}_1$ for \shap and \expectedgradients, and $\frac{1}{d_G}\normbig{G(q_{\bbX}; f, g) - G(q_{\bbXtilde}; f, g)}_1$ with $d_G = d$ for \sage.  We have $d_G = 100\cdot(d + d^2)$ for \featureeffects, since we use 100 uniformly distributed grid points for 1-dimensional effects and 10$\times$10 uniformly distributed grid points for 2-dimensional effects (see Appendix~\ref{app:experiments-setup}). 
For broader context, in Section~\ref{sec:experiments-accuracy}, we also measure the precision of correctly indicating the top $k$ features.

\textbf{Efficiency.} 
We measure the efficiency of compression and explanation estimation with CPU wall-clock time (in seconds), assuming the time of \iid sampling is~$0$. 
We rely on popular open-source implementations of the algorithms (see Appendix~\ref{app:code}) and perform efficiency experiments on a personal computer with an M3 chip. 
This is to imitate the most standard workflow of explanation estimation, while we acknowledge that specific time estimates will vary in more sophisticated settings.

\subsection{\ctetitle improves the accuracy of estimating feature attributions \& importance}\label{sec:experiments-accuracy}

\begin{figure*}[t]
    \centering
    \includegraphics[width=0.495\textwidth]{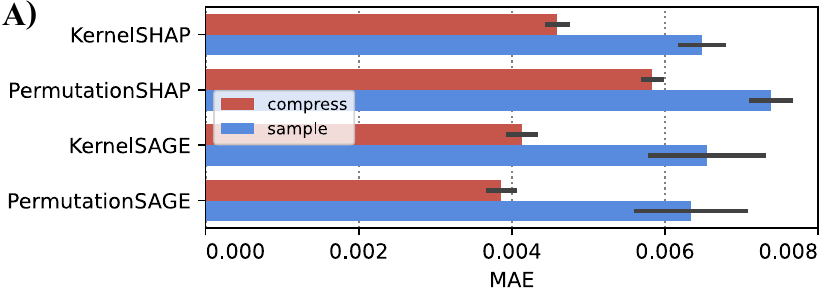}
    \includegraphics[width=0.495\textwidth]{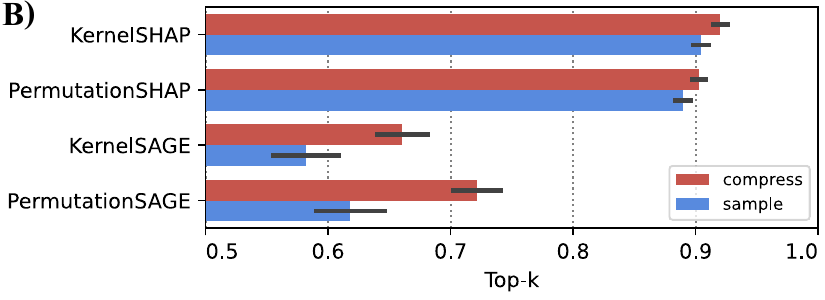}
    \includegraphics[width=0.495\textwidth]{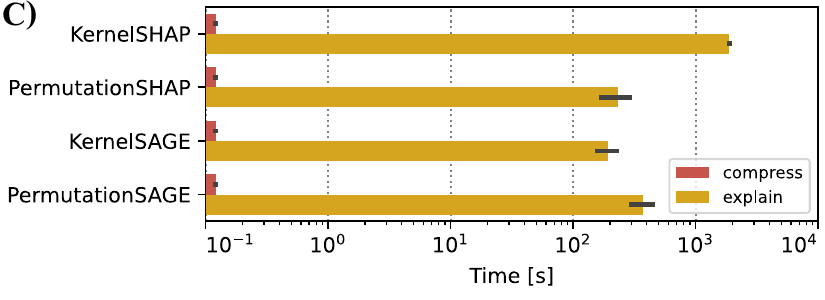}
    \includegraphics[width=0.495\textwidth]{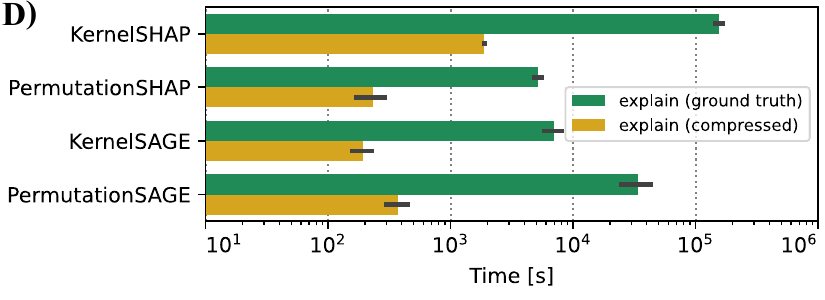}
    \caption{
    Comparison between \cte and \iid sampling for the two estimators of \shap and \sage explanations on the \texttt{adult} dataset. \textbf{A)}~We measure mean absolute error (MAE~$\downarrow$) between feature attribution and importance values, and \textbf{B)}~the precision in correctly identifying the 5 most important features (Top-k~$\uparrow$). 
    \textbf{C)}~Comparison between the computational time of distribution compression and explanation estimation on the compressed sample, assuming the time of \iid sampling is~$0$.
    \textbf{D)}~Comparison between the computational time of explanation estimation on the compressed sample and on full data (in green).
    Analogous results for the other 4 datasets are in Appendix~\ref{app:experiments-openxai}. (mean $\pm$ se.)}
    \label{fig:adult-shap-sage}
\end{figure*}

We use the preprocessed datasets and pretrained neural network models from the well-established OpenXAI benchmark~\citep{agarwal2022openxai}. 
We filter out three datasets with less than 1000 observations in the validation set, where sampling is not crucial, which results in five tasks: \texttt{adult}~($\nvalid=9045$, $d=13$), \texttt{compas} ($\nvalid=1235$, $d=7$), \texttt{gaussian} (a synthetic dataset, $\nvalid=1250$, $d=20$), \texttt{gmsc} (Give Me Some Credit, $\nvalid=20442$, $d=10$), and \texttt{heloc} (aka FICO, $\nvalid=1975$, $d=23$). 
Further details on datasets and models are provided in Appendix~\ref{app:setup-datasets-models}. 

We aim to show that \cte improves the estimation of feature attribution and importance explanations, namely for \shap and \sage. 
We experiment with two model-agnostic estimators: kernel-based and permutation-based. 
For example, Figure~\ref{fig:adult-shap-sage} shows the differences in MAE and Top-k between \cte and standard \iid sampling on the \texttt{adult} dataset. 
Analogous results for the other 4 datasets from OpenXAI are shown in Appendix~\ref{app:experiments-openxai}. 
On all the considered tasks, \cte results in a notable decrease in approximation error when compared to \iid sampling and an increase in precision (for top-$k$ feature identification) with negligible computational overhead.
Moreover, \cte results in explanation estimates of \underline{significantly smaller variance} on average as shown in Table~\ref{tab:openxai-mae}.

\begin{table}[t]
    \centering
    \caption{We report the improvement in the mean absolute error of estimating four popular explanations on five datasets from the OpenXAI benchmark. \cte not only improves the accuracy over \iid by 20--45\%, but also leads to more stable estimates by about 50\%. MAE ($\downarrow$, $\pm$ sd.) values are scaled and rounded to improve readability while detailed and extended results are reported in Appendix~\ref{app:experiments-openxai}.}
    \label{tab:openxai-mae}
    \small
    \begin{tabular}{l|cccc}
        \toprule
         \textbf{Task} & \multicolumn{4}{c}{\textbf{Explanation estimator}} \\
          & \kernshap & \permshap & \kernsage & \permsage \\
        \midrule
         \texttt{adult} & 
         \textbf{\big( {\color{myblue}\iid} $\xrightarrow{\mathrm{\textbf{diff.}}}$ {\color{myred}\cte} \big)} &
         $73_{\pm14}\xrightarrow{21\%}58_{\pm6}$ &
         $65_{\pm42}\xrightarrow{37\%}41_{\pm10}$ & 
         $63_{\pm40}\xrightarrow{40\%}38_{\pm10}$ \\
         \texttt{compas} & 
         $10_{\pm4}\xrightarrow{40\%}6_{\pm2}$ &
         $11_{\pm4}\xrightarrow{45\%}6_{\pm2}$ &
         $29_{\pm17}\xrightarrow{38\%}18_{\pm9}$ & 
         $28_{\pm16}\xrightarrow{39\%}17_{\pm8}$ \\
         \texttt{gaussian} &
         $13_{\pm2}\xrightarrow{38\%}8_{\pm1}$ &
         $15_{\pm2}\xrightarrow{27\%}11_{\pm1}$ &
         $52_{\pm27}\xrightarrow{42\%}30_{\pm7}$ & 
         $52_{\pm26}\xrightarrow{44\%}29_{\pm7}$ \\
         \texttt{gmsc} &
         $23_{\pm6}\xrightarrow{39\%}14_{\pm3}$ &
         $25_{\pm5}\xrightarrow{32\%}17_{\pm3}$ &
         $30_{\pm13}\xrightarrow{40\%}18_{\pm5}$ & 
         $28_{\pm14}\xrightarrow{43\%}16_{\pm5}$ \\
         \texttt{heloc} &
         $67_{\pm15}\xrightarrow{39\%}41_{\pm7}$ &
         $72_{\pm13}\xrightarrow{33\%}48_{\pm6}$ &
         $34_{\pm10}\xrightarrow{21\%}27_{\pm6}$ & 
         $29_{\pm11}\xrightarrow{28\%}21_{\pm6}$ \\
        \bottomrule
    \end{tabular}
\end{table}

\subsection{\ctetitle as an efficient alternative to \iid sampling in explanation estimation}\label{sec:experiments-efficient}
\begin{wrapfigure}{r}{0.38\textwidth}
    \centering
    \vspace{-0.5em}
    \includegraphics[width=0.31\textwidth]{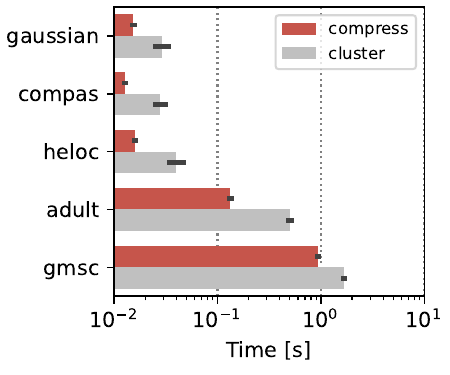}
    \caption{Compressing a distribution from 20k to 128 samples takes less than 1 second to compute on a CPU. (mean $\pm$ se.)}
    \label{fig:benchmark_compress_time}
\end{wrapfigure}

We find \cte to be a very efficient alternative to standard \iid sampling in explanation estimation. 
For example, compressing a distribution from 1k to 32 samples takes less than 0.1 seconds, and from 20k to 128 samples takes less than 1 second. 
The exact runtime will, of course, differ based on the number of features. Figure~\ref{fig:benchmark_compress_time} reports the wall-clock time for datasets of different sizes. 
Note that the potential runtimes for distribution compression are of magnitudes smaller than the typical runtime of explanation estimation. 
For example, estimating \kernshap for 9k samples using 128 background samples takes 30 minutes, which is about 60$\times$ less than estimating the ground truth explanation~(Figure~\ref{fig:adult-shap-sage}). 
Moreover, estimating \kernshap or \permsage for 1k samples using 32 background samples takes about 10 seconds, which is about 30$\times$ less than estimating the ground truth explanation (Appendix~\ref{app:experiments-openxai}).

\subsection{\ctetitle improves gradient-based explanations specific to neural networks}\label{sec:experiments-expected-gradients}
\begin{wrapfigure}{r}{0.38\textwidth}
    \centering
    \vspace{-0.5em}
    \includegraphics[width=0.32\textwidth]{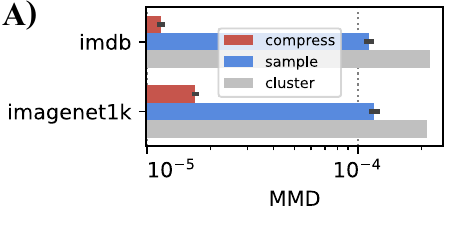}
    \includegraphics[width=0.32\textwidth]{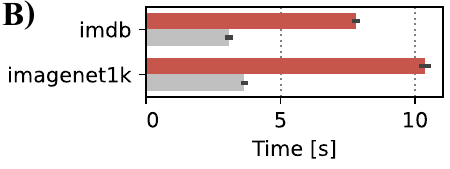}
    \caption{\textbf{A)} \compresspp effectively optimizes $\dmmd_{\bk}$ on unstructured IMDB and Imagenet-1k datasets. \textbf{B)} Compressing a distribution from 25k--50k to 128 samples in 512--768 dimensions takes about 5--10 sec. to compute on a CPU. (mean $\pm$ se.)}
    \label{fig:benchmark-compress-text-image}
\end{wrapfigure}

Next, we aim to show the broader applicability of \cte by evaluating it on gradient-based explanations specific to neural networks, often fitted to larger unstructured datasets.

\textbf{Sanity check.} 
We first compress the validation sets of IMDB and ImageNet-1k on a single CPU as a sanity check for the viability of \cte in settings considering larger datasets. For the IMDB dataset ($\nvalid=25000$, $d=768$), \cte takes as an input text embeddings from the pretrained DistilBERT model's last layer (preceding a classifier) that has a dimension of size 768. Similarly, for ImageNet-1k ($\nvalid=50000$, $d=512$), \cte operates on the hidden representation extracted from ResNet-18. 
Figure~\ref{fig:benchmark-compress-text-image} shows the optimized $\dmmd_{\bk}$ metric between the distributions and computation time in seconds. 
We can see that proper compression results in huge benefits w.r.t. $\dmmd_{\bk}$ (compared to \iid sampling and clustering) and only negligible computational overhead.

\textbf{Accuracy and efficiency.} 
We now study \cte together with \expectedgradients of neural network models trained to 18 datasets ($\nvalid > 1000$, $d \geq 32$) from the OpenML-CC18~\citep{bischl2021openml} and OpenML-CTR23~\citep{fischer2023openmlctr} benchmark suites. 
Details on datasets and models are provided in Appendix~\ref{app:setup-datasets-models}. 
In Figure~\ref{fig:openml-expected-gradients-small}, we show the explanation approximation error for 4 image classification tasks, while analogous results for the remaining 14 datasets are provided in Appendix~\ref{app:experiments-expected-gradients}.
Additionally here, we vary the number of data points sampled from \iid to inspect the increase in efficiency of \cte.
In all cases, \cte achieves on-par approximation error using fewer samples than \iid sampling, i.e. requiring fewer model inferences, resulting in faster computation and saved resources. 
The accuracy improvements are significant, i.e. \cte decreases the estimation error for \expectedgradients by 35\% on \texttt{mnist\_784} (Welch's t-test: $p < 1\mathrm{e}{-10}$), by 40\% on \texttt{Fashion-MNIST} ($p < 1\mathrm{e}{-10}$), and by 21\% on \texttt{CIFAR\_10} ($p < 1\mathrm{e}{-10}$). 
Moreover, \cte provides $2$--$3\times$ efficiency improvements as measured by the number of samples required for \iid to reach the error of \cte.

\begin{figure}[t]
    \centering
    \includegraphics[width=0.99\textwidth]{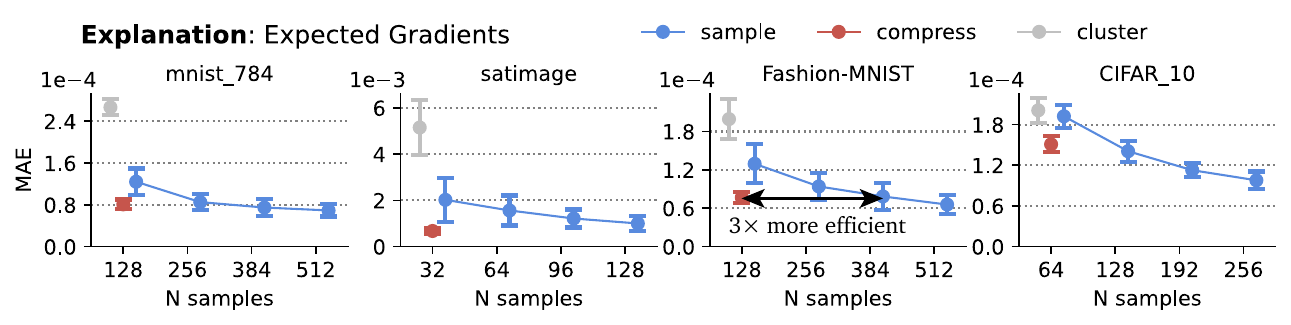}
    \caption{ 
    Comparison between \cte, \iid sampling and clustering for \expectedgradients explanations on the 4 image classification datasets. 
    We measure mean absolute error (MAE~$\downarrow$) between feature attribution values.
    \cte is not only more accurate but also more stable as measured with deviation.
    Analogous results for the remaining 14 datasets are in Appendix~\ref{app:experiments-expected-gradients}. (mean $\pm$~sd.)}
    \label{fig:openml-expected-gradients-small}
\end{figure}

\textbf{Model-agnostic explanation of a language model.} In Appendix~\ref{app:experiments-expected-gradients}, we further experiment with applying \cte to improve the estimation of \glime~\citep{li2023glime} explaining the predictions of a DistilBERT language model trained on the IMDB dataset for sentiment analysis.

\subsection{Ablations with another 30 datasets, an XGBoost model \& feature effects}\label{sec:experiments-openml}

For a convincing case to use \cte instead of \iid in practice, we perform additional empirical analysis on various datasets, with a different model class, and include another global explanation method. 
More specifically, we use \cte to improve \featureeffects of XGBoost models trained on further 30 datasets ($\nvalid > 1000$, $d < 32$) from OpenML-CC18 and OpenML-CTR23. 
Details on datasets and models are provided in Appendix~\ref{app:setup-datasets-models}. 
Moreover, we include \shap and \sage in the benchmark similarly to Section~\ref{sec:experiments-accuracy}. 
As another ablation, \sage is evaluated in two variants that consider either compressing only the background data (a rather typical scenario), or using the compressed samples as both background and foreground data (as indicated with ``fg.''; refer to Listing~\ref{listing:sage} for this distinction).

\begin{figure}[ht]
    \centering
    \includegraphics[width=0.99\textwidth]{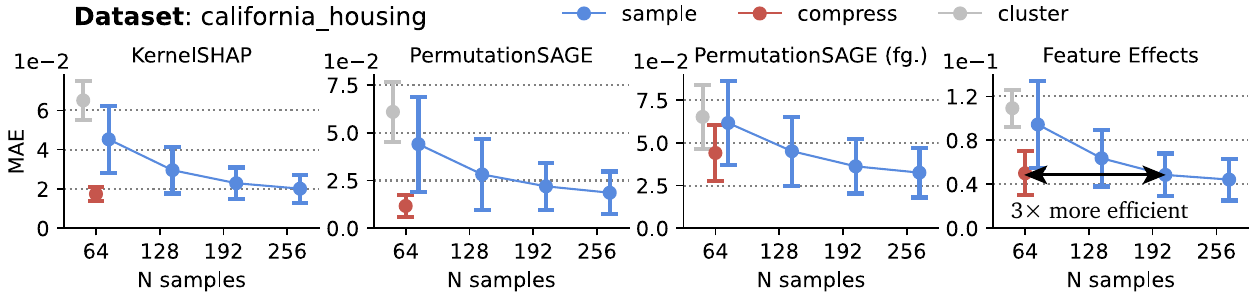}
    \includegraphics[width=0.99\textwidth]{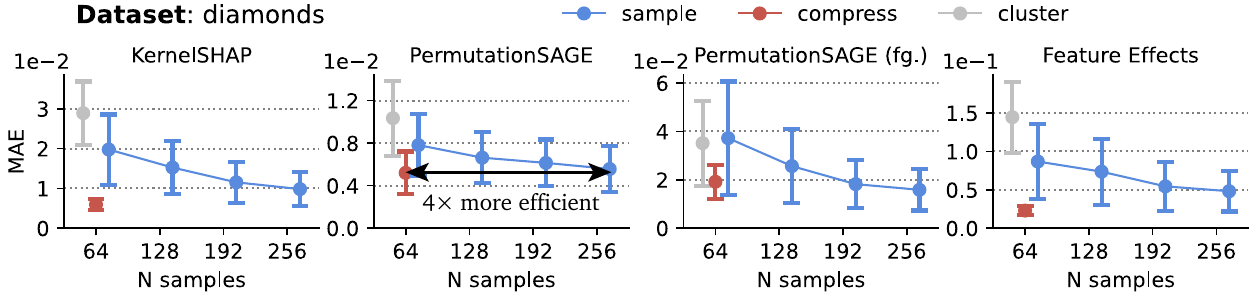}
    \includegraphics[width=0.99\textwidth]{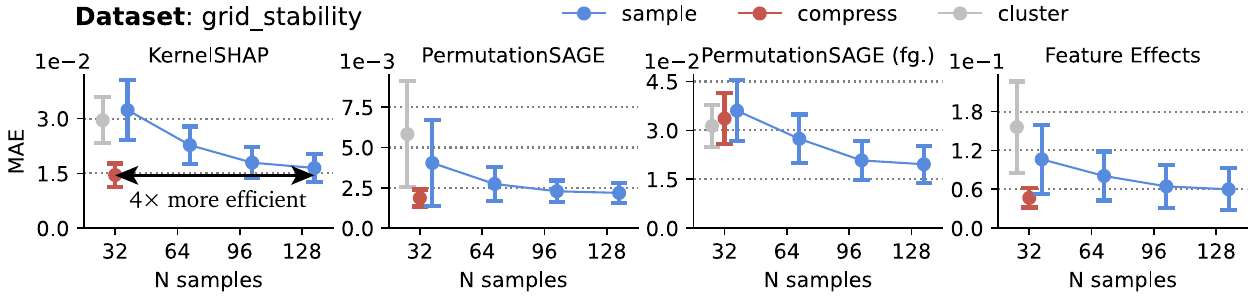}
    \caption{\cte improves the approximation error of local and global removal-based explanations. \sage is evaluated in two variants that consider either compressing only the background data (default), or using the compressed samples as both background and foreground data (as indicated with ``fg.''). Analogous results for the remaining estimators and 27 datasets are in Appendix~\ref{app:experiments-openml}. (mean $\pm$ sd.)}
    \label{fig:openml-shap-sage-small}
\end{figure}

Figure~\ref{fig:openml-shap-sage-small} shows the explanation approximation error for 3 predictive tasks, while analogous results for the remaining explanation estimators and 27 datasets are provided in Appendix~\ref{app:experiments-openml}. 
We observe that \cte significantly improves the estimation of \featureeffects in all cases. 
We further confirm the conclusions from Sections~\ref{sec:experiments-accuracy}~\&~\ref{sec:experiments-efficient} that \cte improves \shap and \sage. 
Another insight is that, on average, \cte provides a smaller improvement over \iid sampling when considering compressing foreground data in \sage.

\begin{figure}[h]
    \includegraphics[width=0.49\textwidth]{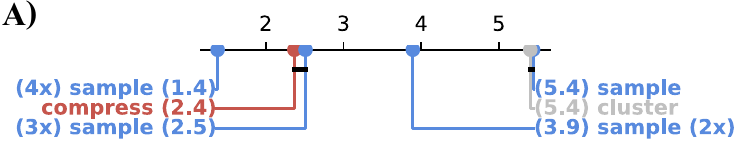}
    \includegraphics[width=0.49\textwidth]{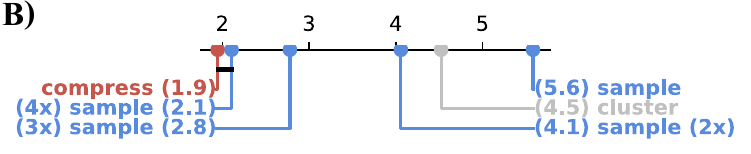}
    \caption{Critical difference diagrams of average ranks (lower is better) aggregated over 6 explanation estimators and 48 dataset--model pairs: \textbf{A)} for MAE averaged over repeats, and \textbf{B)} for the sd. of MAE over repeats that corresponds to the stability of explanation estimation. \cte often achieves on-par explanation approximation error using 2--3$\times$ fewer samples, i.e. requiring 2--3$\times$ fewer model inferences, which is efficient. Moreover, \cte guarantees more stable estimates than \iid sampling.}
    \label{fig:cdd-mae}
\end{figure}

\textbf{Conclusion from the experiments.} 
In Figure~\ref{fig:cdd-mae}, we aggregated the results from Sections~\ref{sec:experiments-expected-gradients}~\&~\ref{sec:experiments-openml} to conclude the main claim that \cte offers 2--3$\times$ improvements in efficiency over \iid sampling.

\subsection{Kernel thinning on datasets for (explainable) machine learning}\label{sec:experiments-sanity-check}

We have already established that distribution compression is a viable approach to data sampling, which regularly entails a better approximation of explanations. 
Moreover, its computational overhead is negligible when applied before explanation estimation. 
To provide more context on discovering the theoretical justification for \cte from Section~\ref{sec:cte}, we aim to show that \compresspp entails a better approximation of feature distribution on popular datasets for (explainable) machine learning. 
This is out-of-the-box, without tuning its hyperparameters, which is a natural direction for future work. 

\textbf{Measuring distribution change.} 
In general, measuring the similarity of distributions or datasets is challenging, and many metrics with various properties have been proposed for this task~\citep{gibbs2002choosing}. 
Here, we report the following distance metric values between the original and compressed distribution: the optimized $\dmmd_{\bk}$, total variation distance \citep[TV,][]{gibbs2002choosing}, Kullback–Leibler divergence~\citep[KL,][]{gibbs2002choosing}, and d-dimensional Wasserstein distance~\citep[WD,][]{feydy2019interpolating,laberge2023fooling}. 
Since approximating d-dimensional TV and KL is infeasible in practice \citep[see e.g.][section 5]{sriperumbudur2012empirical}, we report an average of the top-3 largest discrepancies between the 1-dimensional distributions of features as a proxy.

\begin{figure}[ht]
    \centering
    \includegraphics[width=0.495\textwidth]{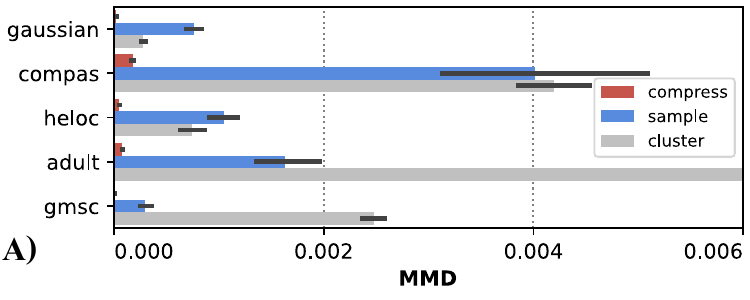}
    \includegraphics[width=0.495\textwidth]{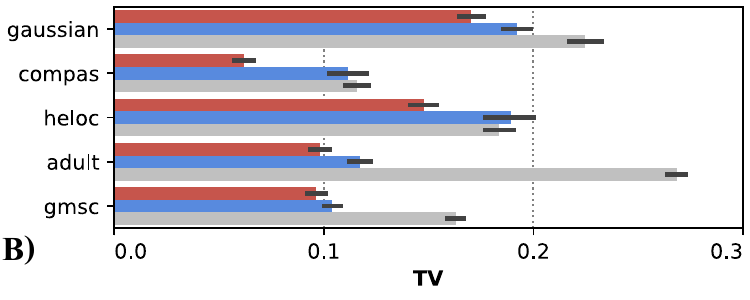}
    \includegraphics[width=0.495\textwidth]{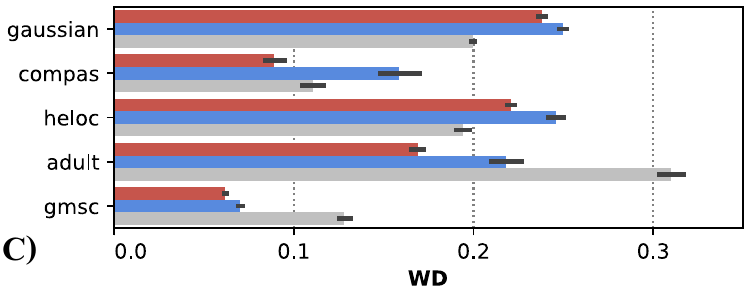}
    \includegraphics[width=0.495\textwidth]{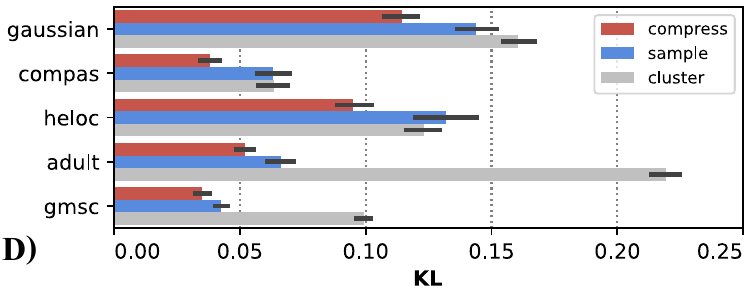}
    \caption{\compresspp with Gaussian kernel on five datasets (rows) for the four considered distribution change metrics (panels): 
    \textbf{A)} $\dmmd_{\bk}$, \textbf{B)} total variation distance, \textbf{C)} d-dimensional Wasserstein distance, and \textbf{D)} Kullback–Leibler divergence. 
    The length of the bar is the mean value $\pm$ standard error across statistical repetitions, where color indicates the applied downsampling method.
    }
    \label{fig:benchmark_compresss_distance}
\end{figure}

\textbf{Result.} 
In Figure~\ref{fig:benchmark_compresss_distance}, we observe that \compresspp works much better in terms of $\dmmd_{\bk}$ on all five datasets from OpenXAI, compared to standard \iid sampling or the clustering baseline, which is no surprise as this metric is internally optimized by the former.  
Note that it also leads to notable improvements in all other metrics.
Overall, there is no consistent improvement in approximating the distribution using clustering, which explains why it leads to higher error in explanation estimation.

\section{Discussion}\label{sec:conclusion}

We propose \emph{compress then explain} as a powerful alternative to the conventional \emph{sample then explain} paradigm in explanation estimation. 
\cte has the potential to improve approximation error across a wide range of explanation methods for various predictive tasks. 
Specifically, we show accuracy and stability improvements in popular removal-based explanations that marginalize feature influence, and in general, global explanations that aggregate local explanations over a subset of data. 
Moreover, \cte leads to more efficient explanation estimation by decreasing the computational resources (time, model inferences) required to achieve error on par with a larger \iid sample size. 

\textbf{Future work} on methods for marginal distribution compression other than kernel thinning and clustering will bring further improvements in the performance of explanation estimation.
Distribution compression methods, by design, have hyperparameters that may impact the empirical results. 
Although we have shown that the default \compresspp algorithm is a robust baseline, exploring the tunability of its hyperparameters is a natural future work direction~\citep[similarly as in the case of conditional sampling methods,][]{olsen2024comparative}. 
We used the Gaussian kernel because it is the standard in the field of distribution compression~\citep{dwivedi2022generalized,shetty2022distribution,domingo2023compress}, especially in experimental analysis, and is generally adopted within machine learning applications. 
Although our empirical validation shows that the Gaussian kernel works well for over 50 datasets, exploring other kernels for which theoretical thinning error bounds exist, like Mat\'ern or B-spline~\citep{dwivedi2021kernel}, is a viable future work direction.
Furthermore, especially for tabular datasets, dealing with categorical features can be an issue, which we elaborate on in Appendix~\ref{app:setup-datasets-models}. 
Specifically for supervised learning, a stratified variant of kernel thinning taking into account a distribution of the target feature could further improve loss-based explanations like \sage, or even the estimation of group fairness metrics.
In concurrent work, \citet{gong2024supervised} generalize \kt to speed up supervised learning problems involving kernel methods.
One could also investigate how influence functions \citep{koh2017understanding}, which aim to attribute the importance of data to the model’s prediction, can guide sampling for efficient explanation estimation.

\textbf{Broader impact.} In general, improving explanation methods has positive implications for humans interacting with AI systems~\citep{rong2024towards}. But, specifically in the context of this work, biased sampling can be exploited to manipulate the explanation results~\citep{slack2020fooling,baniecki2022manipulating,laberge2023fooling}. \cte could minimize the risk of such adversaries and prevent ``random seed/state hacking'' based on the rather unstable \iid sampling from data in empirical research~\citep{herrmann2024position}.

\textbf{Code.} 
We provide additional details on reproducibility in the Appendix, as well as the code to reproduce all experiments in this paper is available at \url{https://github.com/hbaniecki/compress-then-explain}.

\section*{Acknowledgments}

This work was financially supported by the Polish National Science Centre grant number 2021/43/O/ST6/00347, and carried out with the support of the Laboratory of Bioinformatics and Computational Genomics and the High Performance Computing Center of the Faculty of Mathematics and Information Science, Warsaw University of Technology. Hubert Baniecki gratefully acknowledges scholarship funding from the Polish National Agency for Academic Exchange under the Preludium Bis NAWA 3 program. We want to thank Maximilian Muschalik, Mateusz Biesiadowski, Paulina Kaczynska, Anna Semik, and anonymous reviewers for their valuable feedback regarding this work.

\bibliography{references}

\begin{thebibliography}{66}
\providecommand{\natexlab}[1]{#1}
\providecommand{\url}[1]{\texttt{#1}}
\expandafter\ifx\csname urlstyle\endcsname\relax
  \providecommand{\doi}[1]{doi: #1}\else
  \providecommand{\doi}{doi: \begingroup \urlstyle{rm}\Url}\fi

\bibitem[Aas et~al.(2021)Aas, Jullum, and Løland]{aas2021explaining}
Kjersti Aas, Martin Jullum, and Anders Løland.
\newblock Explaining individual predictions when features are dependent: More accurate approximations to {Shapley} values.
\newblock \emph{Artificial Intelligence}, 298:\penalty0 103502, 2021.

\bibitem[Agarwal et~al.(2022)Agarwal, Krishna, Saxena, Pawelczyk, Johnson, Puri, Zitnik, and Lakkaraju]{agarwal2022openxai}
Chirag Agarwal, Satyapriya Krishna, Eshika Saxena, Martin Pawelczyk, Nari Johnson, Isha Puri, Marinka Zitnik, and Himabindu Lakkaraju.
\newblock {OpenXAI}: Towards a transparent evaluation of model explanations.
\newblock In \emph{NeurIPS}, 2022.

\bibitem[Agarwal et~al.(2004)Agarwal, Har-Peled, and Varadarajan]{agarwal2004approximating}
Pankaj~K. Agarwal, Sariel Har-Peled, and Kasturi~R. Varadarajan.
\newblock Approximating extent measures of points.
\newblock \emph{Journal of the ACM}, 51\penalty0 (4):\penalty0 606--635, 2004.

\bibitem[Apley \& Zhu(2020)Apley and Zhu]{apley2020visualizing}
Daniel~W. Apley and Jingyu Zhu.
\newblock Visualizing the effects of predictor variables in black box supervised learning models.
\newblock \emph{Journal of the Royal Statistical Society: Series B (Statistical Methodology)}, 82\penalty0 (4):\penalty0 1059--1086, 2020.

\bibitem[Baniecki \& Biecek(2022)Baniecki and Biecek]{baniecki2022manipulating}
Hubert Baniecki and Przemyslaw Biecek.
\newblock {Manipulating SHAP via Adversarial Data Perturbations (Student Abstract)}.
\newblock In \emph{AAAI}, 2022.

\bibitem[Baniecki et~al.(2024)Baniecki, Casalicchio, Bischl, and Biecek]{baniecki2024robustness}
Hubert Baniecki, Giuseppe Casalicchio, Bernd Bischl, and Przemyslaw Biecek.
\newblock {On the Robustness of Global Feature Effect Explanations}.
\newblock In \emph{ECML PKDD}, 2024.

\bibitem[Bischl et~al.(2021)Bischl, Casalicchio, Feurer, Gijsbers, Hutter, Lang, Mantovani, van Rijn, and Vanschoren]{bischl2021openml}
Bernd Bischl, Giuseppe Casalicchio, Matthias Feurer, Pieter Gijsbers, Frank Hutter, Michel Lang, Rafael~Gomes Mantovani, Jan~N. van Rijn, and Joaquin Vanschoren.
\newblock Open{ML} benchmarking suites.
\newblock In \emph{NeurIPS}, 2021.

\bibitem[Chen et~al.(2023)Chen, Covert, Lundberg, and Lee]{chen2023algorithms}
Hugh Chen, Ian~C Covert, Scott~M Lundberg, and Su-In Lee.
\newblock Algorithms to estimate {Shapley} value feature attributions.
\newblock \emph{Nature Machine Intelligence}, 5\penalty0 (6):\penalty0 590--601, 2023.

\bibitem[Chen et~al.(2018)Chen, Song, Wainwright, and Jordan]{chen2018learning}
Jianbo Chen, Le~Song, Martin~J. Wainwright, and Michael~I. Jordan.
\newblock Learning to explain: An information-theoretic perspective on model interpretation.
\newblock In \emph{ICML}, 2018.

\bibitem[Chen et~al.(2024)Chen, Zhang, Liang, Li, and Cao]{chen2024less}
Ruoyu Chen, Hua Zhang, Siyuan Liang, Jingzhi Li, and Xiaochun Cao.
\newblock Less is more: Fewer interpretable region via submodular subset selection.
\newblock In \emph{ICLR}, 2024.

\bibitem[Chen \& Sun(2023)Chen and Sun]{chen2023extracting}
Zhaozheng Chen and Qianru Sun.
\newblock Extracting class activation maps from non-discriminative features as well.
\newblock In \emph{CVPR}, 2023.

\bibitem[Chérief-Abdellatif \& Alquier(2022)Chérief-Abdellatif and Alquier]{Ch_rief_Abdellatif_2022}
Badr-Eddine Chérief-Abdellatif and Pierre Alquier.
\newblock Finite sample properties of parametric mmd estimation: Robustness to misspecification and dependence.
\newblock \emph{Bernoulli}, 28\penalty0 (1), 2022.

\bibitem[Cooper et~al.(2023)Cooper, Guo, Pham, Yuan, Ruan, Lu, and Sa]{cooper2023cdgrab}
A.~Feder Cooper, Wentao Guo, Khiem Pham, Tiancheng Yuan, Charlie~F. Ruan, Yucheng Lu, and Christopher~De Sa.
\newblock Coordinating distributed example orders for provably accelerated training.
\newblock In \emph{NeurIPS}, 2023.

\bibitem[Covert \& Lee(2021)Covert and Lee]{covert2021improving}
Ian Covert and Su-In Lee.
\newblock Improving {KernelSHAP}: Practical {Shapley} value estimation via linear regression.
\newblock In \emph{AISTATS}, 2021.

\bibitem[Covert et~al.(2020)Covert, Lundberg, and Lee]{covert2020understanding}
Ian Covert, Scott~M Lundberg, and Su-In Lee.
\newblock Understanding global feature contributions with additive importance measures.
\newblock In \emph{NeurIPS}, 2020.

\bibitem[Covert et~al.(2021)Covert, Lundberg, and Lee]{covert2021explaining}
Ian Covert, Scott Lundberg, and Su-In Lee.
\newblock Explaining by removing: A unified framework for model explanation.
\newblock \emph{Journal of Machine Learning Research}, 22\penalty0 (209):\penalty0 1--90, 2021.

\bibitem[Domingo-Enrich et~al.(2023)Domingo-Enrich, Dwivedi, and Mackey]{domingo2023compress}
Carles Domingo-Enrich, Raaz Dwivedi, and Lester Mackey.
\newblock Compress then test: Powerful kernel testing in near-linear time.
\newblock In \emph{AISTATS}, 2023.

\bibitem[Donnelly et~al.(2023)Donnelly, Katta, Rudin, and Browne]{donnelly2023rashomon}
Jon Donnelly, Srikar Katta, Cynthia Rudin, and Edward Browne.
\newblock The {Rashomon} importance distribution: Getting {RID} of unstable, single model-based variable importance.
\newblock In \emph{NeurIPS}, 2023.

\bibitem[Dwivedi \& Mackey(2021)Dwivedi and Mackey]{dwivedi2021kernel}
Raaz Dwivedi and Lester Mackey.
\newblock Kernel thinning.
\newblock In \emph{COLT}, 2021.

\bibitem[Dwivedi \& Mackey(2022)Dwivedi and Mackey]{dwivedi2022generalized}
Raaz Dwivedi and Lester Mackey.
\newblock Generalized kernel thinning.
\newblock In \emph{ICLR}, 2022.

\bibitem[Erion et~al.(2021)Erion, Janizek, Sturmfels, Lundberg, and Lee]{erion2021improving}
Gabriel Erion, Joseph~D Janizek, Pascal Sturmfels, Scott~M Lundberg, and Su-In Lee.
\newblock Improving performance of deep learning models with axiomatic attribution priors and expected gradients.
\newblock \emph{Nature Machine Intelligence}, 3\penalty0 (7):\penalty0 620--631, 2021.

\bibitem[Feydy et~al.(2019)Feydy, S{\'e}journ{\'e}, Vialard, Amari, Trouve, and Peyr{\'e}]{feydy2019interpolating}
Jean Feydy, Thibault S{\'e}journ{\'e}, Fran{\c{c}}ois-Xavier Vialard, Shun-ichi Amari, Alain Trouve, and Gabriel Peyr{\'e}.
\newblock Interpolating between optimal transport and {MMD} using sinkhorn divergences.
\newblock In \emph{AISTATS}, 2019.

\bibitem[Fischer et~al.(2023)Fischer, Feurer, and Bischl]{fischer2023openmlctr}
Sebastian~Felix Fischer, Matthias Feurer, and Bernd Bischl.
\newblock Open{ML}-{CTR}23 {\textendash} a curated tabular regression benchmarking suite.
\newblock In \emph{AutoML}, 2023.

\bibitem[Fumagalli et~al.(2023)Fumagalli, Muschalik, Kolpaczki, H\"{u}llermeier, and Hammer]{fumagalli2023shapiq}
Fabian Fumagalli, Maximilian Muschalik, Patrick Kolpaczki, Eyke H\"{u}llermeier, and Barbara Hammer.
\newblock {SHAP-IQ}: Unified approximation of any-order {Shapley} interactions.
\newblock In \emph{NeurIPS}, 2023.

\bibitem[Ghalebikesabi et~al.(2021)Ghalebikesabi, Ter-Minassian, DiazOrdaz, and Holmes]{ghalebikesabi2021locality}
Sahra Ghalebikesabi, Lucile Ter-Minassian, Karla DiazOrdaz, and Chris~C Holmes.
\newblock On locality of local explanation models.
\newblock In \emph{NeurIPS}, 2021.

\bibitem[Gibbs \& Su(2002)Gibbs and Su]{gibbs2002choosing}
Alison~L Gibbs and Francis~Edward Su.
\newblock On choosing and bounding probability metrics.
\newblock \emph{International Statistical Review}, 70\penalty0 (3):\penalty0 419--435, 2002.

\bibitem[Gong et~al.(2024)Gong, Choi, and Dwivedi]{gong2024supervised}
Albert Gong, Kyuseong Choi, and Raaz Dwivedi.
\newblock Supervised kernel thinning.
\newblock In \emph{NeurIPS}, 2024.

\bibitem[Gretton et~al.(2012)Gretton, Borgwardt, Rasch, Sch{\"o}lkopf, and Smola]{gretton2012kernel}
Arthur Gretton, Karsten~M Borgwardt, Malte~J Rasch, Bernhard Sch{\"o}lkopf, and Alexander Smola.
\newblock A kernel two-sample test.
\newblock \emph{Journal of Machine Learning Research}, 13\penalty0 (1):\penalty0 723--773, 2012.

\bibitem[Har-Peled \& Mazumdar(2004)Har-Peled and Mazumdar]{harpeled2004coresets}
Sariel Har-Peled and Soham Mazumdar.
\newblock On coresets for k-means and k-median clustering.
\newblock In \emph{STOC}, 2004.

\bibitem[Hase et~al.(2021)Hase, Xie, and Bansal]{hase2021outofdistribution}
Peter Hase, Harry Xie, and Mohit Bansal.
\newblock The out-of-distribution problem in explainability and search methods for feature importance explanations.
\newblock In \emph{NeurIPS}, 2021.

\bibitem[Herrmann et~al.(2024)Herrmann, Lange, Eggensperger, Casalicchio, Wever, Feurer, Rügamer, Hüllermeier, Boulesteix, and Bischl]{herrmann2024position}
Moritz Herrmann, F.~Julian~D. Lange, Katharina Eggensperger, Giuseppe Casalicchio, Marcel Wever, Matthias Feurer, David Rügamer, Eyke Hüllermeier, Anne-Laure Boulesteix, and Bernd Bischl.
\newblock Position paper: Rethinking empirical research in machine learning: Addressing epistemic and methodological challenges of experimentation.
\newblock In \emph{ICML}, 2024.

\bibitem[Jethani et~al.(2022)Jethani, Sudarshan, Covert, Lee, and Ranganath]{jethani2022fastshap}
Neil Jethani, Mukund Sudarshan, Ian~Connick Covert, Su-In Lee, and Rajesh Ranganath.
\newblock Fast{SHAP}: Real-time {Shapley} value estimation.
\newblock In \emph{ICLR}, 2022.

\bibitem[Kim et~al.(2022)Kim, Kim, Oh, Yun, Song, Jeong, Ha, and Song]{kim2022dataset}
Jang-Hyun Kim, Jinuk Kim, Seong~Joon Oh, Sangdoo Yun, Hwanjun Song, Joonhyun Jeong, Jung-Woo Ha, and Hyun~Oh Song.
\newblock Dataset condensation via efficient synthetic-data parameterization.
\newblock In \emph{ICML}, 2022.

\bibitem[Klein et~al.(2024)Klein, Lüth, Schlegel, Bungert, El-Assady, and Jäger]{klein2024navigating}
Lukas Klein, Carsten~T. Lüth, Udo Schlegel, Till~J. Bungert, Mennatallah El-Assady, and Paul~F. Jäger.
\newblock Navigating the maze of explainable {AI}: {A} systematic approach to evaluating methods and metrics.
\newblock In \emph{NeurIPS}, 2024.

\bibitem[Koh \& Liang(2017)Koh and Liang]{koh2017understanding}
Pang~Wei Koh and Percy Liang.
\newblock Understanding black-box predictions via influence functions.
\newblock In \emph{ICML}, 2017.

\bibitem[Kokhlikyan et~al.(2020)Kokhlikyan, Miglani, Martin, Wang, Alsallakh, Reynolds, Melnikov, Kliushkina, Araya, Yan, and Reblitz-Richardson]{kokhlikyan2020captum}
Narine Kokhlikyan, Vivek Miglani, Miguel Martin, Edward Wang, Bilal Alsallakh, Jonathan Reynolds, Alexander Melnikov, Natalia Kliushkina, Carlos Araya, Siqi Yan, and Orion Reblitz-Richardson.
\newblock Captum: A unified and generic model interpretability library for {PyTorch}.
\newblock \emph{arXiv preprint arXiv:2009.07896}, 2020.

\bibitem[Kolpaczki et~al.(2024)Kolpaczki, Bengs, Muschalik, and Hüllermeier]{kolpaczki2024approximating}
Patrick Kolpaczki, Viktor Bengs, Maximilian Muschalik, and Eyke Hüllermeier.
\newblock Approximating the {S}hapley value without marginal contributions.
\newblock In \emph{AAAI}, 2024.

\bibitem[Krzyziński et~al.(2023)Krzyziński, Spytek, Baniecki, and Biecek]{krzyzinski2023survshap}
Mateusz Krzyziński, Mikołaj Spytek, Hubert Baniecki, and Przemysław Biecek.
\newblock {SurvSHAP(t)}: Time-dependent explanations of machine learning survival models.
\newblock \emph{Knowledge-Based Systems}, 262:\penalty0 110234, 2023.

\bibitem[Laberge et~al.(2023)Laberge, Aivodji, Hara, and Mario~Marchand]{laberge2023fooling}
Gabriel Laberge, Ulrich Aivodji, Satoshi Hara, and Foutse~Khomh Mario~Marchand.
\newblock {Fooling SHAP with Stealthily Biased Sampling}.
\newblock In \emph{ICLR}, 2023.

\bibitem[Li et~al.(2021)Li, Nagarajan, Plumb, and Talwalkar]{li2021learning}
Jeffrey Li, Vaishnavh Nagarajan, Gregory Plumb, and Ameet Talwalkar.
\newblock A learning theoretic perspective on local explainability.
\newblock In \emph{ICLR}, 2021.

\bibitem[Li et~al.(2023)Li, Xiong, Li, Zhang, Liu, Jiang, Chen, and Dou]{li2023glime}
Xuhong Li, Haoyi Xiong, Xingjian Li, Xiao Zhang, Ji~Liu, Haiyan Jiang, Zeyu Chen, and Dejing Dou.
\newblock {G-LIME}: Statistical learning for local interpretations of deep neural networks using global priors.
\newblock \emph{Artificial Intelligence}, 314:\penalty0 103823, 2023.

\bibitem[Lin et~al.(2023)Lin, Covert, and Lee]{lin2023robustness}
Chris Lin, Ian Covert, and Su-In Lee.
\newblock On the robustness of removal-based feature attributions.
\newblock In \emph{NeurIPS}, 2023.

\bibitem[Lundberg \& Lee(2017)Lundberg and Lee]{lundberg2017unified}
Scott~M Lundberg and Su-In Lee.
\newblock A unified approach to interpreting model predictions.
\newblock In \emph{NeurIPS}, 2017.

\bibitem[Lundstrom et~al.(2022)Lundstrom, Huang, and Razaviyayn]{lundstrom2022rigorous}
Daniel~D Lundstrom, Tianjian Huang, and Meisam Razaviyayn.
\newblock A rigorous study of integrated gradients method and extensions to internal neuron attributions.
\newblock In \emph{ICML}, 2022.

\bibitem[Moosbauer et~al.(2021)Moosbauer, Herbinger, Casalicchio, Lindauer, and Bischl]{moosbauer2021explaining}
Julia Moosbauer, Julia Herbinger, Giuseppe Casalicchio, Marius Lindauer, and Bernd Bischl.
\newblock Explaining hyperparameter optimization via partial dependence plots.
\newblock In \emph{NeurIPS}, 2021.

\bibitem[Muandet et~al.(2017)Muandet, Fukumizu, Sriperumbudur, and Schölkopf]{muandet2017kernel}
Krikamol Muandet, Kenji Fukumizu, Bharath Sriperumbudur, and Bernhard Schölkopf.
\newblock Kernel mean embedding of distributions: A review and beyond.
\newblock \emph{Foundations and Trends in Machine Learning}, 10\penalty0 (1--2):\penalty0 1--141, 2017.

\bibitem[Muschalik et~al.(2024)Muschalik, Fumagalli, Hammer, and Hüllermeier]{muschalik2024beyond}
Maximilian Muschalik, Fabian Fumagalli, Barbara Hammer, and Eyke Hüllermeier.
\newblock Beyond {TreeSHAP}: Efficient computation of any-order {Shapley} interactions for tree ensembles.
\newblock In \emph{AAAI}, 2024.

\bibitem[Olsen et~al.(2022)Olsen, Glad, Jullum, and Aas]{olsen2022using}
Lars H.~B. Olsen, Ingrid~K. Glad, Martin Jullum, and Kjersti Aas.
\newblock Using {Shapley} values and variational autoencoders to explain predictive models with dependent mixed features.
\newblock \emph{Journal of Machine Learning Research}, 23\penalty0 (213):\penalty0 1--51, 2022.

\bibitem[Olsen et~al.(2024)Olsen, Glad, Jullum, and Aas]{olsen2024comparative}
Lars Henry~Berge Olsen, Ingrid~Kristine Glad, Martin Jullum, and Kjersti Aas.
\newblock A comparative study of methods for estimating model-agnostic {Shapley} value explanations.
\newblock \emph{Data Mining and Knowledge Discovery}, pp.\  1--48, 2024.

\bibitem[Owen(2017)]{owen2017statistically}
Art~B Owen.
\newblock Statistically efficient thinning of a markov chain sampler.
\newblock \emph{Journal of Computational and Graphical Statistics}, 26\penalty0 (3):\penalty0 738--744, 2017.

\bibitem[Petsiuk et~al.(2018)Petsiuk, Das, and Saenko]{petsiuk2018rise}
Vitali Petsiuk, Abir Das, and Kate Saenko.
\newblock {RISE}: Randomized input sampling for explanation of black-box models.
\newblock In \emph{BMVC}, 2018.

\bibitem[Ribeiro et~al.(2016)Ribeiro, Singh, and Guestrin]{ribeiro2016should}
Marco~Tulio Ribeiro, Sameer Singh, and Carlos Guestrin.
\newblock {``Why should I trust you?'' Explaining the predictions of any classifier}.
\newblock In \emph{KDD}, 2016.

\bibitem[Rong et~al.(2024)Rong, Leemann, Nguyen, Fiedler, Qian, Unhelkar, Seidel, Kasneci, and Kasneci]{rong2024towards}
Yao Rong, Tobias Leemann, Thai-Trang Nguyen, Lisa Fiedler, Peizhu Qian, Vaibhav Unhelkar, Tina Seidel, Gjergji Kasneci, and Enkelejda Kasneci.
\newblock Towards human-centered explainable {AI}: A survey of user studies for model explanations.
\newblock \emph{IEEE Transactions on Pattern Analysis and Machine Intelligence}, 46\penalty0 (4):\penalty0 2104--2122, 2024.

\bibitem[Scholbeck et~al.(2020)Scholbeck, Molnar, Heumann, Bischl, and Casalicchio]{scholbeck2020sampling}
Christian~A Scholbeck, Christoph Molnar, Christian Heumann, Bernd Bischl, and Giuseppe Casalicchio.
\newblock Sampling, intervention, prediction, aggregation: a generalized framework for model-agnostic interpretations.
\newblock In \emph{ECML PKDD}, 2020.

\bibitem[Sener \& Savarese(2018)Sener and Savarese]{sener2018active}
Ozan Sener and Silvio Savarese.
\newblock Active learning for convolutional neural networks: A core-set approach.
\newblock In \emph{ICLR}, 2018.

\bibitem[Shetty et~al.(2022)Shetty, Dwivedi, and Mackey]{shetty2022distribution}
Abhishek Shetty, Raaz Dwivedi, and Lester Mackey.
\newblock Distribution compression in near-linear time.
\newblock In \emph{ICLR}, 2022.

\bibitem[Slack et~al.(2020)Slack, Hilgard, Jia, Singh, and Lakkaraju]{slack2020fooling}
Dylan Slack, Sophie Hilgard, Emily Jia, Sameer Singh, and Himabindu Lakkaraju.
\newblock Fooling {LIME} and {SHAP}: Adversarial attacks on post hoc explanation methods.
\newblock In \emph{AIES}, 2020.

\bibitem[Slack et~al.(2021)Slack, Hilgard, Singh, and Lakkaraju]{slack2021reliable}
Dylan Slack, Anna Hilgard, Sameer Singh, and Himabindu Lakkaraju.
\newblock Reliable post hoc explanations: Modeling uncertainty in explainability.
\newblock In \emph{NeurIPS}, 2021.

\bibitem[Sriperumbudur et~al.(2010)Sriperumbudur, Gretton, Fukumizu, Sch{{\"o}}lkopf, and Lanckriet]{sriperumbudur2010hilbert}
Bharath~K. Sriperumbudur, Arthur Gretton, Kenji Fukumizu, Bernhard Sch{{\"o}}lkopf, and Gert~R.G. Lanckriet.
\newblock Hilbert space embeddings and metrics on probability measures.
\newblock \emph{Journal of Machine Learning Research}, 11\penalty0 (50):\penalty0 1517--1561, 2010.

\bibitem[Sriperumbudur et~al.(2012)Sriperumbudur, Fukumizu, Gretton, Sch{\"o}lkopf, and Lanckriet]{sriperumbudur2012empirical}
Bharath~K Sriperumbudur, Kenji Fukumizu, Arthur Gretton, Bernhard Sch{\"o}lkopf, and Gert~RG Lanckriet.
\newblock On the empirical estimation of integral probability metrics.
\newblock \emph{Electronic Journal of Statistics}, 6:\penalty0 1550--1599, 2012.

\bibitem[Strumbelj \& Kononenko(2010)Strumbelj and Kononenko]{strumbelj2010efficient}
Erik Strumbelj and Igor Kononenko.
\newblock An efficient explanation of individual classifications using game theory.
\newblock \emph{Journal of Machine Learning Research}, 11:\penalty0 1--18, 2010.

\bibitem[Van~Looveren \& Klaise(2021)Van~Looveren and Klaise]{van2021interpretable}
Arnaud Van~Looveren and Janis Klaise.
\newblock Interpretable counterfactual explanations guided by prototypes.
\newblock In \emph{ECML PKDD}, 2021.

\bibitem[Wang et~al.(2018)Wang, Zhu, Torralba, and Efros]{wang2018dataset}
Tongzhou Wang, Jun-Yan Zhu, Antonio Torralba, and Alexei~A. Efros.
\newblock Dataset distillation.
\newblock \emph{arXiv preprint arXiv:1811.10959}, 2018.

\bibitem[Zhang et~al.(2024)Zhang, Zheng, Zhou, and Lu]{zhang2024path}
Borui Zhang, Wenzhao Zheng, Jie Zhou, and Jiwen Lu.
\newblock Path choice matters for clear attributions in path methods.
\newblock In \emph{ICLR}, 2024.

\bibitem[Zhao et~al.(2021)Zhao, Mopuri, and Bilen]{zhao2021dataset}
Bo~Zhao, Konda~Reddy Mopuri, and Hakan Bilen.
\newblock Dataset condensation with gradient matching.
\newblock In \emph{ICLR}, 2021.

\bibitem[Zimmerman et~al.(2024)Zimmerman, Giusti, and Guzzi]{zimmerman2024resourceaware}
Nicky Zimmerman, Alessandro Giusti, and Jérôme Guzzi.
\newblock Resource-aware collaborative monte carlo localization with distribution compression.
\newblock \emph{arXiv preprint arXiv:2404.02010}, 2024.

\end{thebibliography}
\bibliographystyle{iclr2025}

\clearpage
\appendix
\section*{Appendix for ``Efficient and Accurate Explanation Estimation with Distribution Compression''}

In Appendix~\ref{app:proofs}, we derive proofs for Propositions~\ref{prop:bound-removal}~\&~\ref{prop:bound-global}. Appendix~\ref{app:code} provides code listings for \shap, \expectedgradients and \featureeffects, analogous to Listing~\ref{listing:sage} for \sage. Additional details on the experimental setup are provided in Appendix~\ref{app:experiments-setup}. Appendices~\ref{app:experiments-openxai},~\ref{app:experiments-expected-gradients}~\&~\ref{app:experiments-openml} report experimental results for the remaining datasets. Appendix~\ref{app:setup-compute-resources} comments on compute resources used for experiments. Appendix~\ref{app:visual-comparison} provides exemplary visual comparisons of explanations. The code to reproduce all experiments in this paper is available at \url{https://github.com/hbaniecki/compress-then-explain}.

\startcontents[sections]
\printcontents[sections]{l}{1}{\setcounter{tocdepth}{2}}

\clearpage
\section{Motivation: standard \iid sampling in explanation estimation}\label{app:motivation-quotes}

We find that \iid sampling from datasets is a heuristic often used (and overlooked) in various estimators of post-hoc explanations. Our work aims to first quantify the approximation error introduced by \emph{sample then explain}, and then propose a method to efficiently reduce it. Below are a few examples from the literature on explainability that motivate the shift to our introduced \emph{compress then explain} paradigm.

In \citep{laberge2023fooling}, we read ``\emph{For instance, when a dataset is used to represent a background distribution, explainers in the SHAP library such as the ExactExplainer and TreeExplainer will subsample this dataset by selecting 100 instances uniformly at random when the size of the dataset exceeds 100.}''

In \citep{chen2023extracting}, we read ``\emph{[Footnote 1.] We use a random subset of samples for each class in the real implementation, to reduce the computation costs of clustering.}''

In \citep{ghalebikesabi2021locality}, we read ``\emph{After training a convolutional neural network on the MNIST dataset, we explain digits with the predicted label 8 given a background dataset of 100 images with labels 3 and 8.}'', as well as ``\emph{Feature attributions are sorted by similarity according to a preliminary PCA analysis across a subset of 2000 samples from the Adult Income dataset, using 2000 reference points.}''

In \citep{erion2021improving}, we read ``\emph{During training, we let k be the number of samples we draw to compute expected gradients for each mini-batch.} and ``\emph{This expectation-based formulation lends itself to a natural, sampling based approximation method: (1) draw samples of $x'$ from the training dataset [...], (2) compute the value inside the expectation for each sample and (3) average over samples.}''

In \citep{van2021interpretable}, we read ``\emph{We also need a representative, unlabeled sample of the training dataset.}'', and in Algorithms 1 and 2: ``\emph{A sample
$X = \{x_1, \ldots , x_n\}$ from training set.}''

In \citep{covert2020understanding}, we read ``\emph{When calculating feature importance, our sampling approximation for SAGE (Algorithm 1) was run using draws from the marginal distribution. We used a fixed set of 512 background samples for the bank, bike and credit datasets, 128 for MNIST, and all 334 training examples for BRCA.}''

In the \texttt{shap} Python package \citep{lundberg2017unified}, there is a warning saying ``\emph{Using 110 background data samples could cause slower run times. Consider using shap.sample(data, K) or shap.kmeans(data, K) to summarize the background as K samples.}'', and the documentation mentions ``\emph{For small problems, this background dataset can be the whole training set, but for larger problems consider using a single reference value or using the \textbf{kmeans} function to summarize the dataset.}''

\clearpage
\section{Proofs}\label{app:proofs}
\setcounter{proposition}{0}

Below, we derive proofs for Propositions~\ref{prop:bound-removal}~\&~\ref{prop:bound-global}.

\begin{proposition}[Feature marginalization is bounded by the maximum mean discrepancy between data samples]
For two empirical distributions $q_{\bbX}, q_{\bbXtilde}$ approximated with a kernel density estimator $\bk$, we have $\big|f(\mathbf{x}_s; q_{\bbX}) - f(\mathbf{x}_s; q_{\bbXtilde})\big| \leq C_f \cdot \widehat{\mathrm{MMD}}_{\bk}(q_{\bbX}, q_{\bbXtilde})$, where $C_f$ denotes a constant that bounds the model function $f$, i.e. $\forall_{\bx \in \bbR^p} \big|f(\bx)\big| \leq C_f$.
\begin{proof}
We derive the following inequality: 
\begin{align}
    \big( f(\bxs; \pXsb) - f(\bxs; \qXsb) \big)^2 & = \left( \bbE_{\bXsb \sim p_{\bXsb}} \left[f(\bxs, \bXsb)\right] - \bbE_{\bXsb \sim q_{\bXsb}} \left[f(\bxs, \bXsb)\right] \right)^2 \\
    & = \left( \int f(\bxs, \bxsb) \pXsb(\bxsb) d\bxsb - \int f(\bxs, \bxsb) \qXsb(\bxsb) d\bxsb \right)^2 \\
    \text{(linearity)} & = \left( \int f(\bxs, \bxsb) \big(\pXsb(\bxsb) - \qXsb(\bxsb)\big) d\bxsb \right)^2 \\
    \text{(Cauchy--Schwarz)} & \leq \int \big( f(\bxs, \bxsb) \big)^2 d\bxsb \int \big(\pXsb(\bxsb) - \qXsb(\bxsb)\big)^2  d\bxsb \\
    \text{(boundedness)} & \leq C_f^2 \cdot \int \big(\pXsb(\bxsb) - \qXsb(\bxsb)\big)^2  d\bxsb \\
    \text{(Definition~\ref{def:mmdhat})} & = C_f^2 \cdot \widehat{\dmmd}^{2}_{\bk}(\pXsb, \qXsb).
\end{align}

Substituting with empirical distributions, we have
\begin{equation}
\big|f(\mathbf{x}_s; q_{\bbX}) - f(\mathbf{x}_s; q_{\bbXtilde})\big| \leq C_f \cdot \widehat{\mathrm{MMD}}_{\bk}(q_{\bbX}, q_{\bbXtilde}).
\end{equation}
\end{proof}
\end{proposition}

\begin{proposition}[Global explanation is bounded by the maximum mean discrepancy between data samples]
For two empirical distributions $q_{\bbX}, q_{\bbXtilde}$ approximated with a kernel density estimator $\bk$, we have $\normbig{G(q_{\bbX}; f, g) - G(q_{\bbXtilde}; f, g)}_2 \leq C_g\cdot\widehat{\dmmd}_{\bk}(q_{\bbX}, q_{\bbXtilde})$, where $C_g$ denotes a constant that bounds the local explanation function $g$, i.e. $\forall_{\bx \in \bbR^p} \normbig{g(\bx; \cdot)}_2 \leq C_g$.
\begin{proof}
We derive the following inequality: 
\begin{align}
    \normbig{G(p_{\bX}; f, g) - G(q_{\bX}; f, g)}_2^2 & = \normbig{\bbE_{\bX \sim p_{\bX}} \left[g(\bX; f, \cdot)\right] - \bbE_{\bX \sim q_{\bX}} \left[g(\bX; f, \cdot)\right]}_2^2 \\
    & = \norm{\int g(\bx) p_{\bX}(\bx) d\bx - \int g(\bx) q_{\bX}(\bx) d\bx}_2^2 \\
    \text{(linearity)} & = \norm{\int g(\bx) \big(p_{\bX}(\bx) -  q_{\bX}(\bx)\big) d\bx}_2^2 \\
    \text{(Cauchy--Schwarz)} & \leq \int \normbig{g(\bx)}_2^2 d\bx \int \big(p_{\bX}(\bx) -  q_{\bX}(\bx)\big)^2 d\bx \\
    \text{(boundedness)} & \leq C_g^2 \cdot \int \big( p_{\bX}(\bx) -  q_{\bX}(\bx) \big)^2 d\bx \\
    \text{(Definition~\ref{def:mmdhat})} & = C_g^2 \cdot \widehat{\dmmd}^{2}_{\bk}(p_{\bX}, q_{\bX}).
\end{align}

Substituting with empirical distributions, we have
\begin{equation}
\normbig{G(q_{\bbX}; f, g) - G(q_{\bbXtilde}; f, g)}_2 \leq C_g\cdot\widehat{\dmmd}_{\bk}(q_{\bbX}, q_{\bbXtilde}).
\end{equation}
\end{proof}
\end{proposition}

\clearpage
\section{Implementing \ctetitle in practice}\label{app:code}

\cte is simple to plug-into the current workflows for explanation estimation as shown in Listing~\ref{listing:sage} for \sage, Listing~\ref{listing:shap} for \shap, Listing~\ref{listing:expected-gradients} for \expectedgradients, and Listing~\ref{listing:feature-effects} for \featureeffects. We use the \texttt{goodpoints} Python package~\citep[][MIT license]{dwivedi2021kernel}.

\begin{listing}[ht]
\small
\begin{boxminted}{2-4}{Python}
X, model = ...
from goodpoints import compress
ids = compress.compresspp_kt(X, kernel_type=b"gaussian", g=4)
X_compressed = X[ids]
import shap
masker = shap.maskers.Independent(X_compressed)
explainer = shap.PermutationExplainer(model.predict, masker)
explanation = explainer(X)
\end{boxminted}
\caption{Code snippet showing the 3-line plug-in of distribution compression for \shap estimation.}\label{listing:shap}
\end{listing}

\begin{listing}[ht]
\small
\begin{boxminted}{2-4}{Python}
X, model = ...
from goodpoints import compress
ids = compress.compresspp_kt(X, kernel_type=b"gaussian", g=4)
X_compressed = X[ids]
import captum
explainer = captum.attr.IntegratedGradients(model)
import torch
inputs = torch.as_tensor(X)
baselines = torch.as_tensor(X_compressed)
explanation = torch.mean(torch.stack([
    explainer.attribute(inputs, baselines[[i]], target=1) 
    for i in range(baselines.shape[0])
]), dim=0)
\end{boxminted}
\caption{Code snippet showing the plug-in of distribution compression for \expectedgradients.}\label{listing:expected-gradients}
\end{listing}

\begin{listing}[ht]
\small
\begin{boxminted}{2-4}{Python}
X, model = ...
from goodpoints import compress
ids = compress.compresspp_kt(X, kernel_type=b"gaussian", g=4)
X_compressed = X[ids]
import alibi
explainer = alibi.explainers.PartialDependence(predictor=model.predict)
explanation = explainer.explain(X_compressed)
\end{boxminted}
\caption{Code snippet showing the plug-in of distribution compression for \featureeffects.}\label{listing:feature-effects}
\end{listing}

\clearpage
\section{Experimental setup}\label{app:experiments-setup}

\subsection{Explanation hyperparameters}\label{app:setup-explanation-hyperparameters}

In Section~\ref{sec:experiments}, we experiment with 4 explanation methods (6 estimators). Without the loss of generality, in case of classification models, we always explain a prediction for the 2nd class. For \shap, we use the \kernshap and \permshap implementations from the \texttt{shap} Python package~\citep[][MIT license]{lundberg2017unified} with default hyperparameters (notably, \texttt{npermutations=10} in the latter). For \sage, we use the \kernsage and \permsage implementations from the \texttt{sage} Python package~\citep[][MIT license]{covert2020understanding}. We use default hyperparameters; notably, a cross-entropy loss for classification and mean squared error for regression. For \expectedgradients, we aggregate with mean the integrated gradients explanations from the \texttt{captum} Python package~\citep[][BSD-3 license]{kokhlikyan2020captum}, for which we use default hyperparameters; notably, \texttt{n\_steps=50} and \texttt{method="gausslegendre"}. For \featureeffects, we implement the partial dependence algorithm~\citep{apley2020visualizing,moosbauer2021explaining} ourselves for maximum computational speed in case of 2-dimensional plots, mimicking the popular open-source implementations.\footnote{\url{https://docs.seldon.io/projects/alibi/en/latest/api/alibi.explainers.html\#alibi.explainers.PartialDependence}; \url{https://interpret.ml/docs/python/api/PartialDependence}} We use 100 uniformly distributed grid points for 1-dimensional plots and 10$\times$10 uniformly distributed grid points for 2-dimensional plots.

\subsection{Details on datasets and models}\label{app:setup-datasets-models}

Table~\ref{tab:openxai-datasets} shows details of datasets from the OpenXAI~\citep[][MIT license]{agarwal2022openxai} benchmark used in Sections~\ref{sec:experiments-accuracy},~\ref{sec:experiments-efficient}~\&~\ref{sec:experiments-sanity-check}. To each dataset, there is a pretrained neural network with an accuracy of 92\%~(\texttt{gaussian}), 85\%~(\texttt{compas}), 74\%~(\texttt{heloc}), 85\%~(\texttt{adult}) and 93\%~(\texttt{gmsc}). We do not further preprocess data; notably, feature values are already scaled to~$[0, 1]$.

\begin{table}[h]
\centering
\caption{Datasets from OpenXAI with $\nvalid > 1000$ used in experiments.}
\label{tab:openxai-datasets}
\begin{tabular}{l|rrrr}
\toprule
\textbf{Dataset} & $\ntrain$ & $\nvalid$ & $d$ & \textbf{No. classes} \\
\midrule
    \texttt{gaussian} & 3750 & 1250 & 20 & 2 \\
    \texttt{compas} & 4937 & 1235 & 7 & 2 \\
    \texttt{heloc} & 7896 & 1975 & 23 & 2 \\
    \texttt{adult} & 36177 & 9045 & 13 & 2 \\
    \texttt{gmsc} & 81767 & 20442 & 10 & 2 \\
\bottomrule
\end{tabular}
\end{table}

\begin{table}
\centering
\caption{Datasets from OpenML-CC18 and OpenML-CTR23 with $\nvalid > 1000$ used in experiments.}
\label{tab:openml-datasets}
\begin{tabular}{ll|rrrr}
\toprule
\textbf{Dataset} & \textbf{Task ID} & $\ntrain$ & $\nvalid$ & $d$ & \textbf{No. classes} \\
\midrule
        phoneme & 9952 & 4053 & 1351 & 5 & 2 \\
    wilt & 146820 & 3629 & 1210 & 5 & 2 \\
    cps88wages & 361261 & 21116 & 7039 & 6 & -- \\
    jungle\_chess & 167119 & 33614 & 11205 & 6 & 3 \\
    abalone & 361234 & 3132 & 1045 & 8 & -- \\
    electricity & 219 & 33984 & 11328 & 8 & 2 \\
    kin8nm & 361258 & 6144 & 2048 & 8 & -- \\
    california\_housing & 361255 & 15480 & 5160 & 8 & -- \\
    brazilian\_houses & 361267 & 8019 & 2673 & 9 & -- \\
    diamonds & 361257 & 40455 & 13485 & 9 & -- \\
    physiochemical\_protein & 361241 & 34297 & 11433 & 9 & -- \\
    white\_wine & 361249 & 3673 & 1225 & 11 & -- \\
    health\_insurance & 361269 & 16704 & 5568 & 11 & -- \\
    grid\_stability & 361251 & 7500 & 2500 & 12 & -- \\
    adult & 7592 & 36631 & 12211 & 14 & 2 \\
    naval\_propulsion\_plant & 361247 & 8950 & 2984 & 14 & -- \\
    miami\_housing & 361260 & 10449 & 3483 & 15 & -- \\
    letter & 6 & 15000 & 5000 & 16 & 26 \\
    bank-marketing & 14965 & 33908 & 11303 & 16 & 2 \\
    pendigits & 32 & 8244 & 2748 & 16 & 10 \\
    video\_transcoding & 361252 & 51588 & 17196 & 18 & -- \\
    churn & 167141 & 3750 & 1250 & 20 & 2 \\
    kings\_county & 361266 & 16209 & 5404 & 21 & -- \\
    numerai28.6 & 167120 & 72240 & 24080 & 21 & 2 \\
    sarcos & 361254 & 36699 & 12234 & 21 & -- \\
    cpu\_activity & 361256 & 6144 & 2048 & 21 & -- \\
    jm1 & 3904 & 8163 & 2722 & 21 & 2 \\
    wall-robot-navigation & 9960 & 4092 & 1364 & 24 & 4 \\
    fifa & 361272 & 14383 & 4795 & 28 & -- \\
    PhishingWebsites & 14952 & 8291 & 2764 & 30 & 2 \\
    \midrule
    pumadyn32nh & 361259 & 6144 & 2048 & 32 & -- \\
    GestureSegmentation & 14969 & 7404 & 2469 & 32 & 5 \\
    satimage & 2074 & 4822 & 1608 & 36 & 6 \\
    texture & 125922 & 4125 & 1375 & 40 & 11 \\
    connect-4 & 146195 & 50667 & 16890 & 42 & 3 \\
    fps\_benchmark & 361268 & 18468 & 6156 & 43 & -- \\
    wave\_energy & 361253 & 54000 & 18000 & 48 & -- \\
    theorem-proving & 9985 & 4588 & 1530 & 51 & 6 \\
    spambase & 43 & 3450 & 1151 & 57 & 2 \\
    optdigits & 28 & 4215 & 1405 & 64 & 10 \\
    superconductivity & 361242 & 15947 & 5316 & 81 & -- \\
    nomao & 9977 & 25848 & 8617 & 118 & 2 \\
    har & 14970 & 7724 & 2575 & 561 & 6 \\
    isolet & 3481 & 5847 & 1950 & 617 & 26 \\
    mnist\_784 & 3573 & 52500 & 17500 & 784 & 10 \\
    Fashion-MNIST & 146825 & 52500 & 17500 & 784 & 10 \\
    Devnagari-Script & 167121 & 69000 & 23000 & 1024 & 46 \\
    CIFAR\_10 & 167124 & 45000 & 15000 & 3072 & 10 \\
\bottomrule
\end{tabular}
\end{table}

Table~\ref{tab:openml-datasets} shows details of datasets from the OpenML-CC18~\citep[][BSD-3 license]{bischl2021openml} and OpenML-CTR23~\citep[][BSD-3 license]{fischer2023openmlctr} benchmarks used in Sections~\ref{sec:experiments-expected-gradients}~\&~\ref{sec:experiments-openml}. We first split all datasets in 75:25 (train:validation) ratio and left 48 datasets with $\nvalid > 1000$ for our experiments. For the 30 smaller ($d < 32$) datasets, we train an XGBoost model with default hyperparameters (200 estimators) and explain it with \shap, \sage, \featureeffects. For the 18 bigger ($d \geq 32$) datasets, we train a 3-layer neural network model with (128, 64) neurons in hidden ReLU layers and explain it with \expectedgradients. We perform basic preprocessing of data: (1) remove features with a single or $n$ unique values, (2) target encode categorical features, (3) impute missing values with mean, and (4) standardize features. 

In general, categorical features can be an issue for clustering and distribution compression algorithms; so are for many explanation algorithms and conditional distribution samplers. Although target encoding worked well in our setup, we envision two additional heuristics to deal with categorical features: (1) perform distribution compression using a dataset restricted to non-categorical features, (2) target encode categorical features only for distribution compression.

\section{\ctetitle significantly improves the estimation of feature attributions \& importance}\label{app:experiments-openxai}

\begin{figure}[h]
    \centering
    \includegraphics[width=0.49\textwidth]{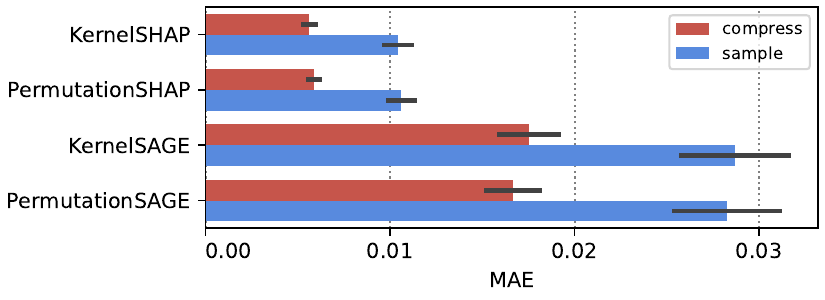}
    \includegraphics[width=0.49\textwidth]{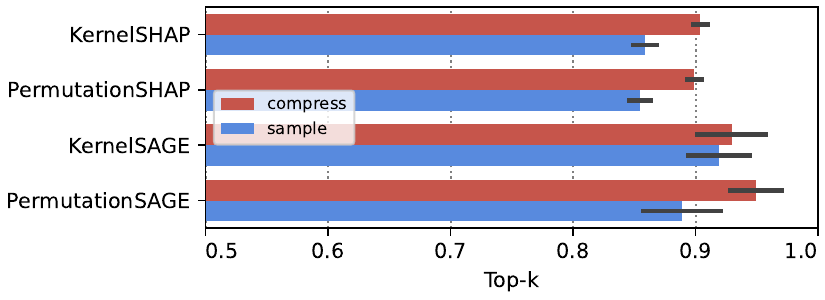}
    \includegraphics[width=0.49\textwidth]{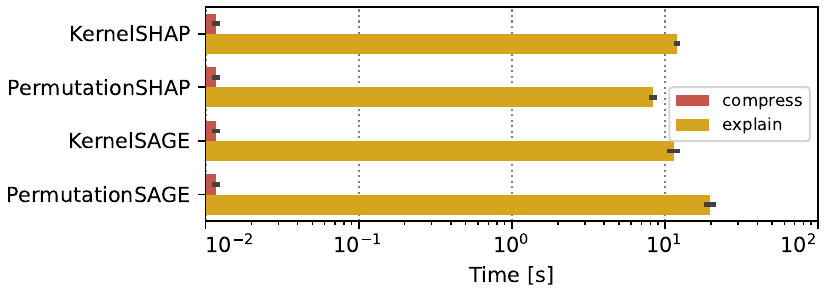}
    \includegraphics[width=0.49\textwidth]{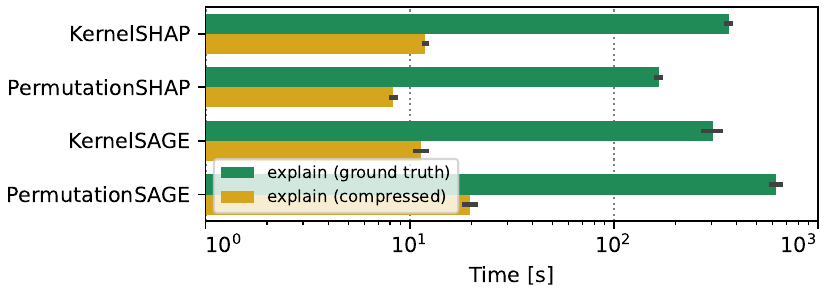}
    \caption{Extended Figure~\ref{fig:adult-shap-sage} (1/4). \cte improves \shap and \sage estimation by using the compressed samples as background data for the \texttt{compas} dataset. 
    We measure mean absolute error (MAE~$\downarrow$) between feature attribution and importance values, as well as the precision in correctly indicating the 3 most important features (Top-k~$\uparrow$). Computational resources required to compress a distribution are negligible in the context of explanation estimation. (mean $\pm$ se.)
    }
    \label{fig:compas-shap-sage}
    \vspace{1em}

    \centering
    \includegraphics[width=0.49\textwidth]{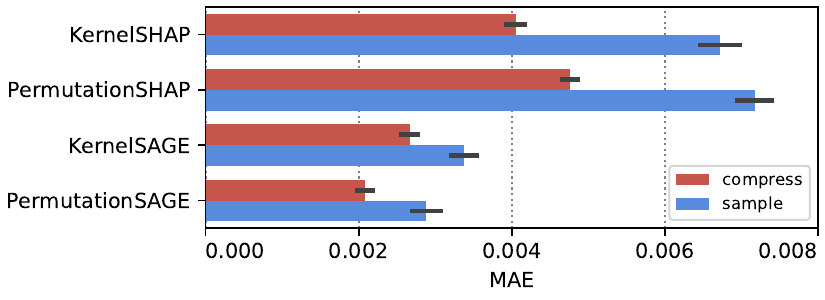}
    \includegraphics[width=0.49\textwidth]{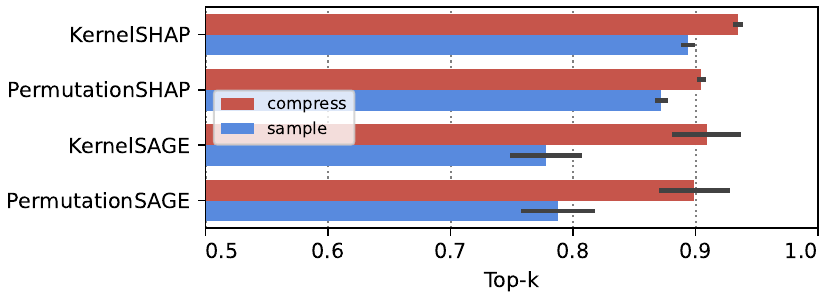}
    \includegraphics[width=0.49\textwidth]{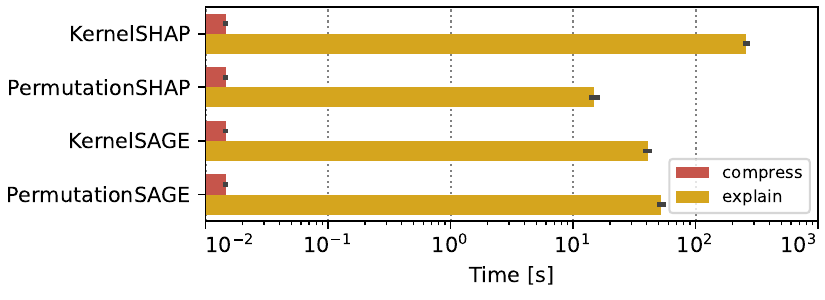}
    \includegraphics[width=0.49\textwidth]{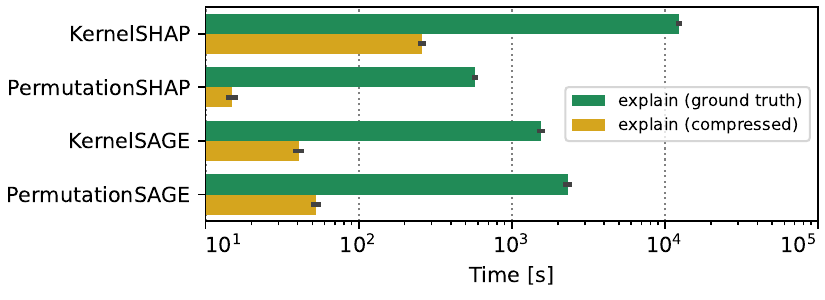}
    \caption{Extended Figure~\ref{fig:adult-shap-sage} (2/4). \cte improves \shap and \sage estimation on the \texttt{heloc} dataset. 
    }
    \label{fig:heloc-shap-sage}
\end{figure}

\begin{figure}
    \centering
    \includegraphics[width=0.49\textwidth]{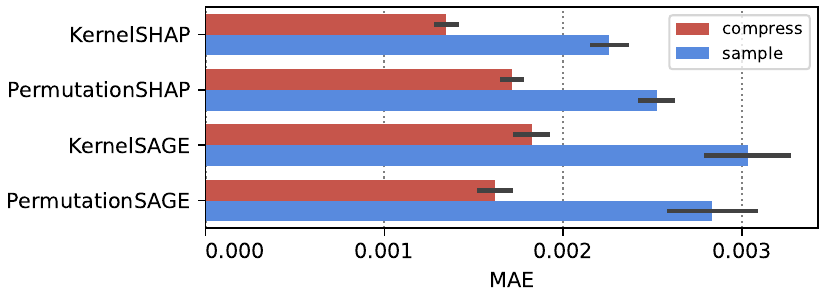}
    \includegraphics[width=0.49\textwidth]{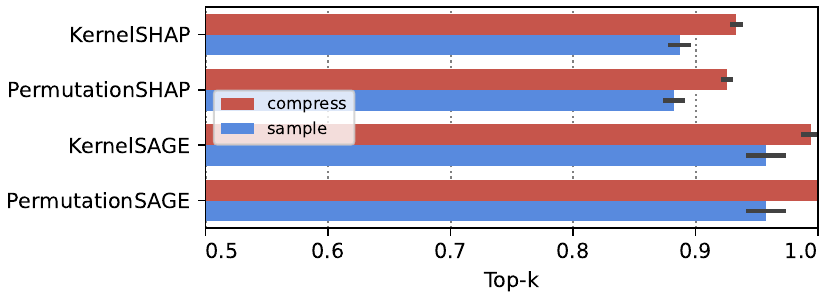}
    \includegraphics[width=0.49\textwidth]{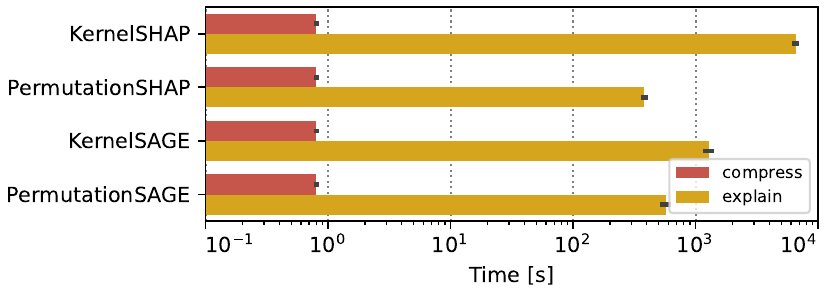}
    \includegraphics[width=0.49\textwidth]{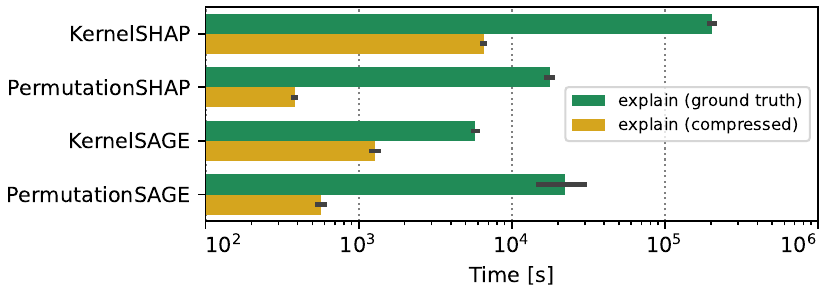}
    \caption{Extended Figure~\ref{fig:adult-shap-sage} (3/4). \cte improves \shap and \sage estimation by using the compressed samples as background data for the \texttt{gmsc} dataset. We measure mean absolute error (MAE~$\downarrow$) between feature attribution and importance values, as well as the precision in correctly indicating the 5 most important features (Top-k~$\uparrow$). Computational resources required to compress a distribution are negligible in the context of explanation estimation. (mean $\pm$ se.)}
    \label{fig:gmsc-shap-sage}
    \vspace{1em}
    
    \includegraphics[width=0.49\textwidth]{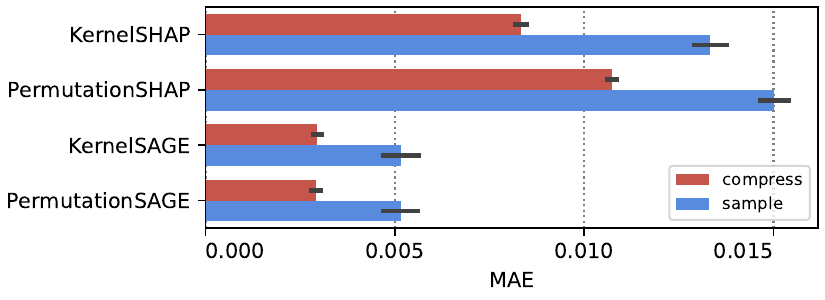}
    \includegraphics[width=0.49\textwidth]{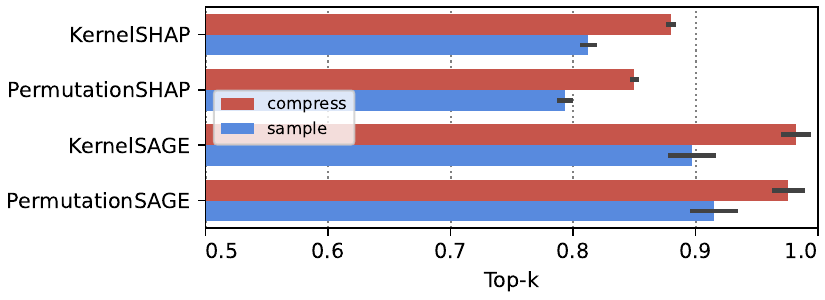}
    \includegraphics[width=0.49\textwidth]{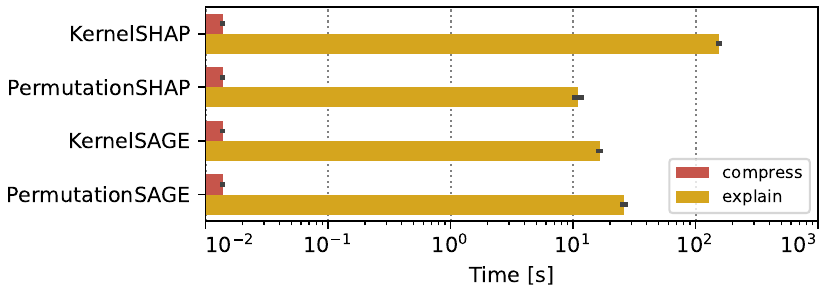}
    \includegraphics[width=0.49\textwidth]{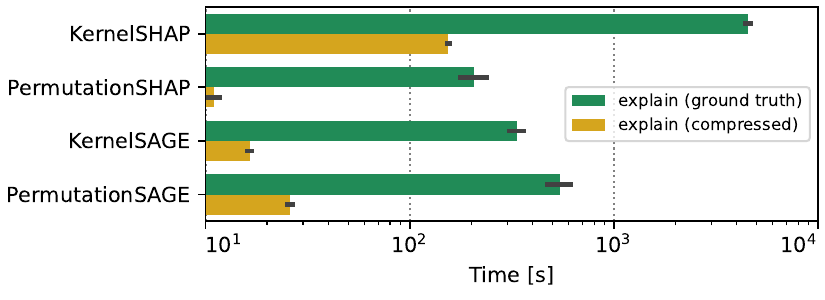}
    \caption{Extended Figure~\ref{fig:adult-shap-sage} (4/4). \cte improves \shap and \sage estimation on the \texttt{gaussian} dataset. 
    }
    \label{fig:gaussian-shap-sage}
\end{figure}

We report the differences in MAE and Top-k between \cte and \iid sampling in 
Figure~\ref{fig:compas-shap-sage}~(\texttt{compas}), 
Figure~\ref{fig:heloc-shap-sage}~(\texttt{heloc}), Figure~\ref{fig:gmsc-shap-sage} (\texttt{gmsc}) and Figure~\ref{fig:gaussian-shap-sage} (\texttt{gaussian}). On all the considered tasks, \cte offers a notable decrease in approximation error of \shap and \sage with negligible computational overhead (as measured by time in seconds).

\clearpage
\section{\ctetitle improves gradient-based explanations}\label{app:experiments-expected-gradients}

\begin{figure}[h]
    \centering
    \includegraphics[width=0.99\textwidth]{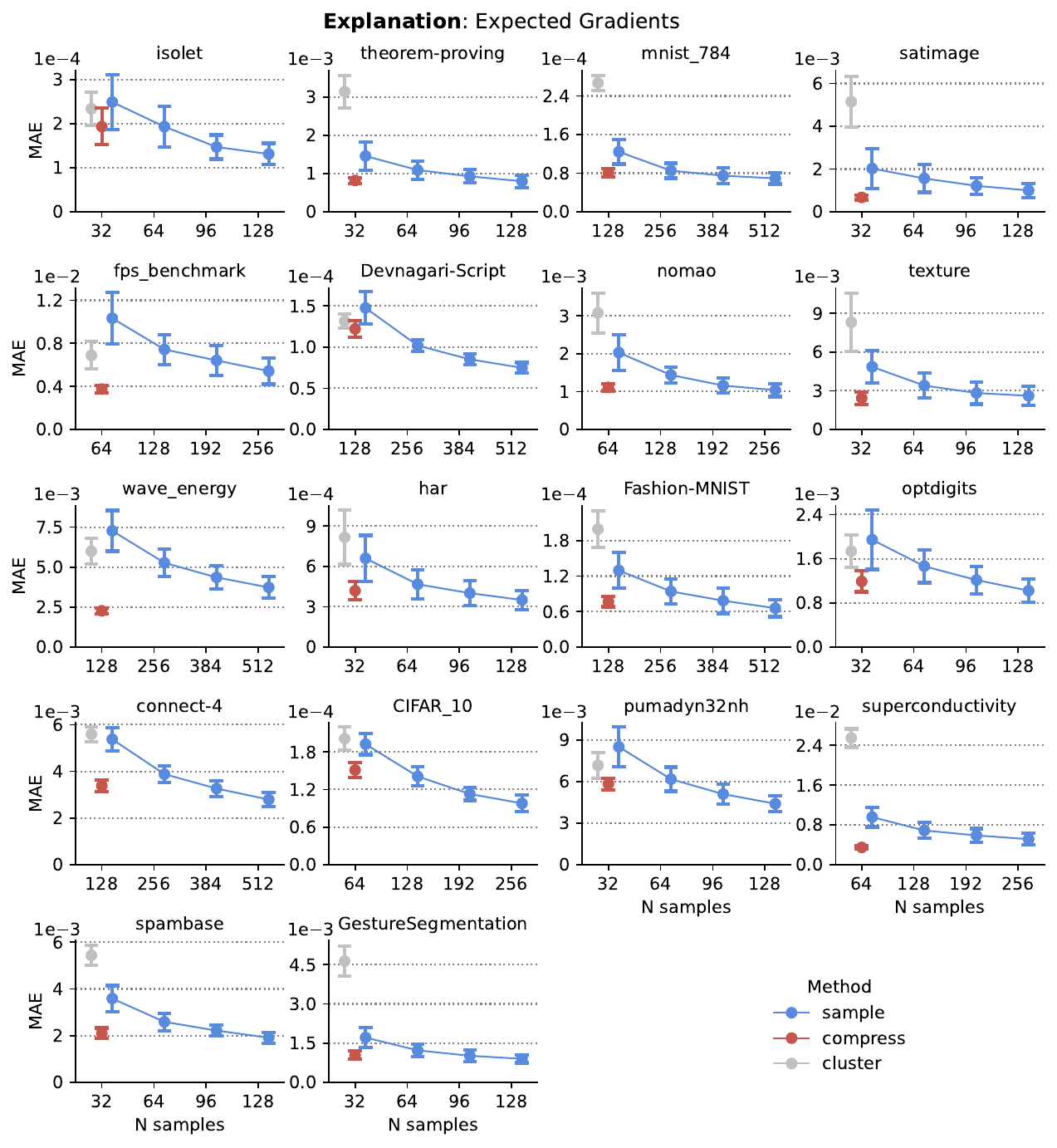}
    \caption{Extended Figure~\ref{fig:openml-expected-gradients-small}. 
    Comparison between \cte, \iid sampling and clustering for \expectedgradients explanations on 18 datasets. 
    We measure mean absolute error (MAE~$\downarrow$) between feature attribution values.
    \cte is not only more efficient and accurate, but also more stable as measured with deviation.
    (mean $\pm$~sd.)}
    \label{fig:openml-expected-gradients-full}
\end{figure}

Figure~\ref{fig:openml-expected-gradients-full} shows the \expectedgradients approximation error for 18 datasets. In all cases, \cte achieves on-par approximation error using fewer samples than \iid sampling, i.e. requiring fewer model inferences, resulting in faster computation and saved resources. 

\begin{figure}[h]
    \centering
    \includegraphics[width=0.66\textwidth]{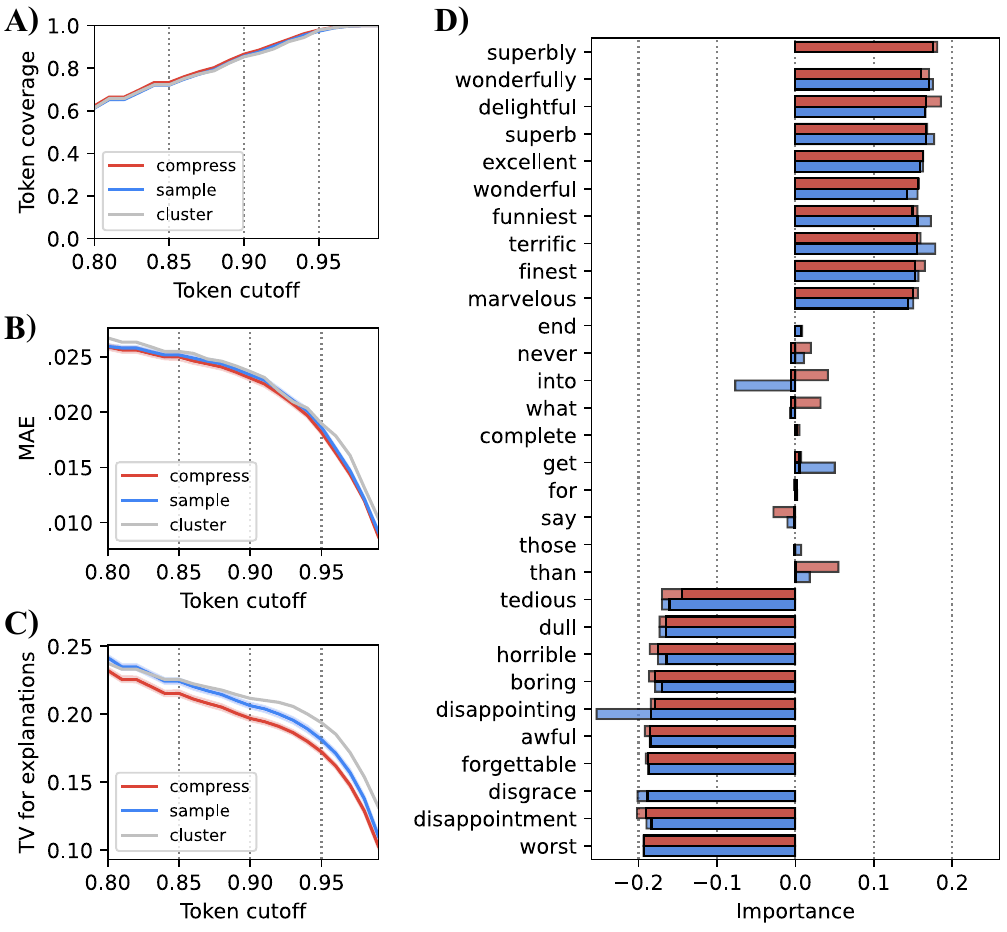}
    \caption{\cte for \glime of a DistilBERT model classifying IMDB reviews. \textbf{A)} It is not obvious how to measure the distance between global explanations containing different sets of tokens (Token coverage in \% w.r.t. ground truth, $\uparrow$). Therefore, we gradually remove rare tokens from the measurement based on their occurrence in the ground truth explanation (Token cutoff in quantiles). \textbf{B)} Measurement of mean absolute error (MAE, $\downarrow$) between aggregated global explanations. \textbf{C)}~Measurement of total variation distance ($\dtv$, $\downarrow$) between token occurrences in local explanations before global aggregation. \textbf{D)} We show an exemplary ``worst-case'' explanation, i.e. with the lowest MAE for cutoff 0.95 where token coverage is over 99\%, for both \cte and \iid sampling. For this visualization, we only show the importance of the 5 most positive/negative tokens, and 5 tokens with the importance closest to zero. Explanation approximation error is indicated with transparent bars. Notably, \iid sampling misses containing any input with an important token ``superbly'', while \cte misses ``disgrace''. Sampling overestimates the global importance of tokens ``disappointing'', ``into'' and ``get'', while \cte, for example, overestimates ``than'' and underestimates ``tedious'' or ``delightful''.}
    \label{fig:imdb-lime}
\end{figure}

\clearpage
\textbf{Model-agnostic explanation of a language model.} 
We further experimented with applying \cte to improve the estimation of global aggregated \lime~\citep{ribeiro2016should}, aka \glime~\citep{li2023glime}, which is a more complex setup that we leave for future work. 
We aim to explain the predictions of a DistilBERT language model\footnote{\url{https://huggingface.co/dfurman/distilbert-base-uncased-imdb}} trained on the IMDB dataset\footnote{\url{https://huggingface.co/datasets/stanfordnlp/imdb}} for sentiment analysis. We calculate \lime with $k=10$ for all samples from the validation set using an A100 GPU and aggregate these local explanations into global token importance with a mean of absolute normalized values~\citep{li2023glime}, which is the ``ground truth'' explanation. We then compress the set with \iid sampling, \cte, and clustering based on the inputs' text embeddings from the model's last layer (preceding a classifier) that has a dimension of size 768. Figure~\ref{fig:imdb-lime} shows results for explanation approximation error and an exemplary comparison between the explanations relating to Figure~\ref{fig:abstract}. To obtain these results, we used $8\times$ more samples than the typical compression scenario (still $25\times$ fewer than the full sample) so as to overcome the issue of rare tokens skewing the results. It becomes challenging to compute the distance between the ground truth and approximated explanations as the latter contains significantly fewer tokens (features), as opposed to previous experiments where these two explanations always had equal dimensions. Thus, MAE becomes biased towards sparse explanations and popular tokens, i.e. an explanation with a single token of well-approximated importance could have an error close to~0. For context, we measure $\dtv$ between the discrete distributions of tokens in local explanations before the global aggregation (lower is better). We report results for different token cutoffs, where we remove the tokens from the ground truth explanation by their rarity, which saturates at 5\% tokens left. 

\clearpage
\section{Ablations}\label{app:experiments-openml}

\begin{figure}[h]
    \centering
    \includegraphics[width=0.9\textwidth]{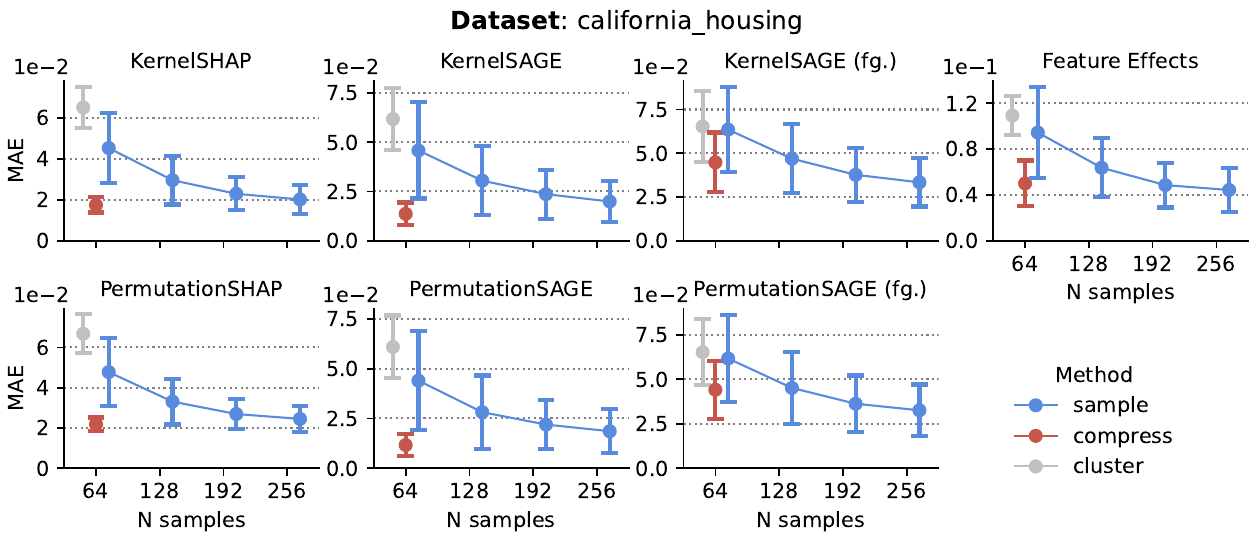}
    
    \vspace{1em}
    \includegraphics[width=0.9\textwidth]{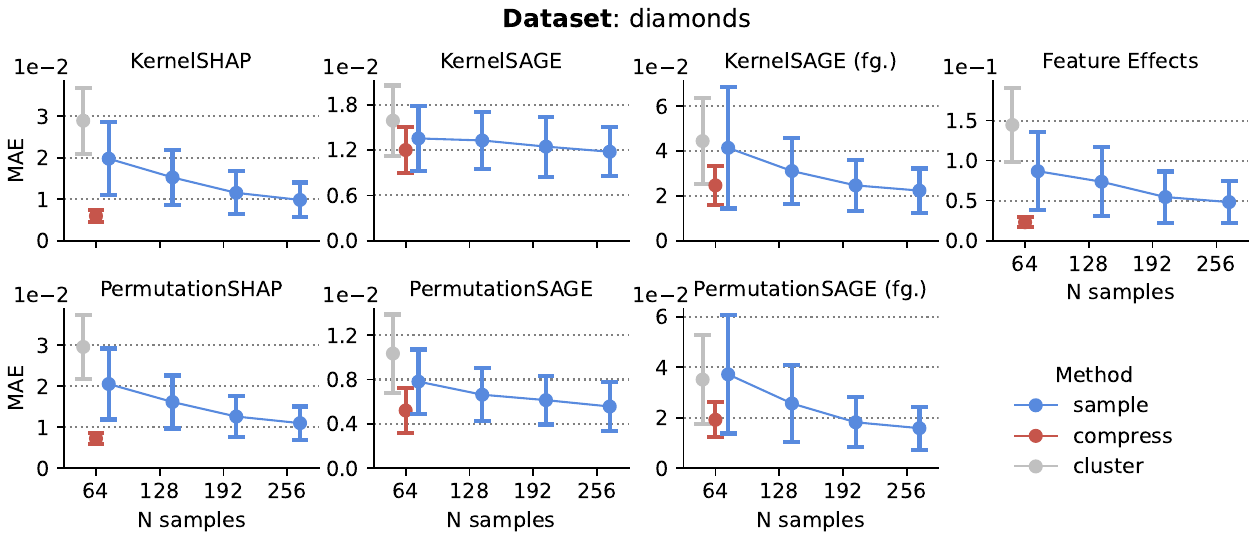}
    
    \vspace{1em}
    \includegraphics[width=0.9\textwidth]{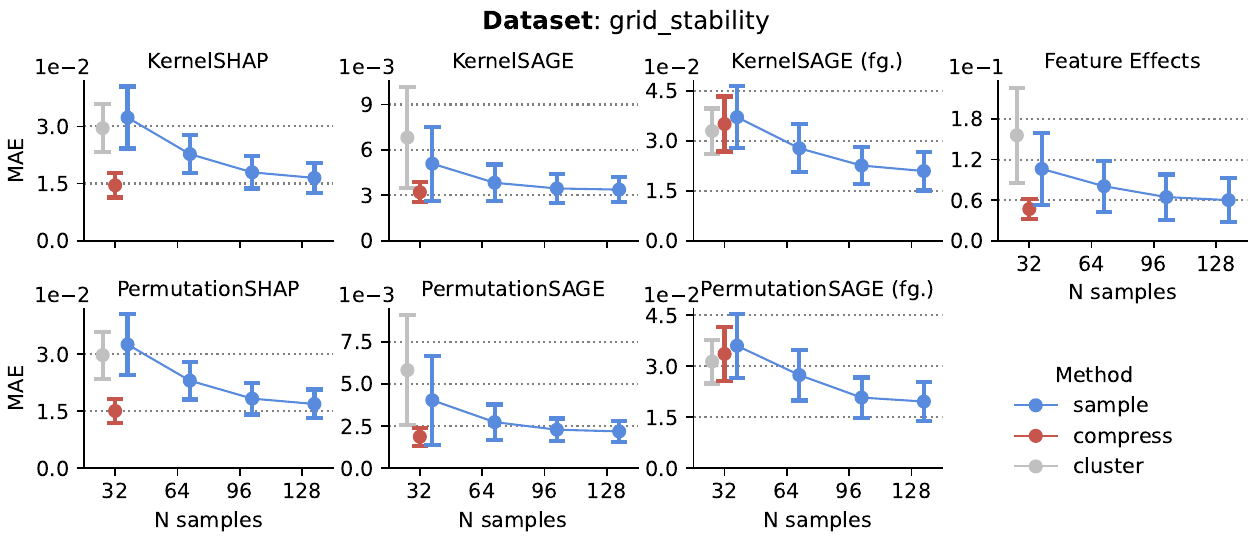}
    \caption{Extended Figure~\ref{fig:openml-shap-sage-small} (1/10). \cte improves the explanation approximation error of various local and global removal-based explanations. \sage is evaluated in two variants that consider either compressing only the background data (default), or using the compressed samples as both background and foreground data (as indicated with ``fg.''). (mean $\pm$ sd.)}
    \label{fig:openml_1}
\end{figure}

\begin{figure}
    \centering
    \includegraphics[width=\textwidth]{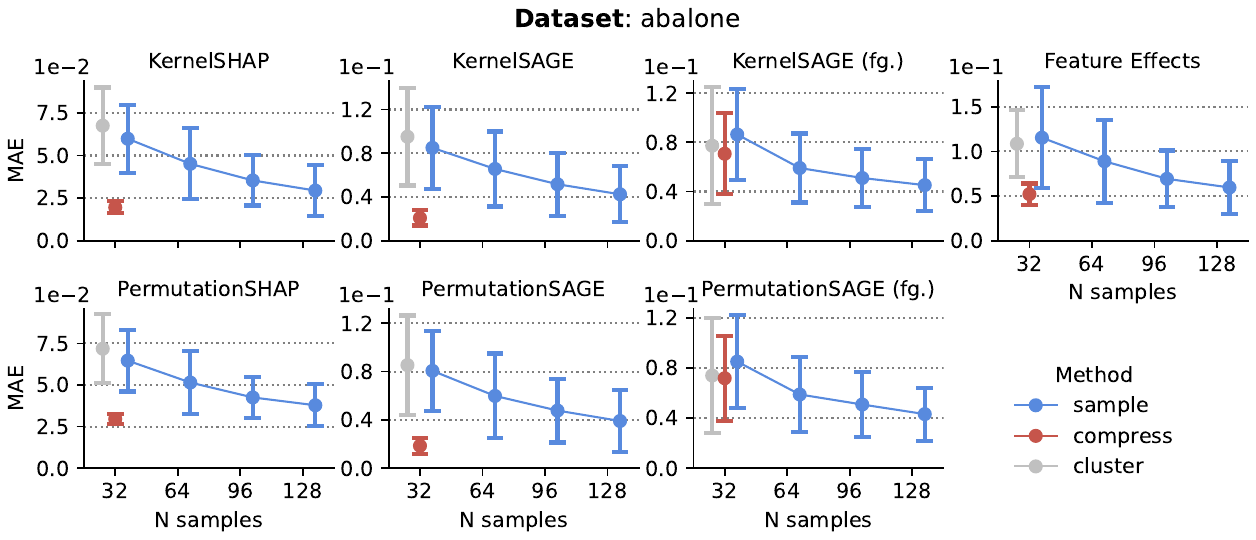}
    
    \vspace{1em}
    \includegraphics[width=\textwidth]{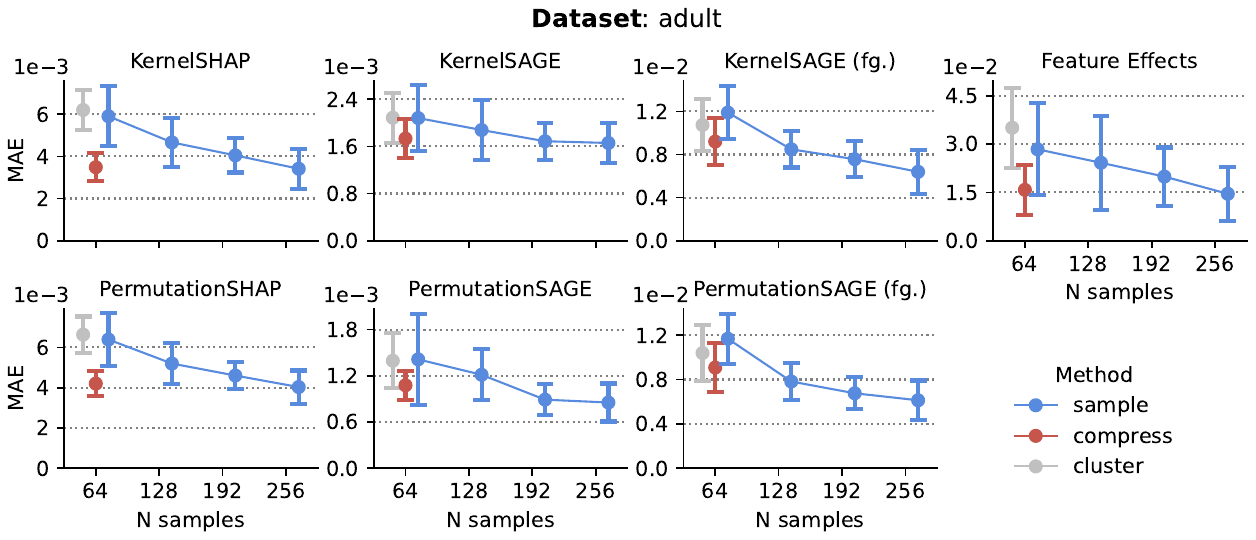}
    
    \vspace{1em}
    \includegraphics[width=\textwidth]{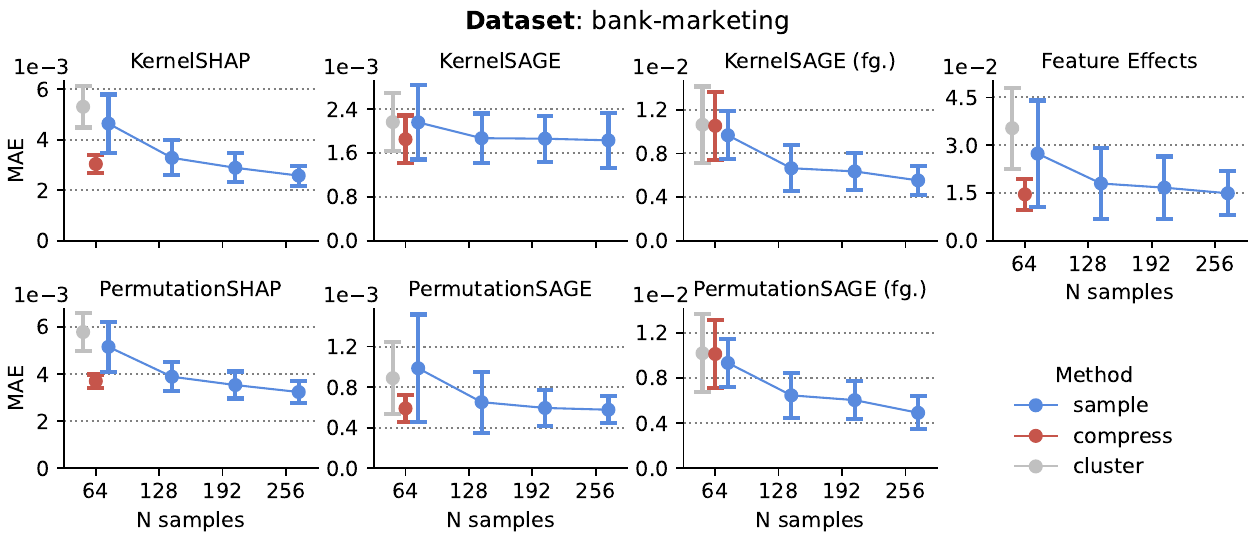}
    \caption{Extended Figure~\ref{fig:openml-shap-sage-small} (2/10). \cte improves the explanation approximation error of various local and global removal-based explanations. \sage is evaluated in two variants that consider either compressing only the background data (default), or using the compressed samples as both background and foreground data (as indicated with ``fg.''). (mean $\pm$ sd.)}
    \label{fig:openml_2}
\end{figure}

\begin{figure}
    \centering
    \includegraphics[width=\textwidth]{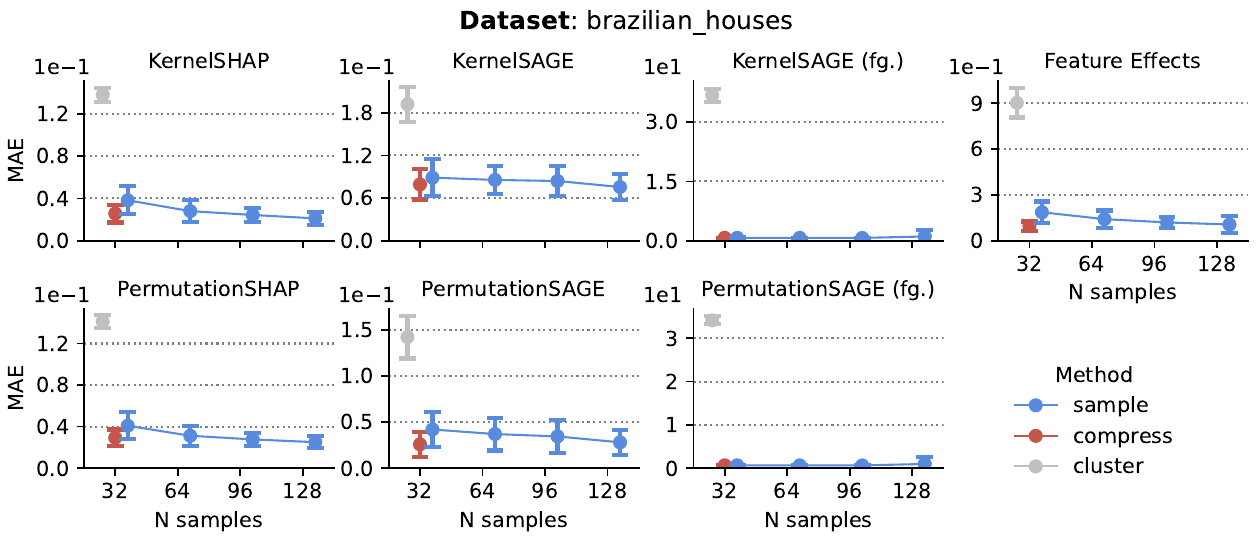}

    \vspace{1em}
    \includegraphics[width=\textwidth]{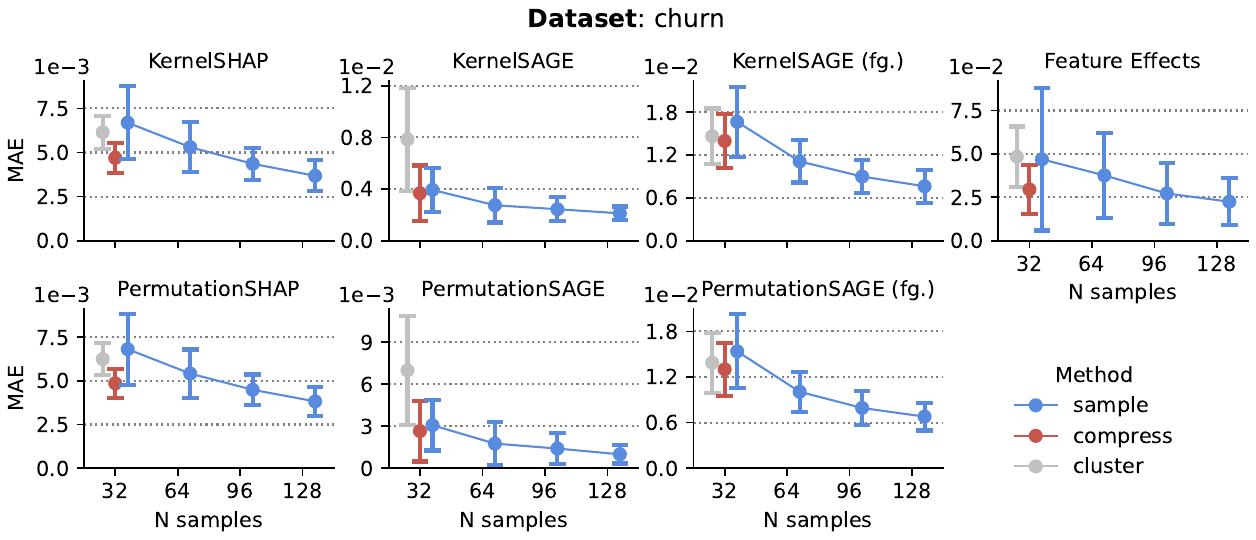}

    \vspace{1em}
    \includegraphics[width=\textwidth]{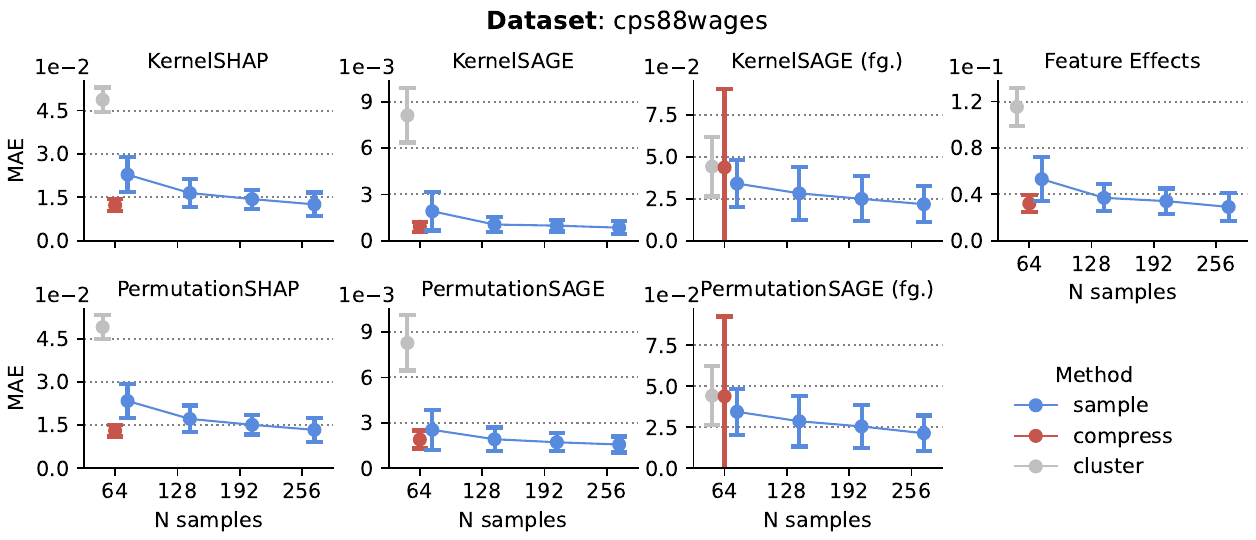}
    \caption{Extended Figure~\ref{fig:openml-shap-sage-small} (3/10). \cte improves the explanation approximation error of various local and global removal-based explanations. \sage is evaluated in two variants that consider either compressing only the background data (default), or using the compressed samples as both background and foreground data (as indicated with ``fg.''). (mean $\pm$ sd.)}
    \label{fig:openml_3}
\end{figure}

\begin{figure}
    \centering
    \includegraphics[width=\textwidth]{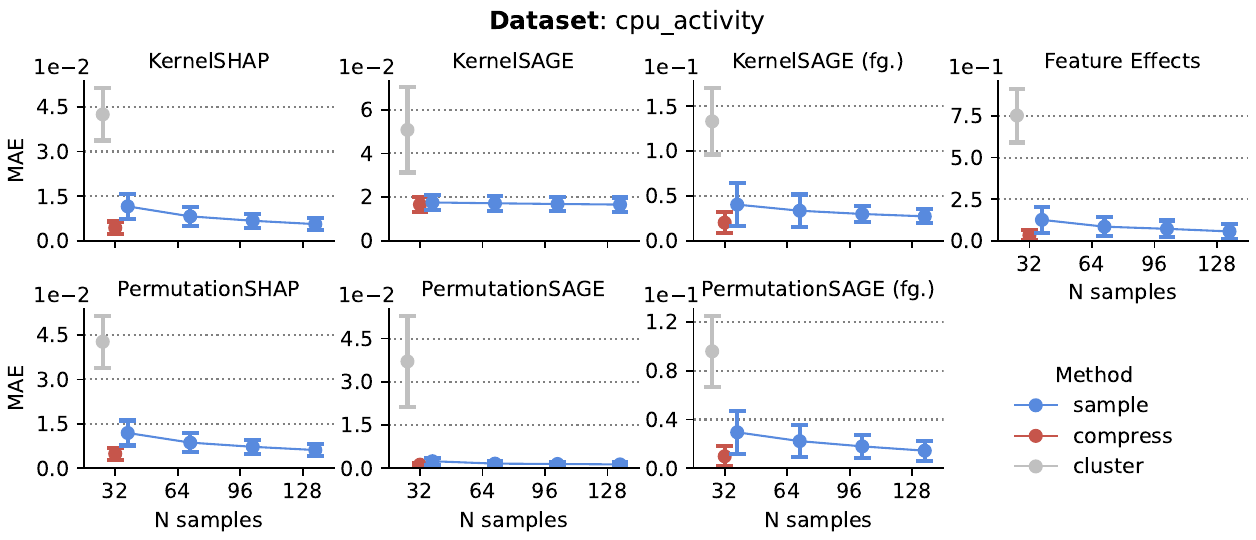}

    \vspace{1em}
    \includegraphics[width=\textwidth]{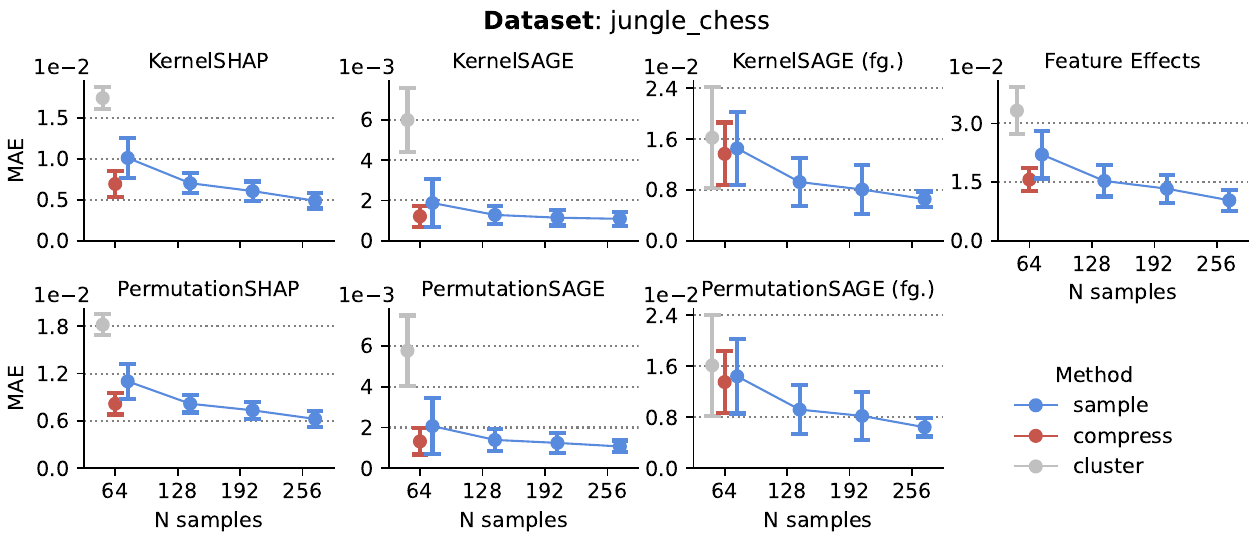}

    \vspace{1em}
    \includegraphics[width=\textwidth]{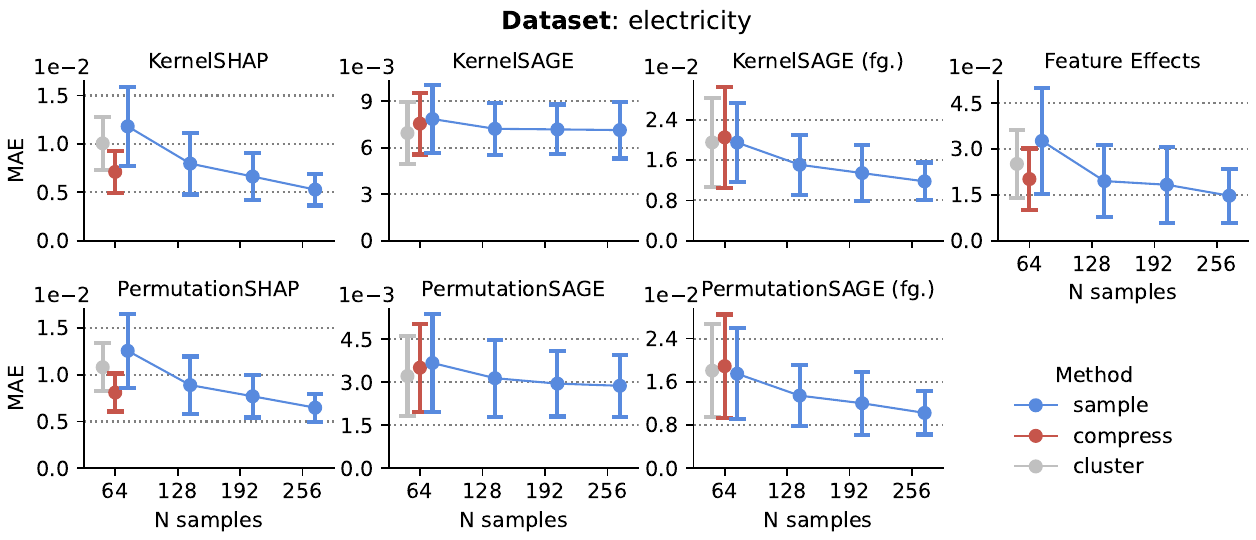}
    \caption{Extended Figure~\ref{fig:openml-shap-sage-small} (4/10). \cte improves the explanation approximation error of various local and global removal-based explanations. \sage is evaluated in two variants that consider either compressing only the background data (default), or using the compressed samples as both background and foreground data (as indicated with ``fg.''). (mean $\pm$ sd.)}
    \label{fig:openml_4}
\end{figure}

\begin{figure}
    \centering
    \includegraphics[width=\textwidth]{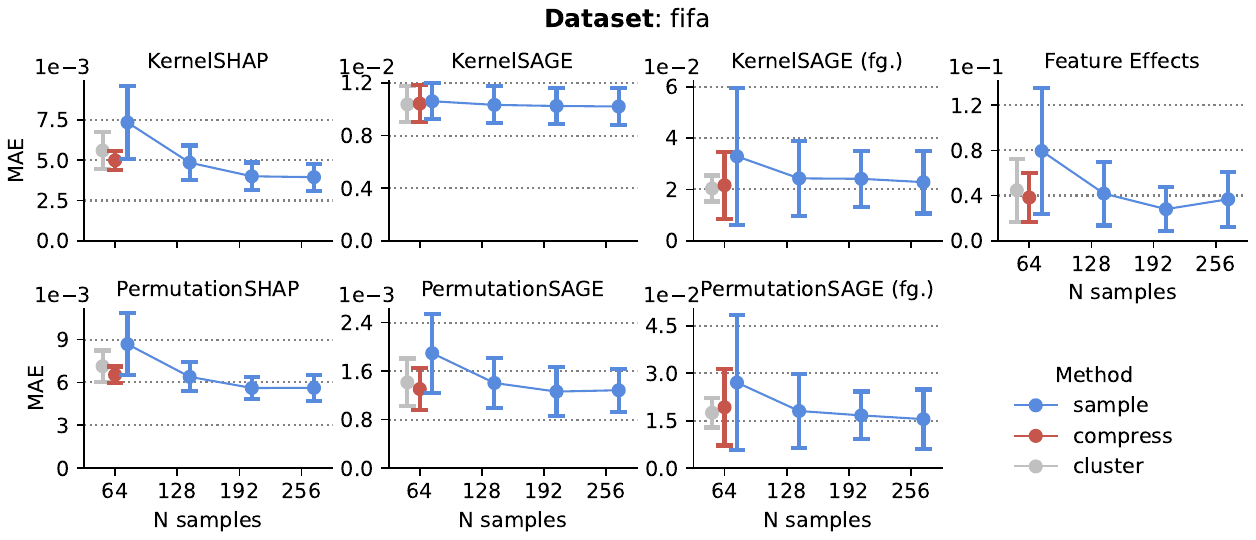}

    \vspace{1em}
    \includegraphics[width=\textwidth]{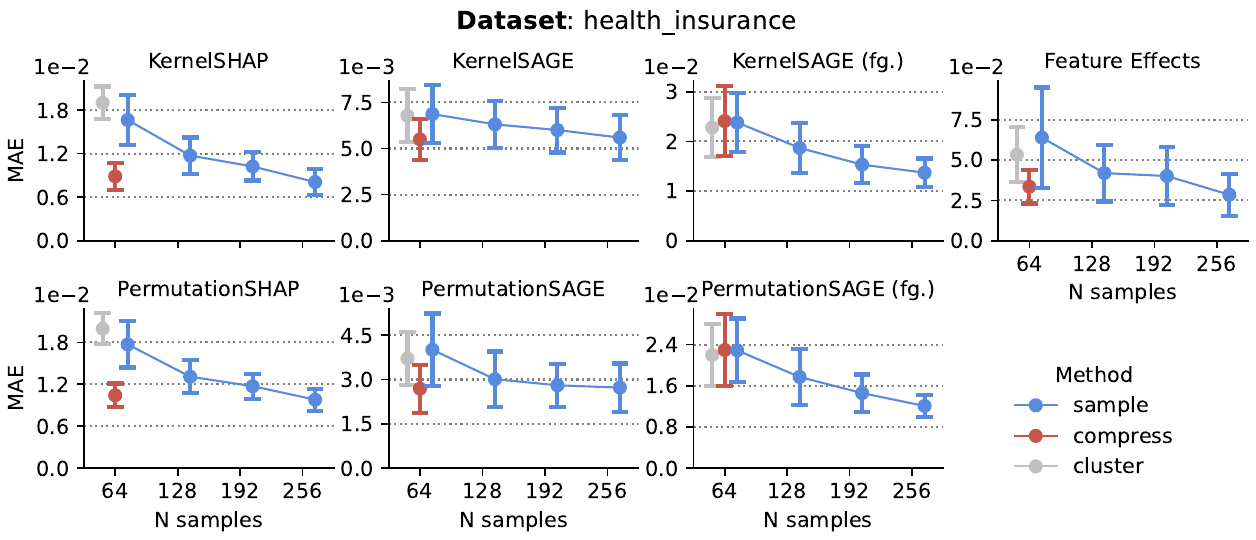}

    \vspace{1em}
    \includegraphics[width=\textwidth]{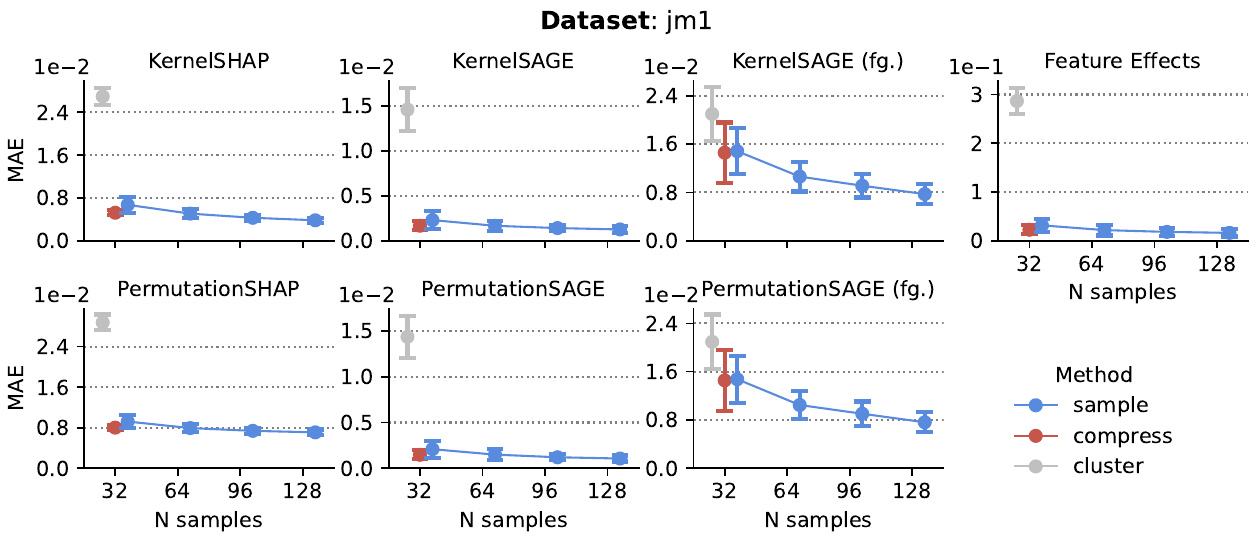}
    \caption{Extended Figure~\ref{fig:openml-shap-sage-small} (5/10). \cte improves the explanation approximation error of various local and global removal-based explanations. \sage is evaluated in two variants that consider either compressing only the background data (default), or using the compressed samples as both background and foreground data (as indicated with ``fg.''). (mean $\pm$ sd.)}
    \label{fig:openml_5}
\end{figure}

\begin{figure}
    \centering
    \includegraphics[width=\textwidth]{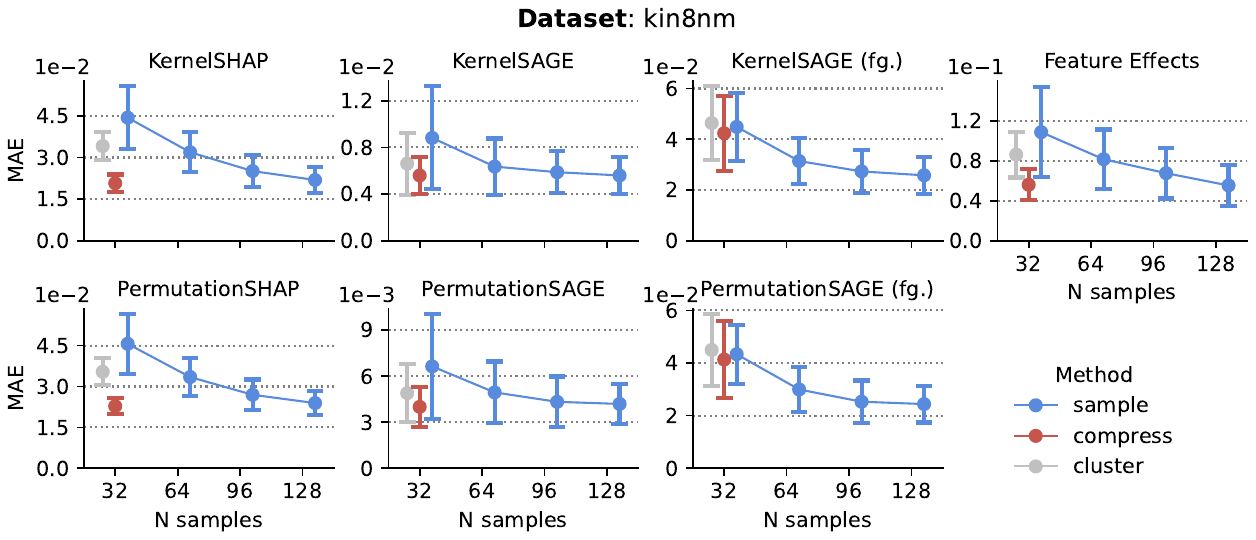}

    \vspace{1em}
    \includegraphics[width=\textwidth]{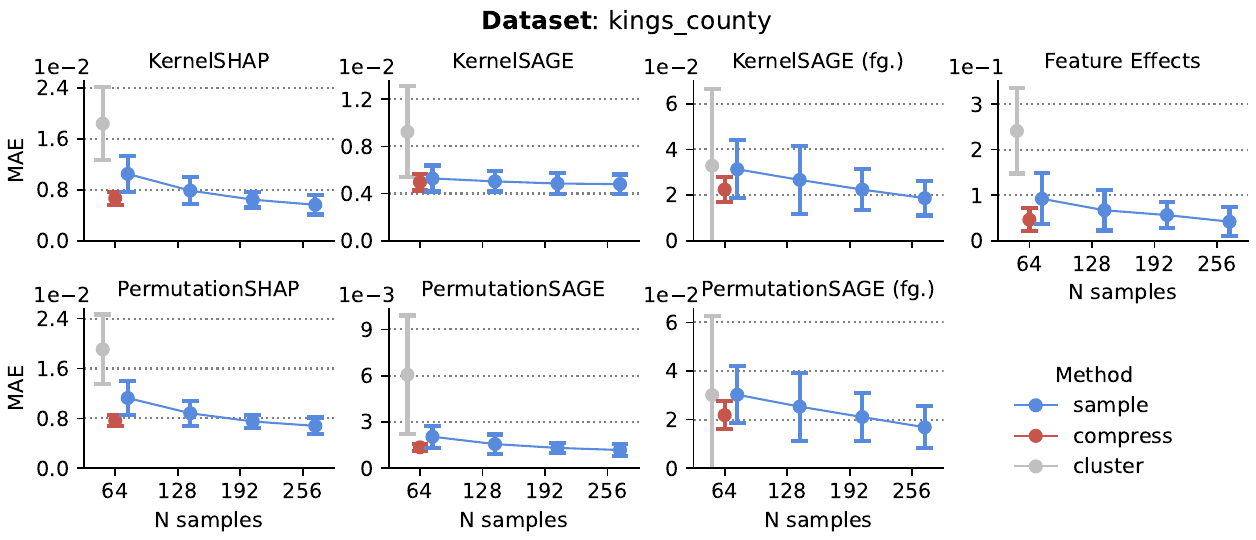}

    \vspace{1em}
    \includegraphics[width=\textwidth]{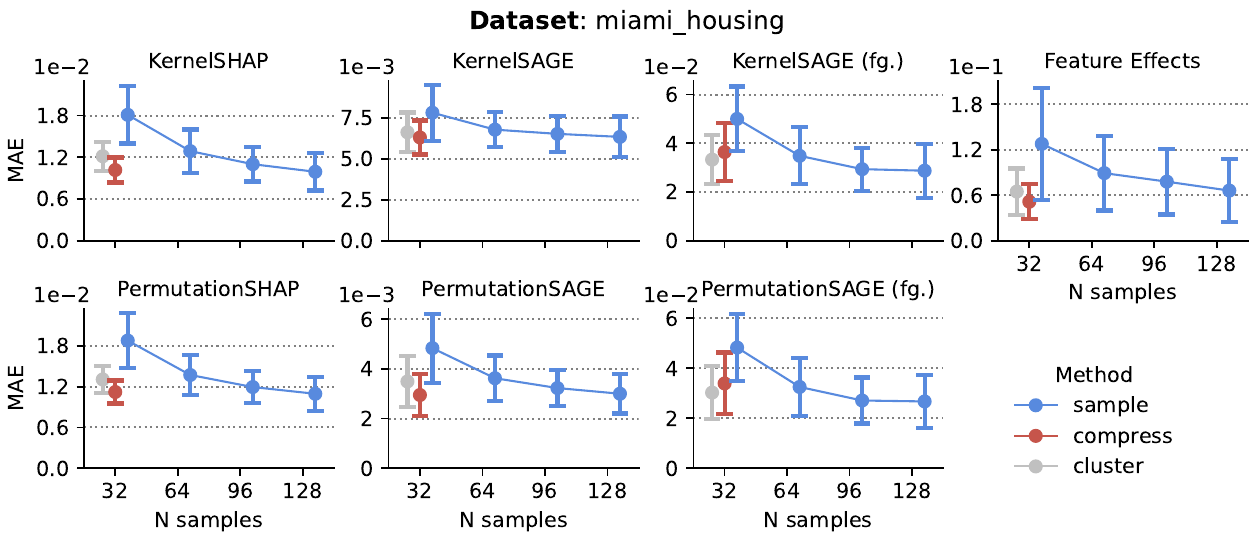}
    \caption{Extended Figure~\ref{fig:openml-shap-sage-small} (6/10). \cte improves the explanation approximation error of various local and global removal-based explanations. \sage is evaluated in two variants that consider either compressing only the background data (default), or using the compressed samples as both background and foreground data (as indicated with ``fg.''). (mean $\pm$ sd.)}
    \label{fig:openml_6}
\end{figure}

\begin{figure}
    \centering
    \includegraphics[width=\textwidth]{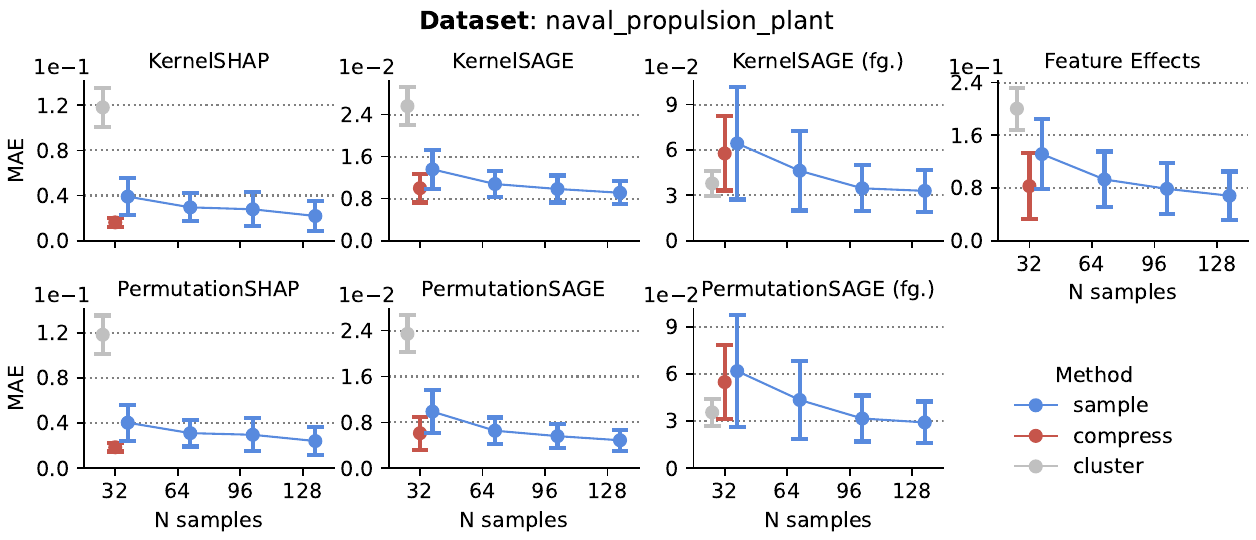}

    \vspace{1em}
    \includegraphics[width=\textwidth]{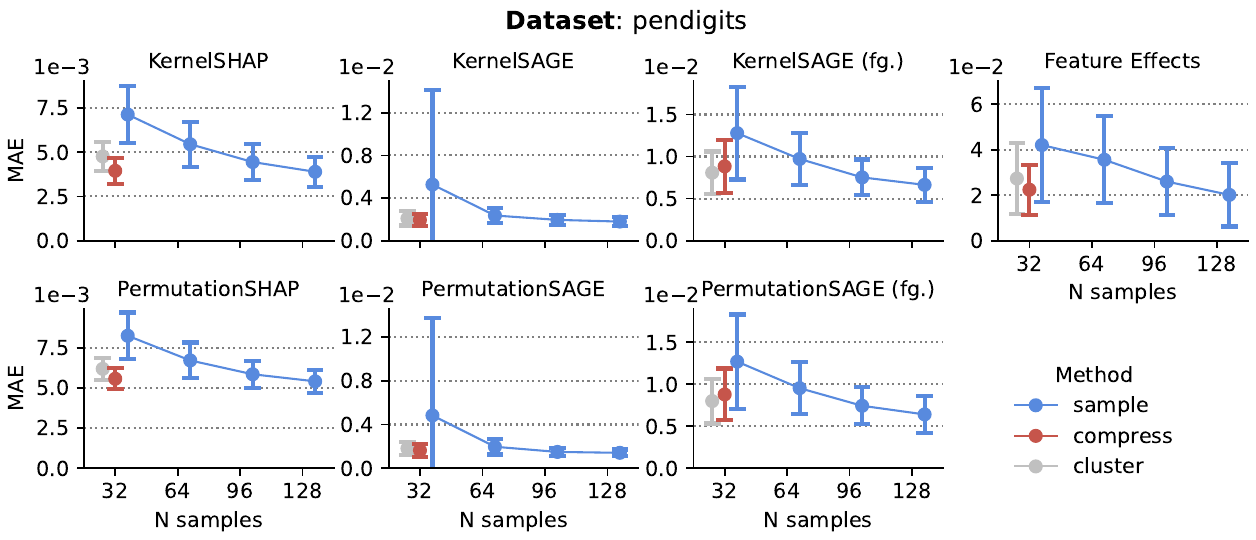}

    \vspace{1em}
    \includegraphics[width=\textwidth]{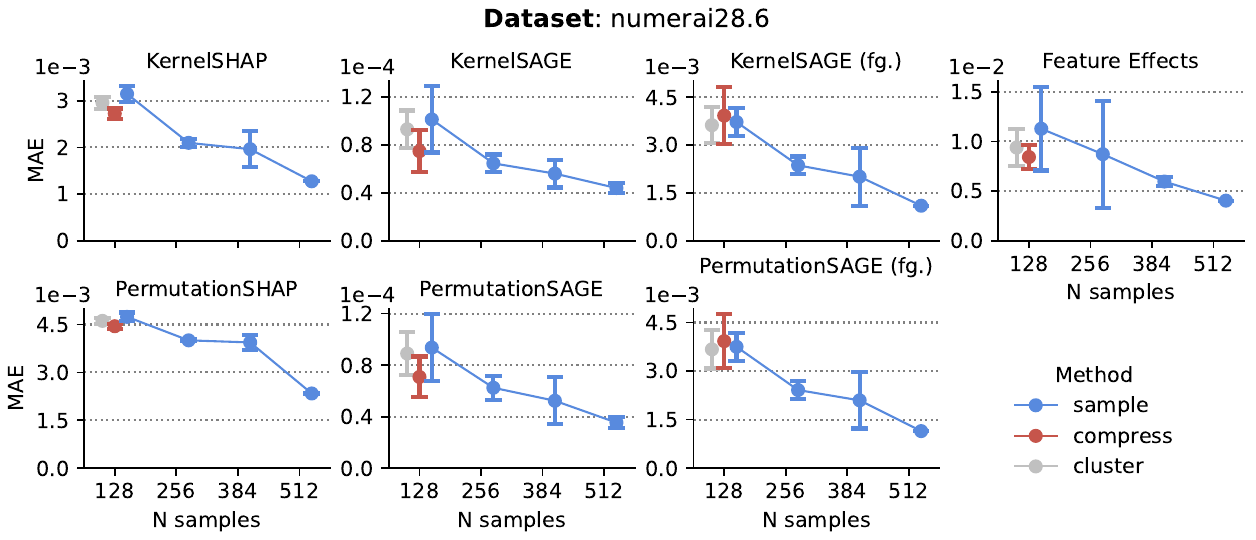}
    \caption{Extended Figure~\ref{fig:openml-shap-sage-small} (7/10). \cte improves the explanation approximation error of various local and global removal-based explanations. \sage is evaluated in two variants that consider either compressing only the background data (default), or using the compressed samples as both background and foreground data (as indicated with ``fg.''). (mean $\pm$ sd.)}
    \label{fig:openml_7}
\end{figure}

\begin{figure}
    \centering
    \includegraphics[width=\textwidth]{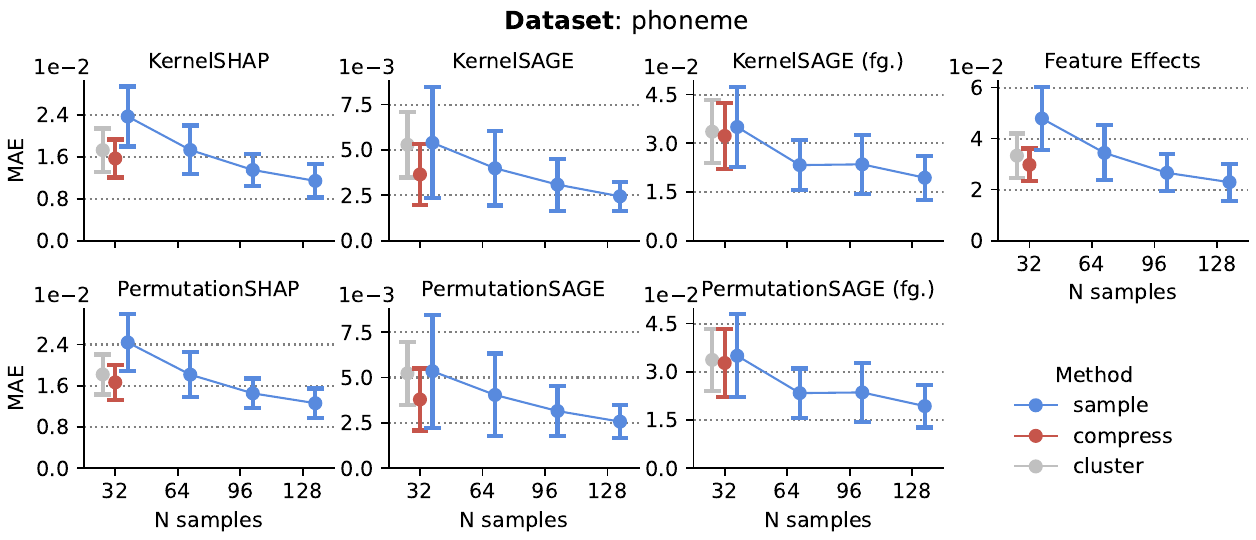}

    \vspace{1em}
    \includegraphics[width=\textwidth]{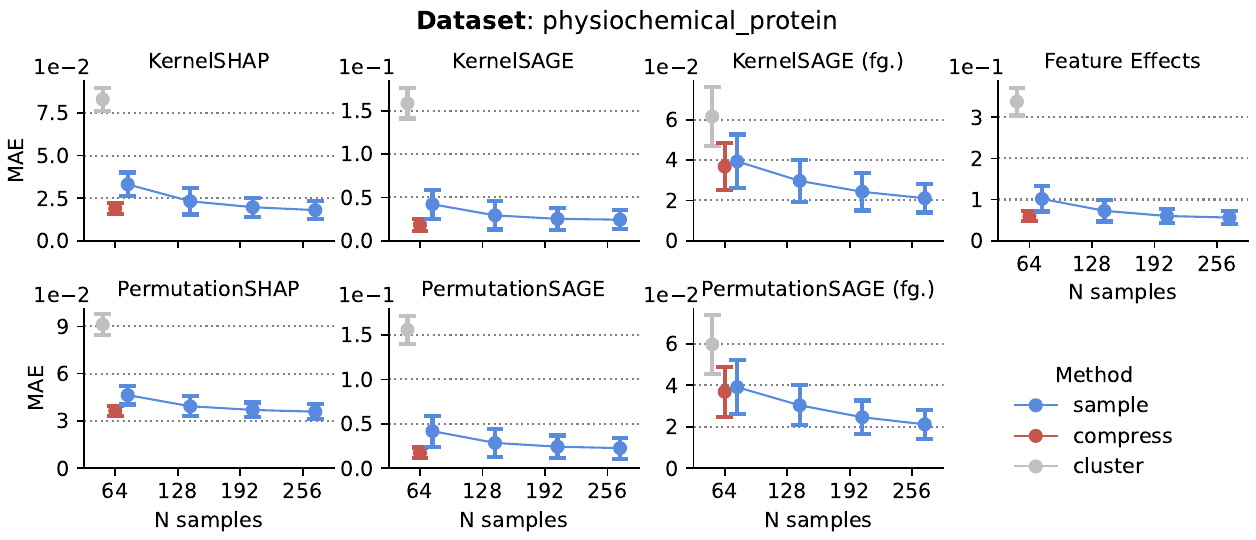}

    \vspace{1em}
    \includegraphics[width=\textwidth]{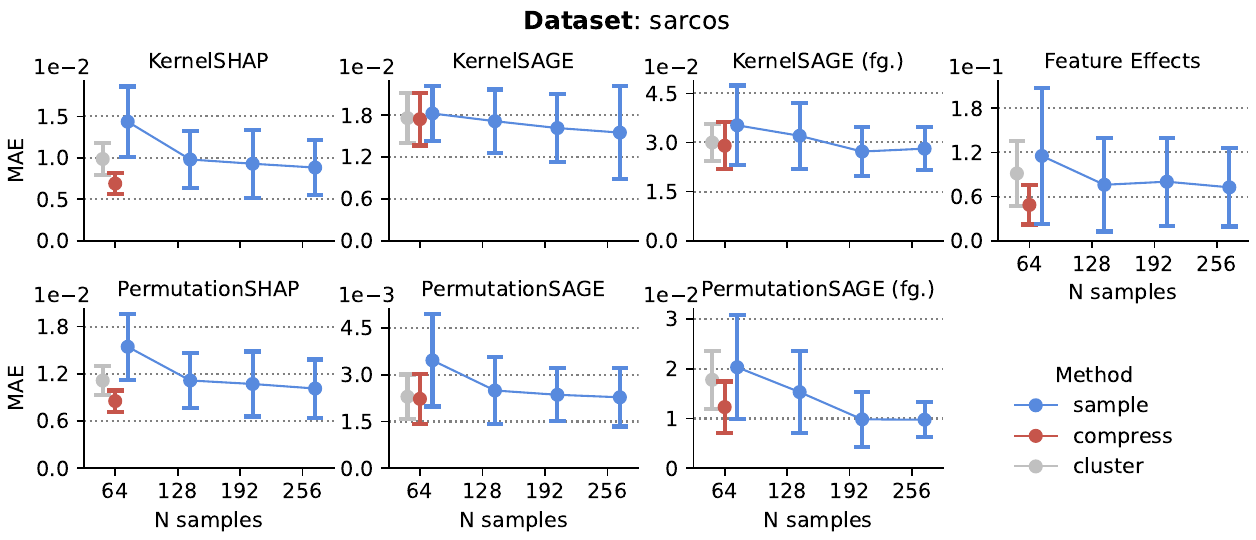}
    \caption{Extended Figure~\ref{fig:openml-shap-sage-small} (8/10). \cte improves the explanation approximation error of various local and global removal-based explanations. \sage is evaluated in two variants that consider either compressing only the background data (default), or using the compressed samples as both background and foreground data (as indicated with ``fg.''). (mean $\pm$ sd.)}
    \label{fig:openml_8}
\end{figure}

\begin{figure}
    \centering
    \includegraphics[width=\textwidth]{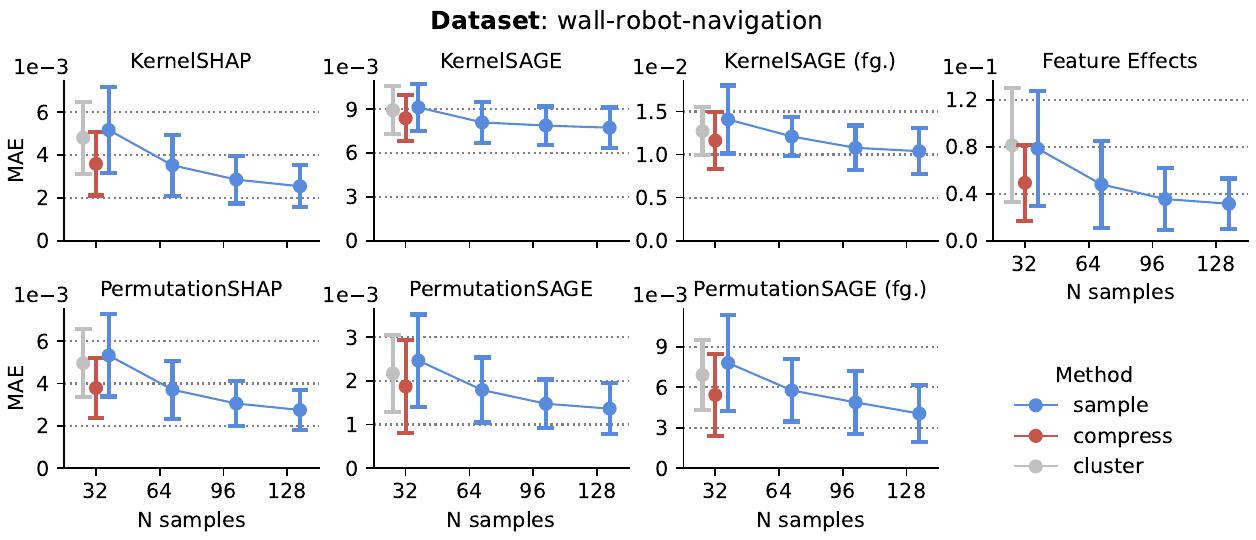}

    \vspace{1em}
    \includegraphics[width=\textwidth]{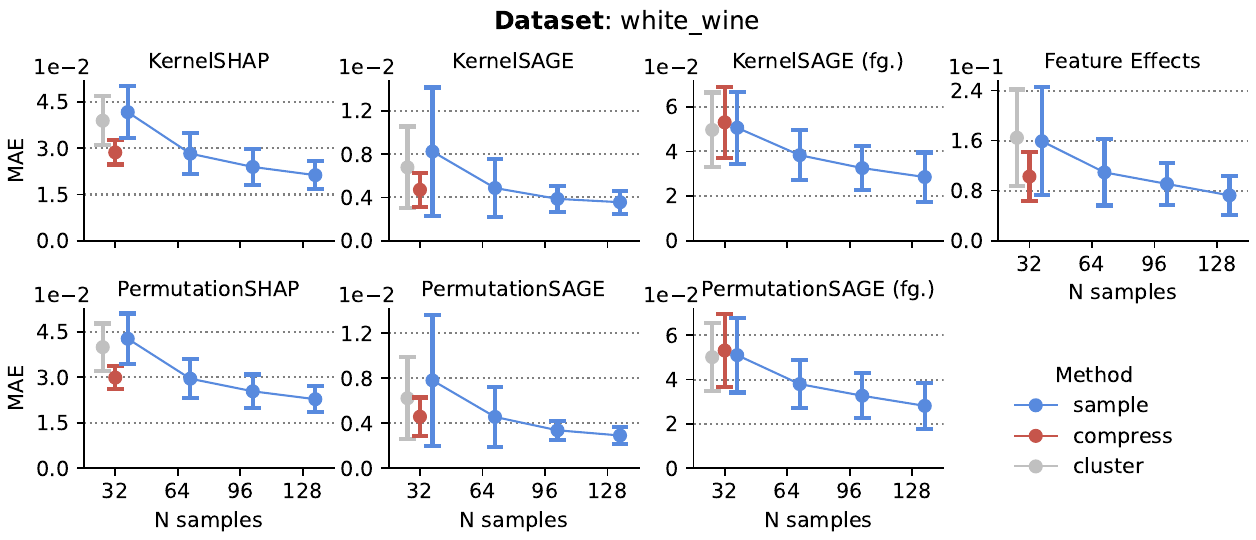}

    \vspace{1em}
    \includegraphics[width=\textwidth]{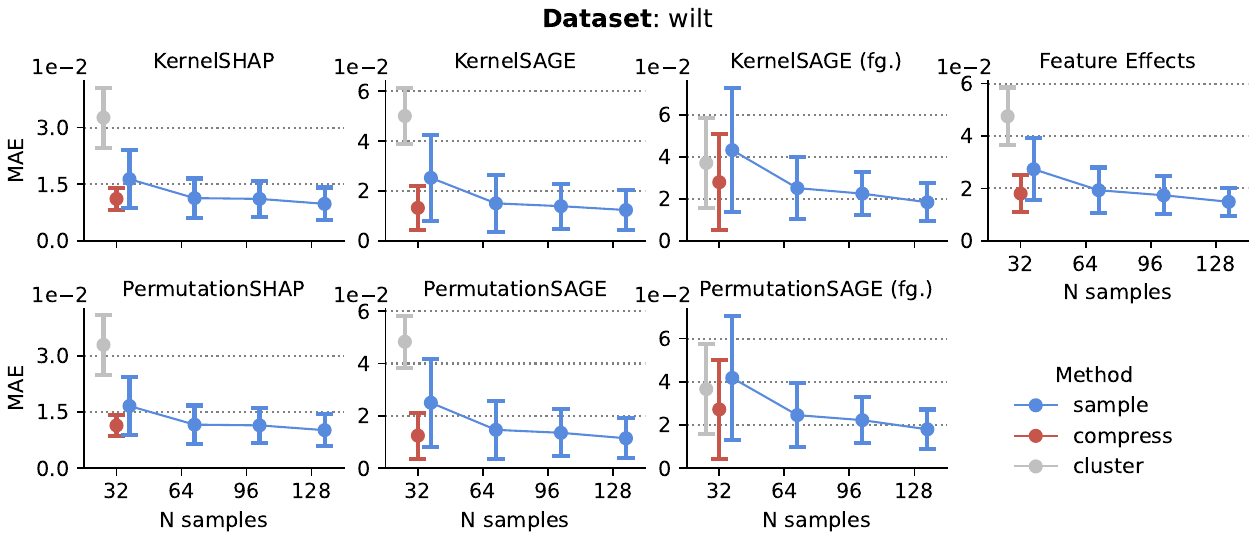}
    \caption{Extended Figure~\ref{fig:openml-shap-sage-small} (9/10). \cte improves the explanation approximation error of various local and global removal-based explanations. \sage is evaluated in two variants that consider either compressing only the background data (default), or using the compressed samples as both background and foreground data (as indicated with ``fg.''). (mean $\pm$ sd.)}
    \label{fig:openml_9}
\end{figure}

Figures~\ref{fig:openml_1}--\ref{fig:openml_9} report the explanation approximation error for 30 predictive tasks. We observe that \cte significantly improves the estimation of \featureeffects in all cases. Another insight is that, on average, \cte provides a smaller improvement over \iid sampling when considering compressing foreground data in \sage.

\begin{figure}
    \centering
    \includegraphics[width=\textwidth]{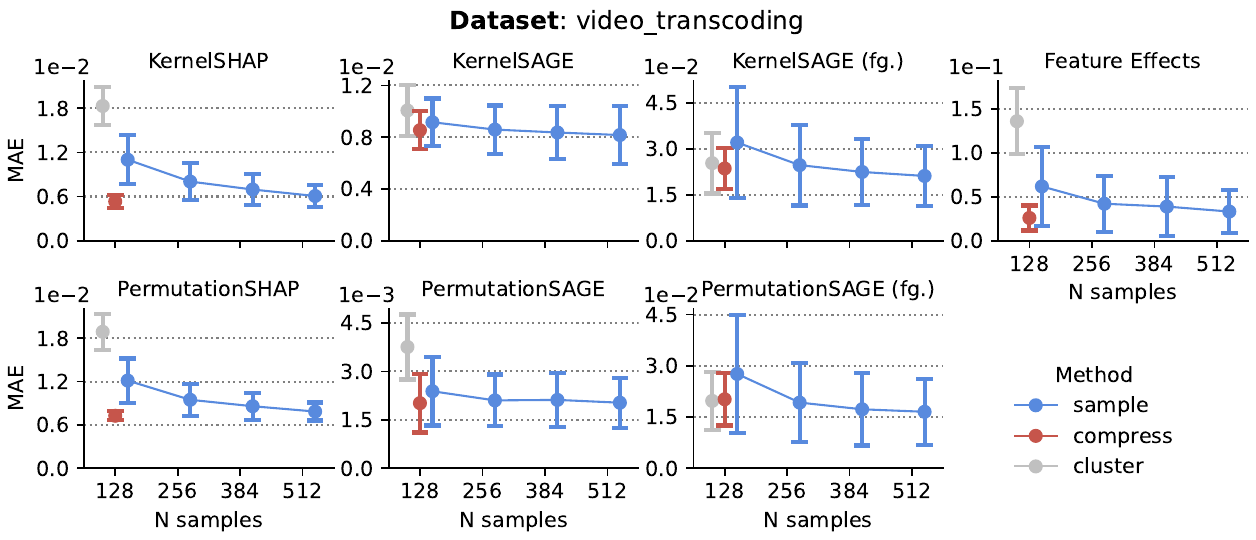}

    \vspace{1em}
    \includegraphics[width=\textwidth]{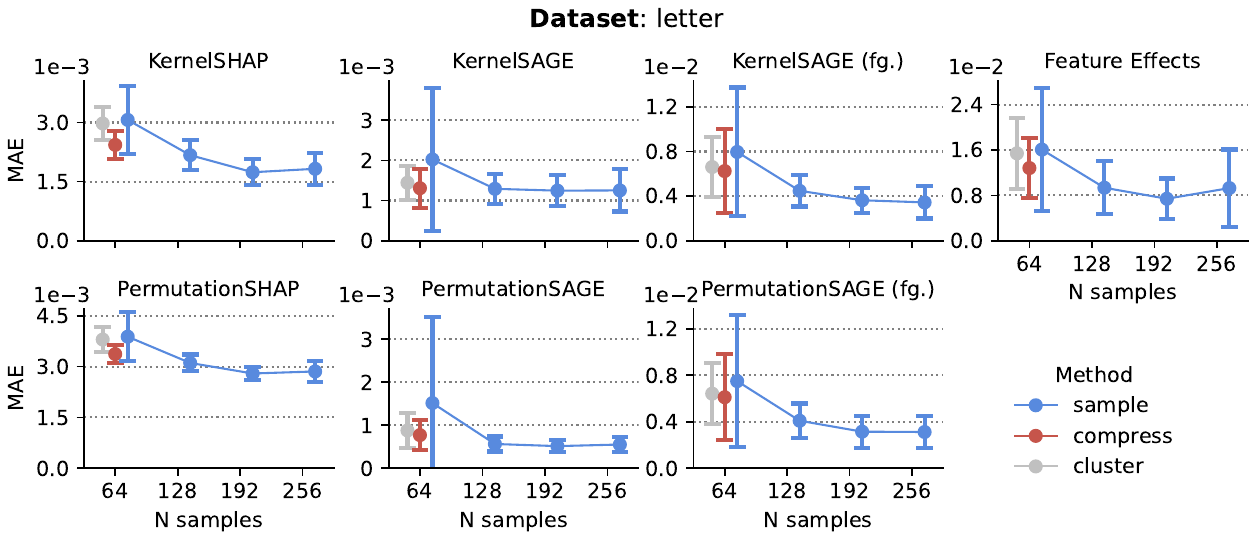}

    \vspace{1em}
    \includegraphics[width=\textwidth]{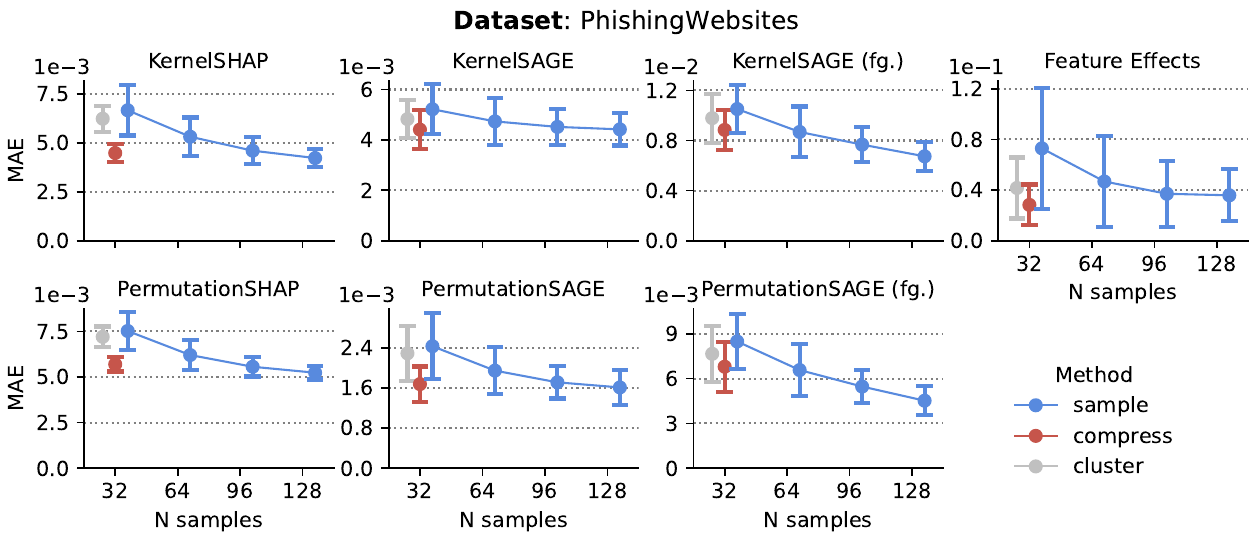}
    \caption{Extended Figure~\ref{fig:openml-shap-sage-small} (10/10). \cte improves the explanation approximation error of various local and global removal-based explanations. \sage is evaluated in two variants that consider either compressing only the background data (default), or using the compressed samples as both background and foreground data (as indicated with ``fg.''). (mean $\pm$ sd.)}
    \label{fig:openml_10}
\end{figure}

\clearpage
\section{Compute resources}\label{app:setup-compute-resources}

Experiments described in Sections~\ref{sec:experiments-accuracy}, \ref{sec:experiments-efficient} \& \ref{sec:experiments-sanity-check}, and Figure~\ref{fig:benchmark-compress-text-image}, were computed on a personal computer with an M3 chip as justified in the beginning of Section~\ref{sec:experiments}. 
Experiments described in Sections~\ref{sec:experiments-expected-gradients}~\&~\ref{sec:experiments-openml} were computed on a cluster with $4\times$ AMD Rome 7742 CPUs (256 cores) and 4TB of RAM for about 14 days combined.

\clearpage
\section{Visual comparison of explanations}\label{app:visual-comparison}

We provide exemplary visual comparisons between ground truth explanations and those estimated on an \iid and compressed sample in 4 experimental settings. 

\begin{figure}[ht]
    \centering
    \includegraphics[width=0.49\linewidth]{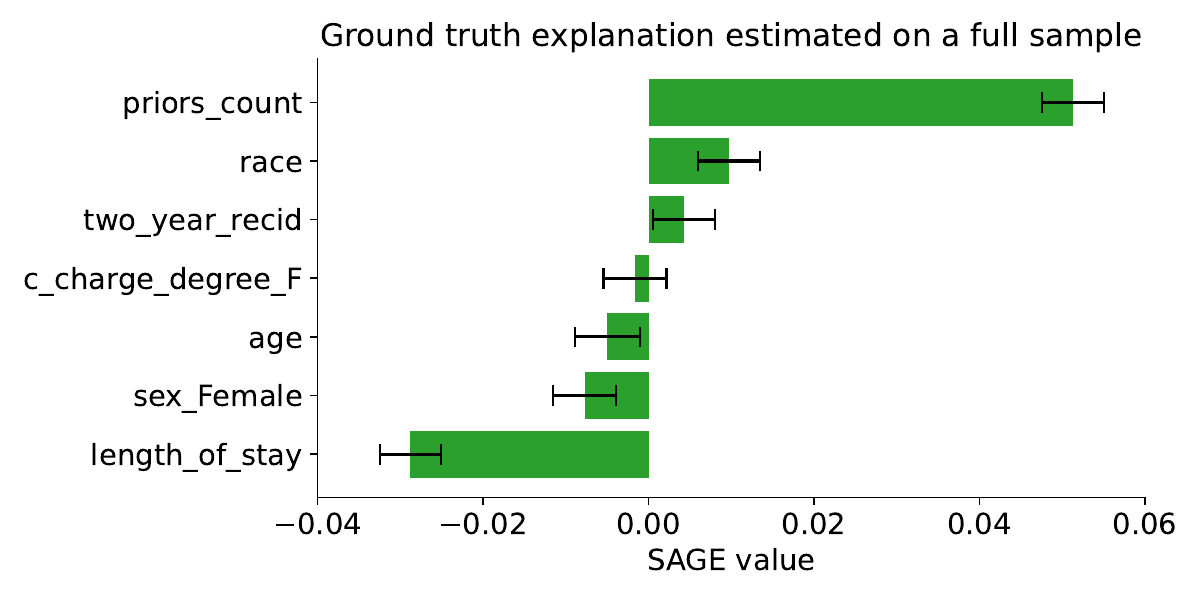}
    \includegraphics[width=0.49\linewidth]{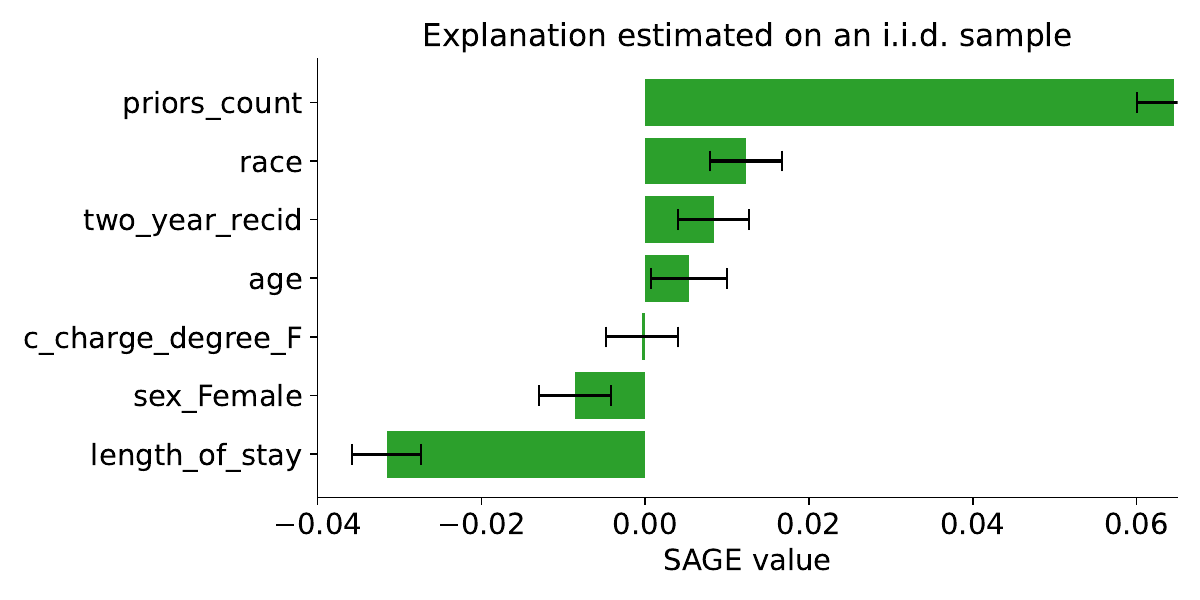}
    \vspace{-0.5em}
    
    \rule{13cm}{0.5pt}  
    
    \vspace{0.5em}
    \includegraphics[width=0.49\linewidth]{figures/example_sage_compas_gt.pdf}
    \includegraphics[width=0.49\linewidth]{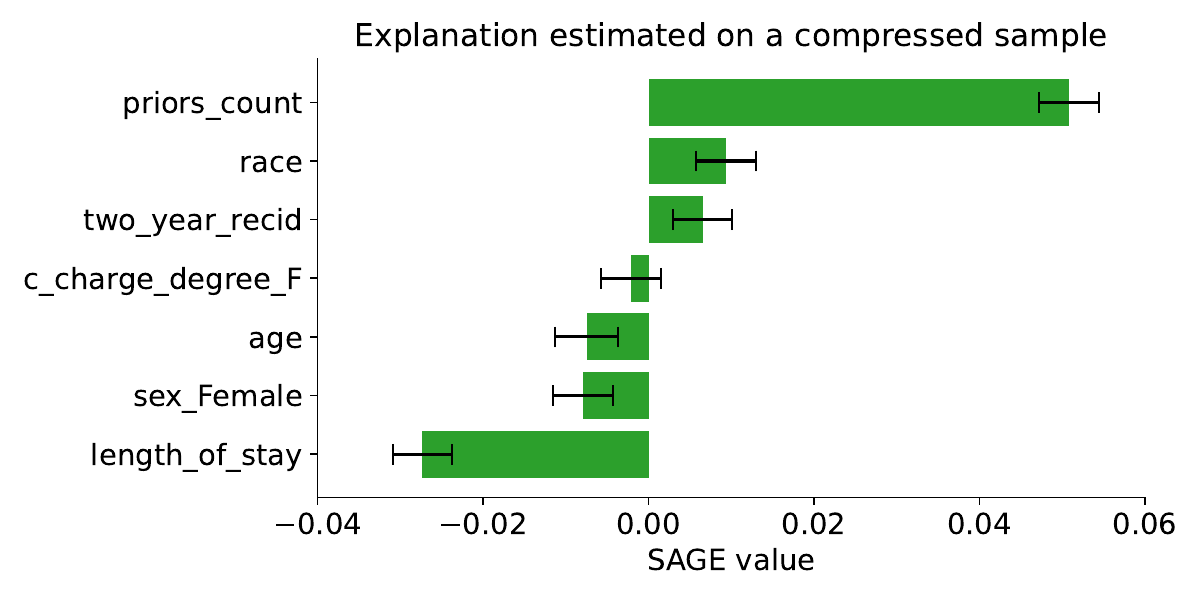}
        \caption{
        Comparison between \kernsage estimated on the full (left), sampled (right top), and compressed (right bottom) subsets of the \texttt{compas} dataset. 
        MAE introduced by \iid sampling equals $0.0050$ for the importance values and $0.00033$ for their standard deviations (error bars), by \cte is $0.0011$ and $0.00007$ respectively, and so the relative improvement of \cte is~78\% in both cases.
        }
    \label{fig:example_sage_compas}
    \vspace{1em}
    \includegraphics[width=0.49\linewidth]{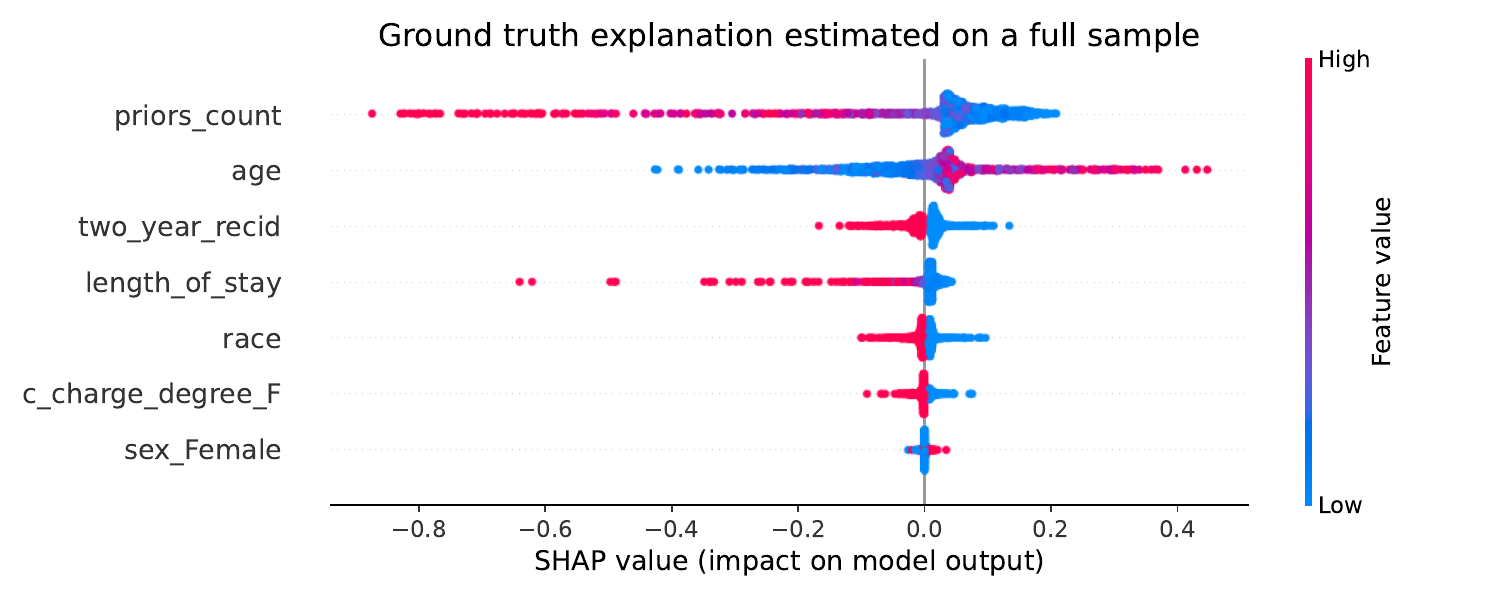}
    \includegraphics[width=0.49\linewidth]{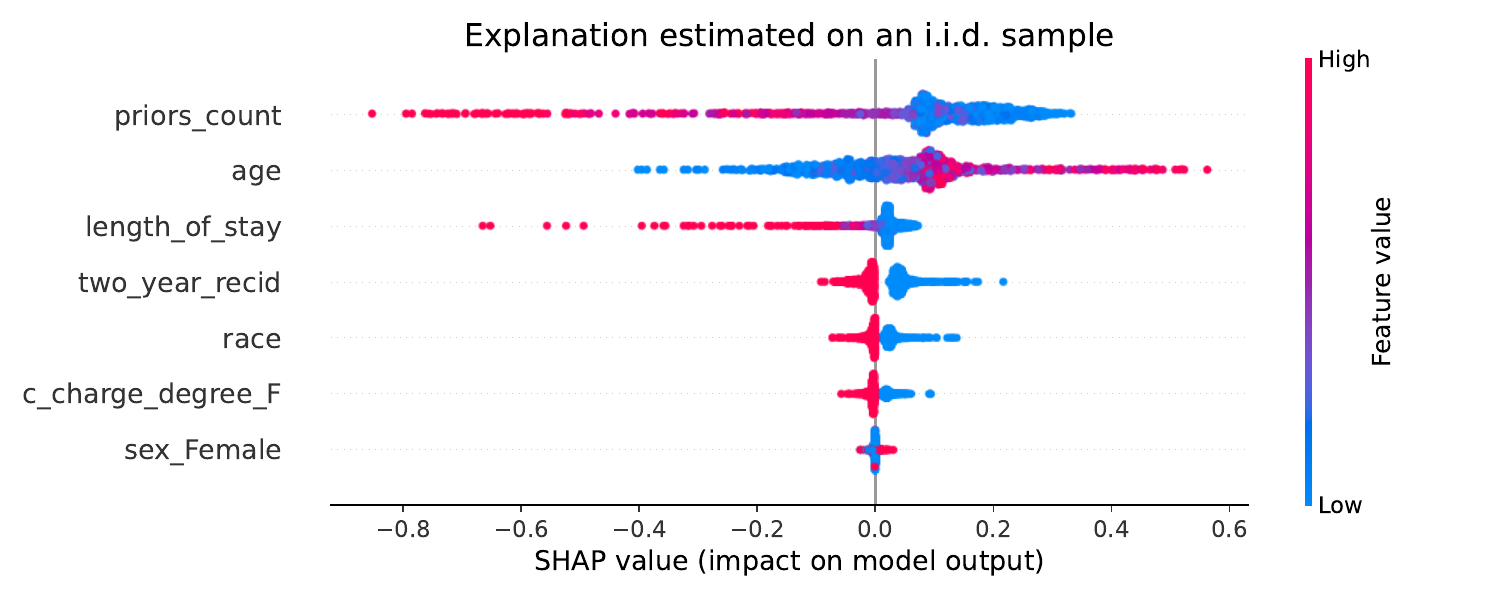}
    \vspace{-0.5em}
    
    \rule{13cm}{0.5pt}  
    
    \vspace{0.5em}
    \includegraphics[width=0.49\linewidth]{figures/example_shap_compas_gt.pdf}
    \includegraphics[width=0.49\linewidth]{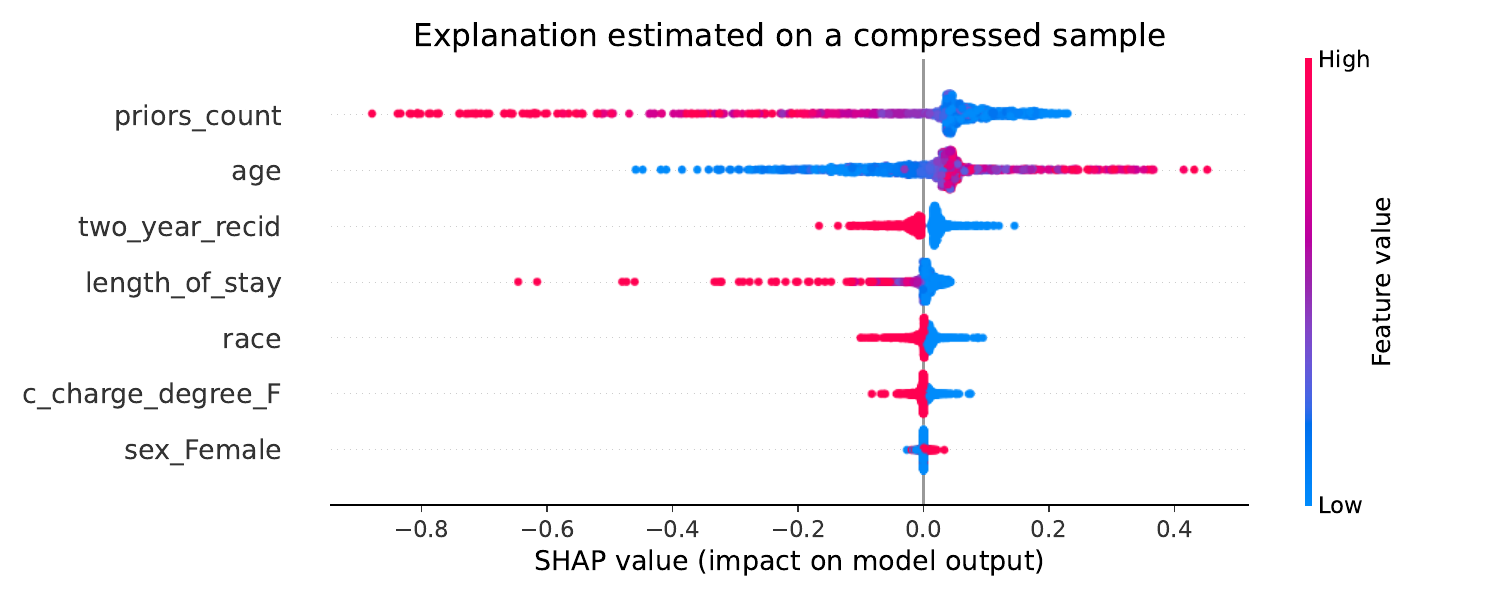}
        \caption{
        Comparison between all local \permshap explanations estimated on full (left), sampled (right top), and compressed (right bottom) subsets of the \texttt{compas} dataset. 
        MAE introduced by \iid sampling equals $0.0227$, by \cte is $0.0032$, and so the relative improvement of \cte is~86\%.
        }
    \label{fig:example_shap_compas}
\end{figure}

\begin{figure}[ht]
    \centering
    \includegraphics[width=0.49\linewidth]{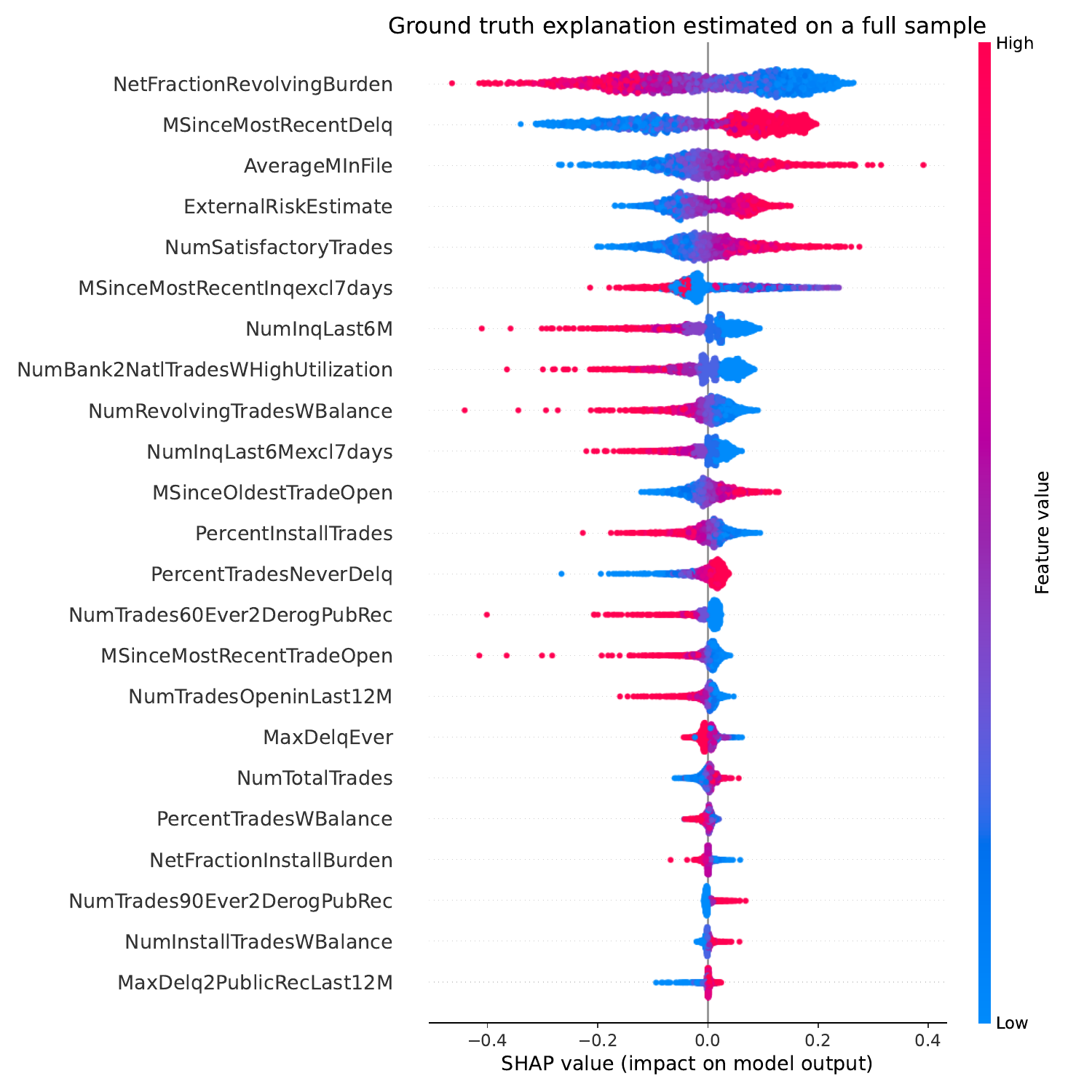}
    \includegraphics[width=0.49\linewidth]{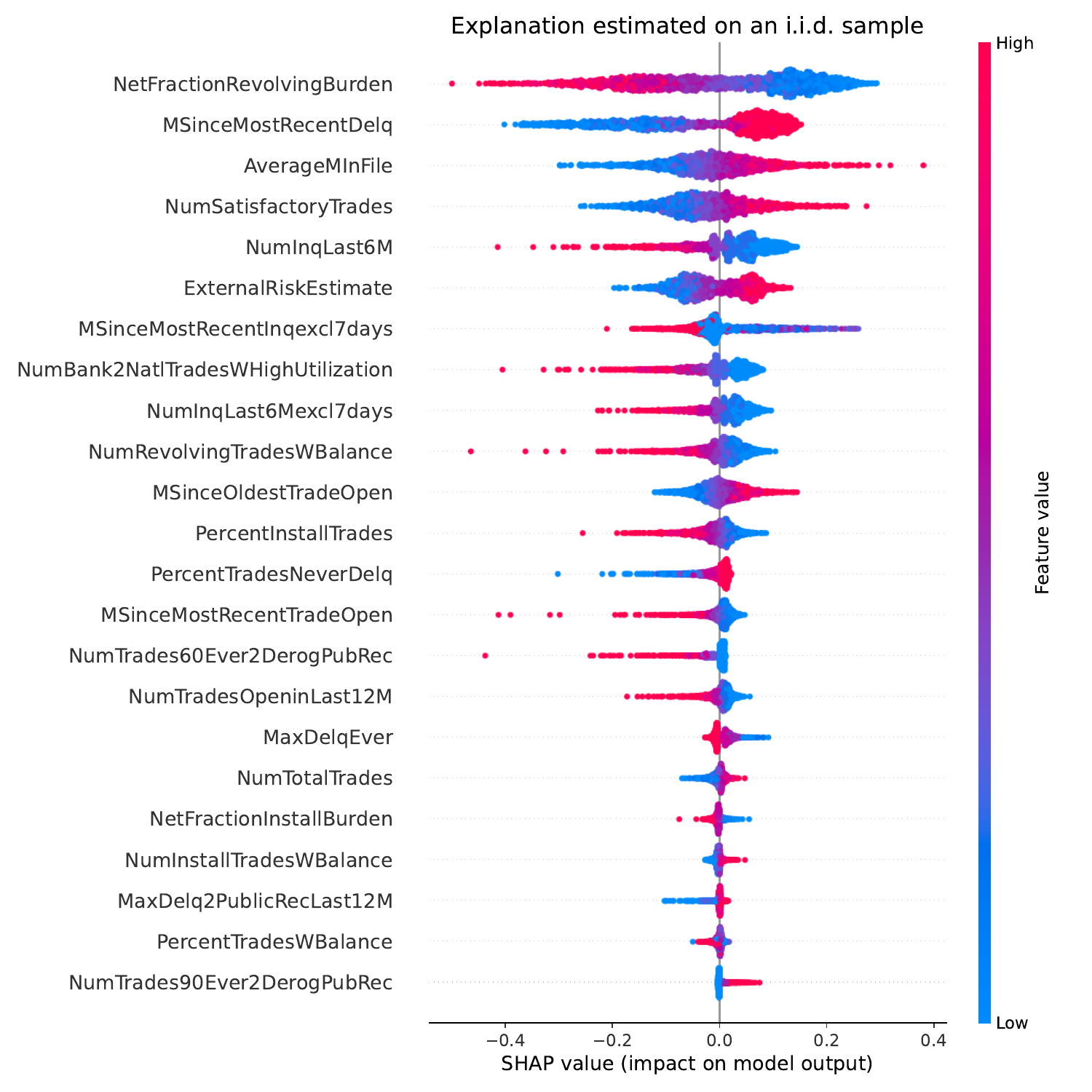}
    \vspace{-0.5em}
    
    \rule{13cm}{0.5pt}  
    
    \vspace{0.5em}   
    \includegraphics[width=0.49\linewidth]{figures/example_shap_heloc_gt.pdf}
    \includegraphics[width=0.49\linewidth]{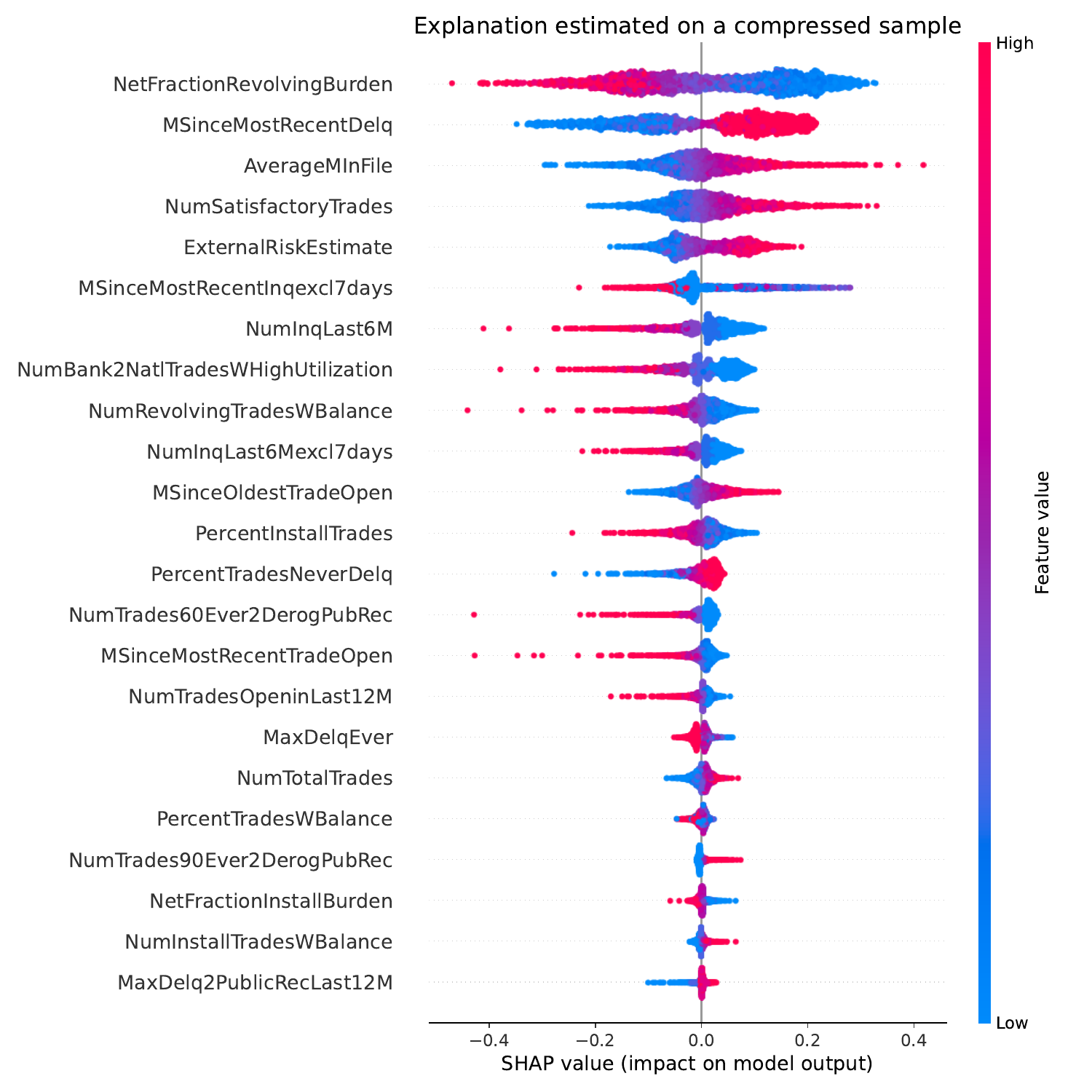}
        \caption{
        Comparison between all local \permshap explanations estimated on full (left), sampled (right top), and compressed (right bottom) subsets of the \texttt{heloc} dataset. 
        MAE introduced by \iid sampling equals $0.0087$, by \cte is $0.0053$, and so the relative improvement of \cte is~38\%.
        }
    \label{fig:example_shap_heloc}
\end{figure}

\begin{figure}[ht]
    \centering
    \includegraphics[width=0.9\linewidth]{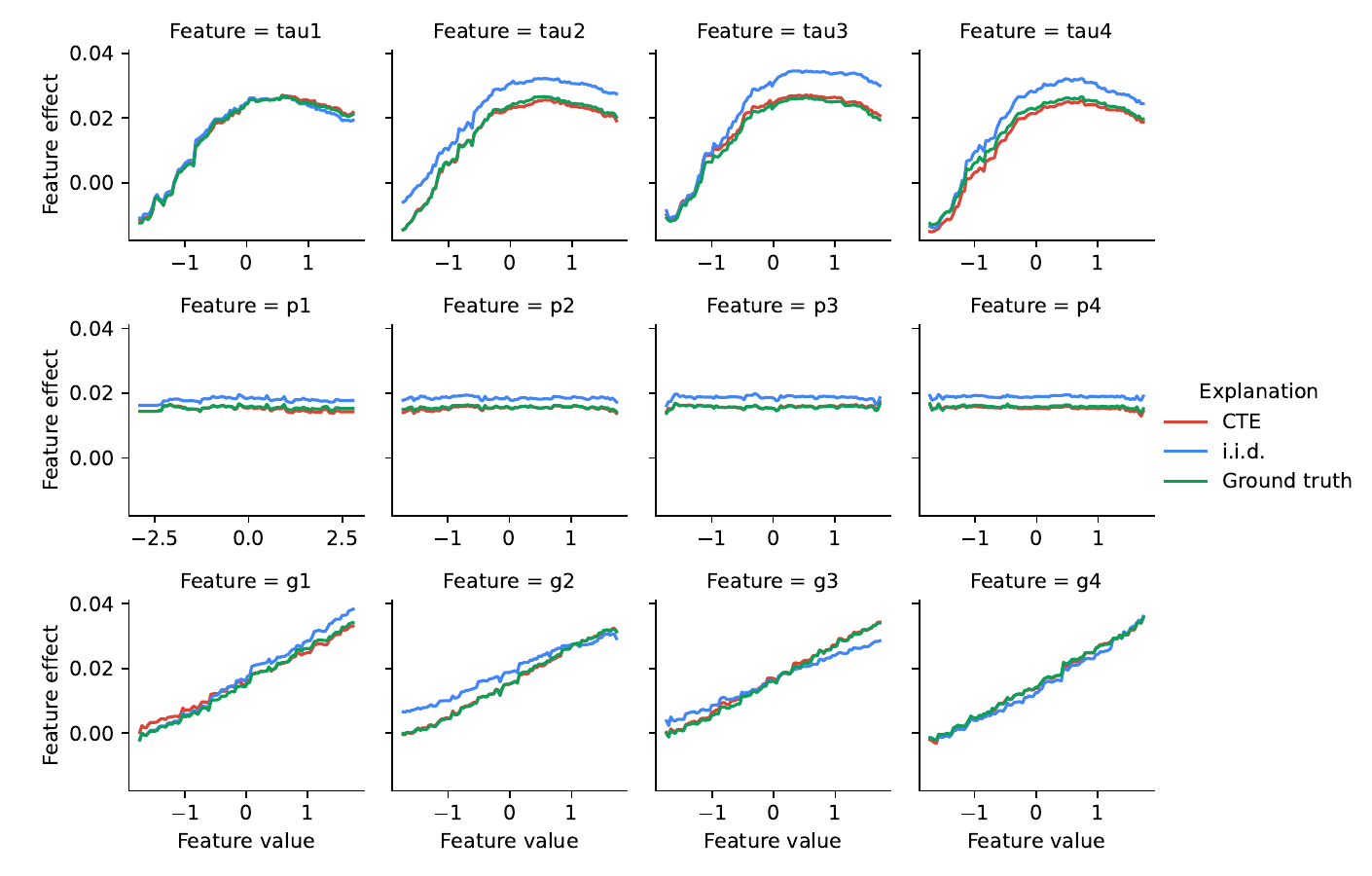}
        \caption{
        Comparison between \featureeffects explanation estimated on the full (Ground truth), sampled (\iid), and compressed (\cte) subsets of the \texttt{grid\_stability} dataset. 
        MAE introduced by \iid sampling equals $0.0032$, by \cte is $0.0007$, and so the relative improvement of \cte is~79\%.
        }
    \label{fig:example_featureffects_grid_stability}
    \vspace{1em}
    \centering
    \includegraphics[width=0.9\linewidth]{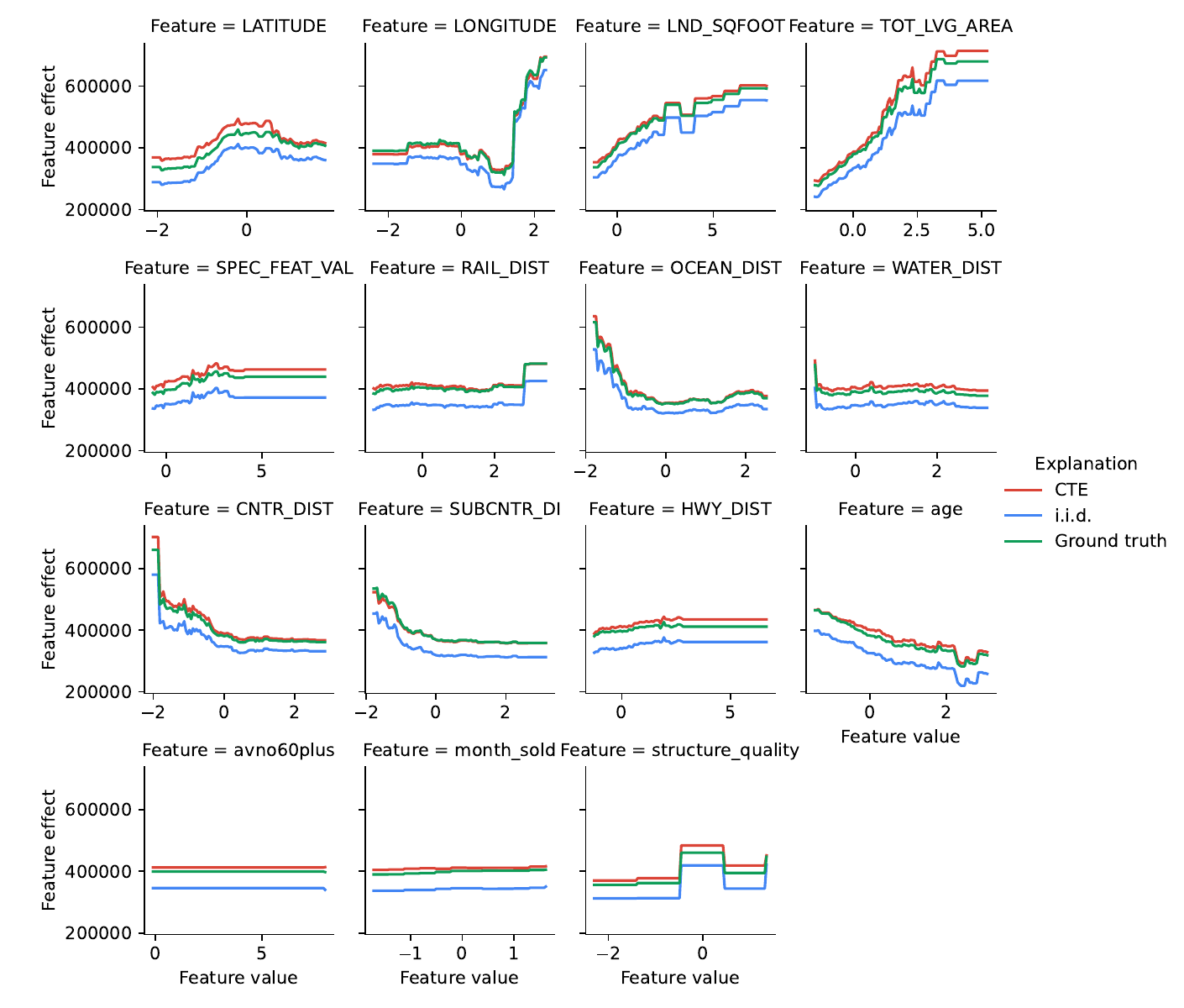}
        \caption{
        Comparison between \featureeffects explanation estimated on the full (Ground truth), sampled (\iid), and compressed (\cte) subsets of the \texttt{miami\_housing} dataset. 
        MAE introduced by \iid sampling equals $49766$, by \cte is $14031$, and so the relative improvement of \cte is~71\%.
        }
    \label{fig:example_featureffects_miami_housing}
\end{figure}

\begin{figure}
    \centering
    \includegraphics[width=0.49\linewidth]{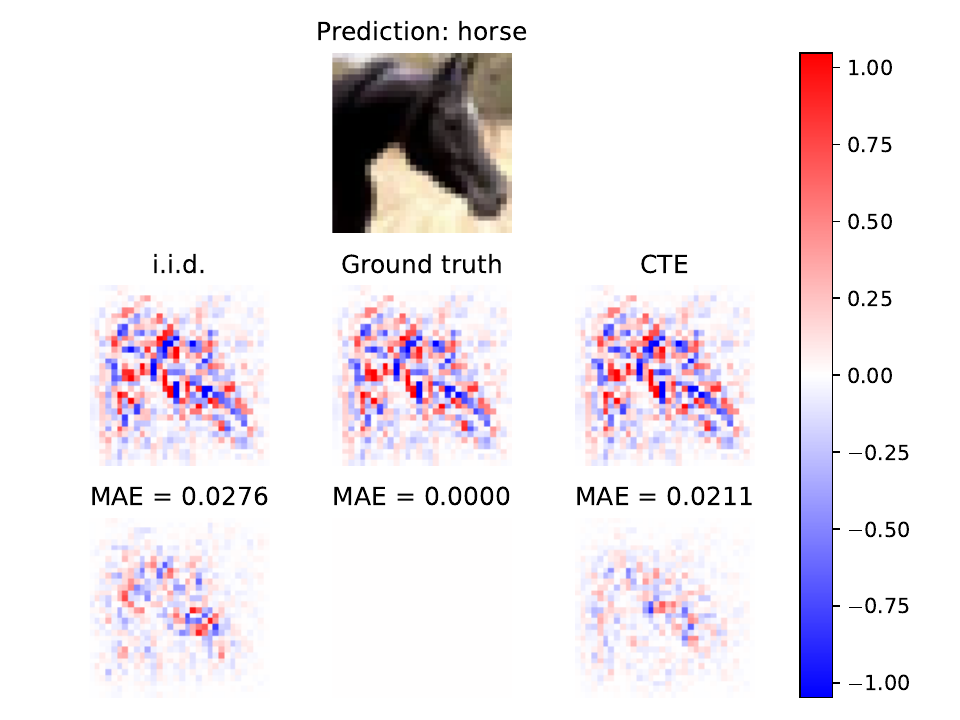}
    \includegraphics[width=0.49\linewidth]{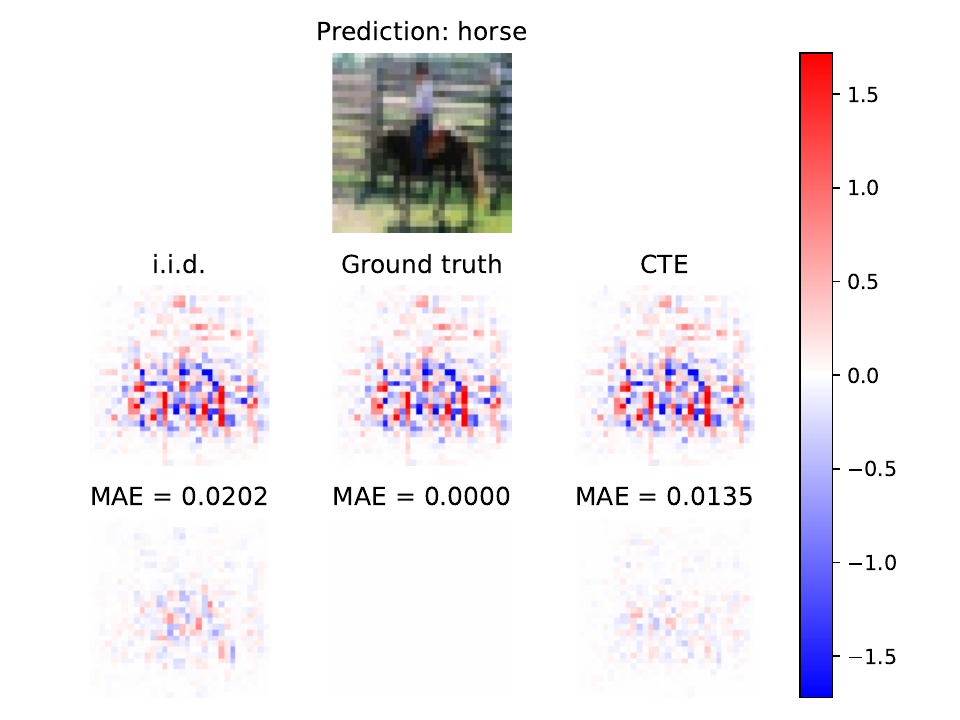}
        \includegraphics[width=0.49\linewidth]{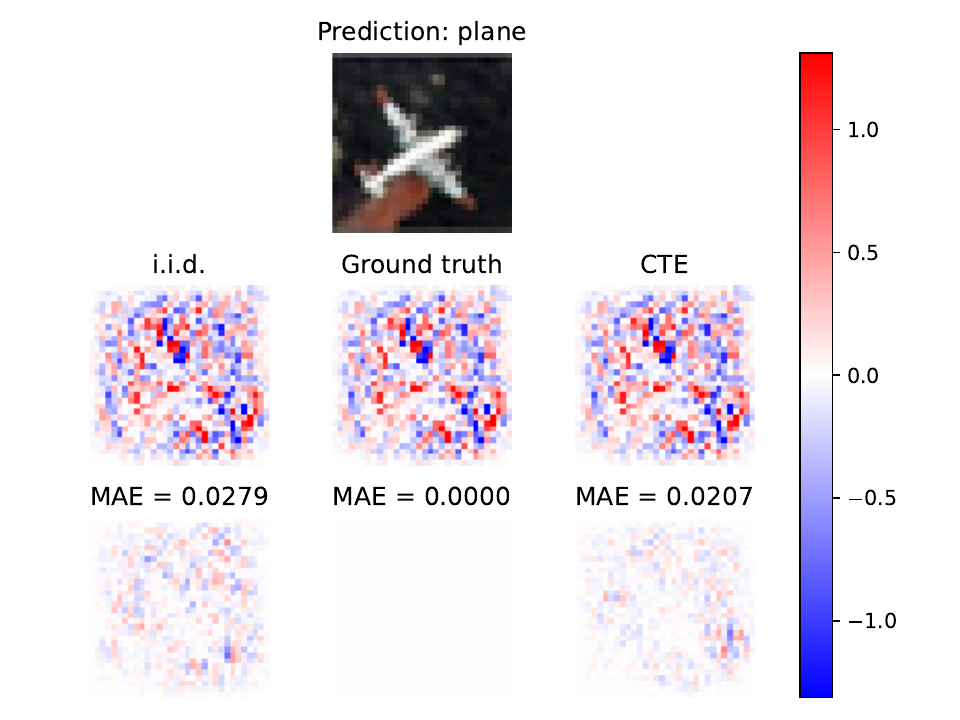}
    \includegraphics[width=0.49\linewidth]{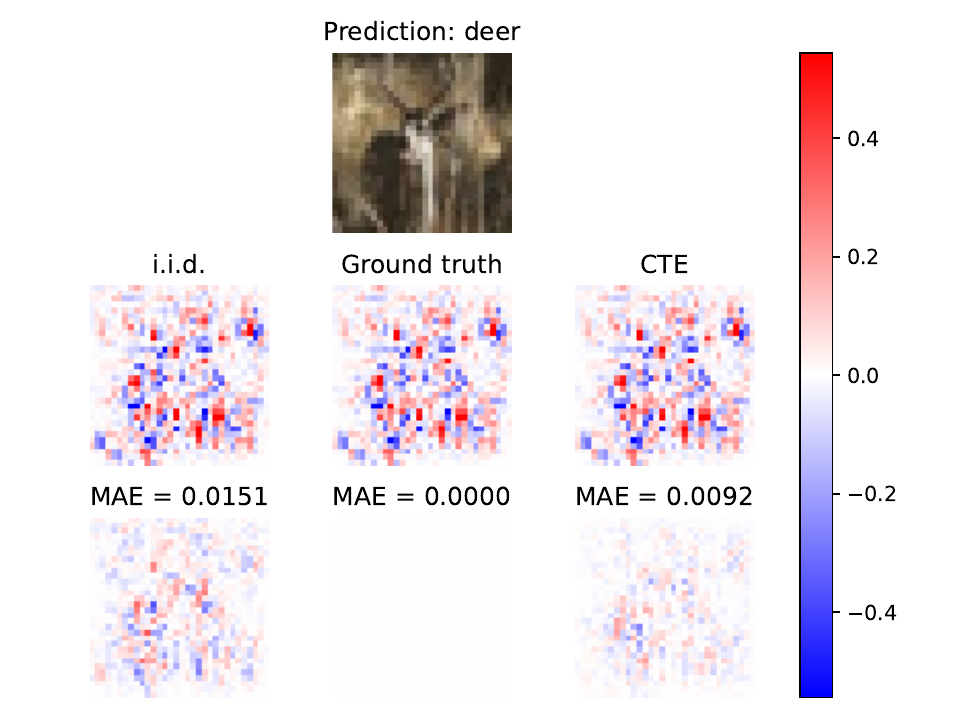}
        \includegraphics[width=0.49\linewidth]{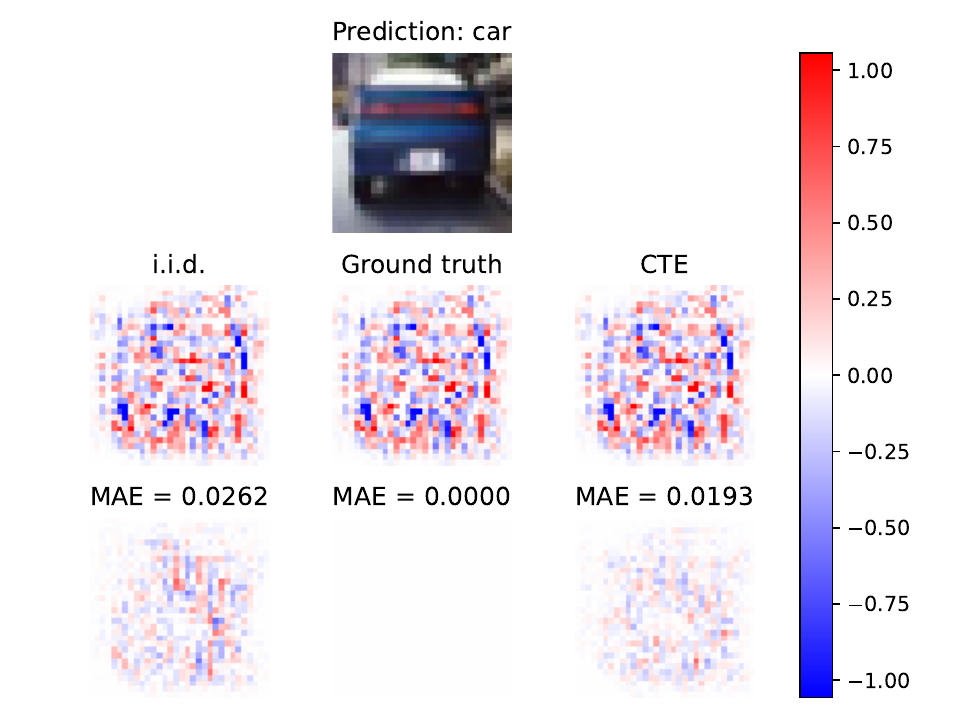}
    \includegraphics[width=0.49\linewidth]{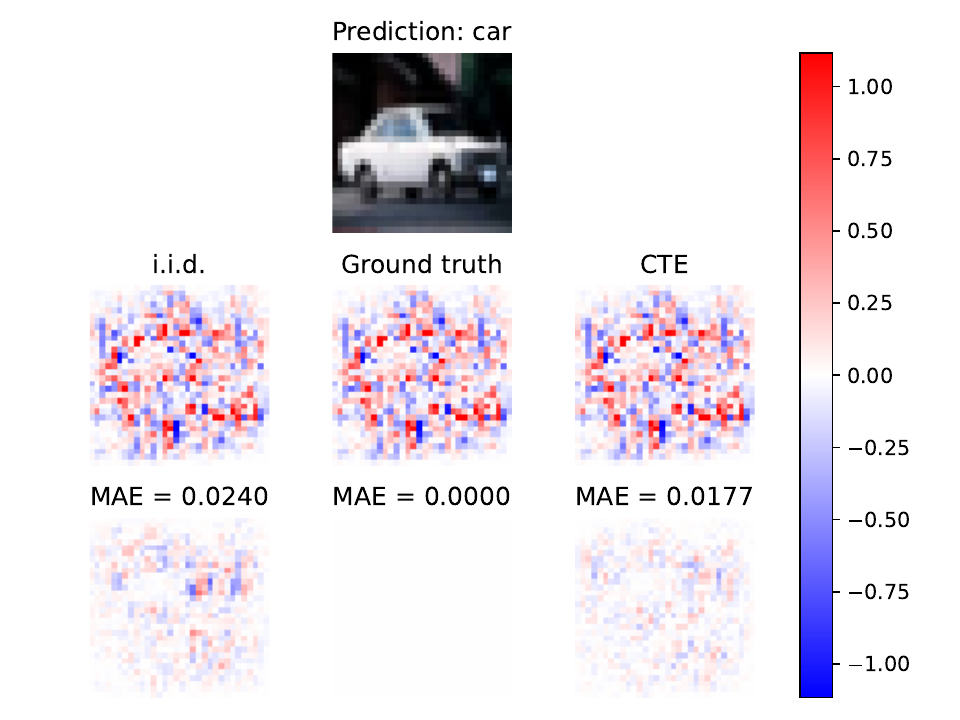}
    \includegraphics[width=0.49\linewidth]{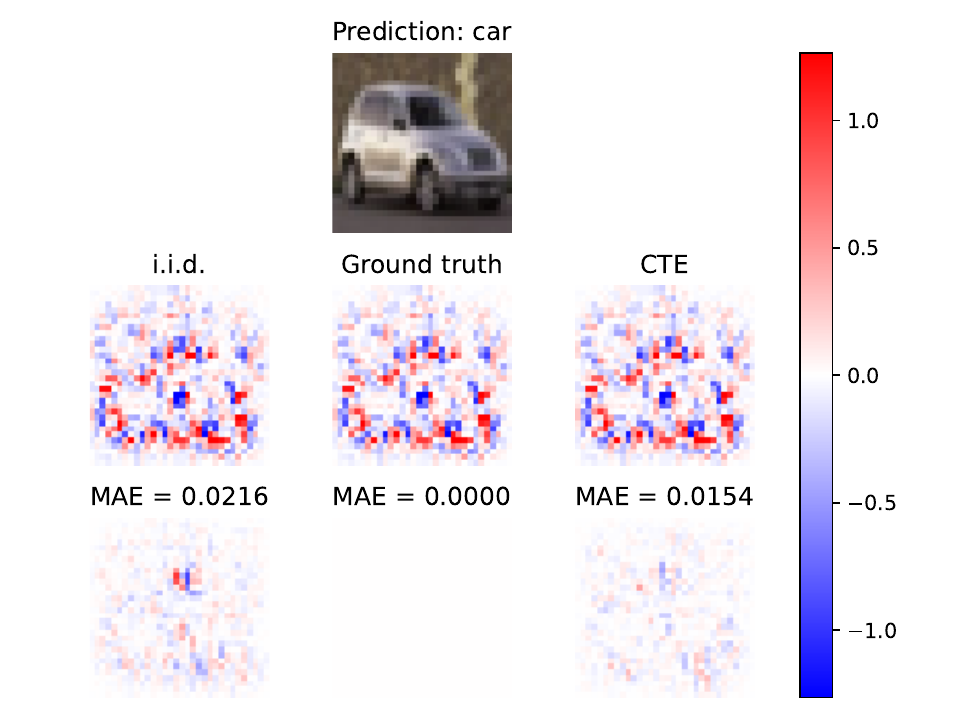}
    \includegraphics[width=0.49\linewidth]{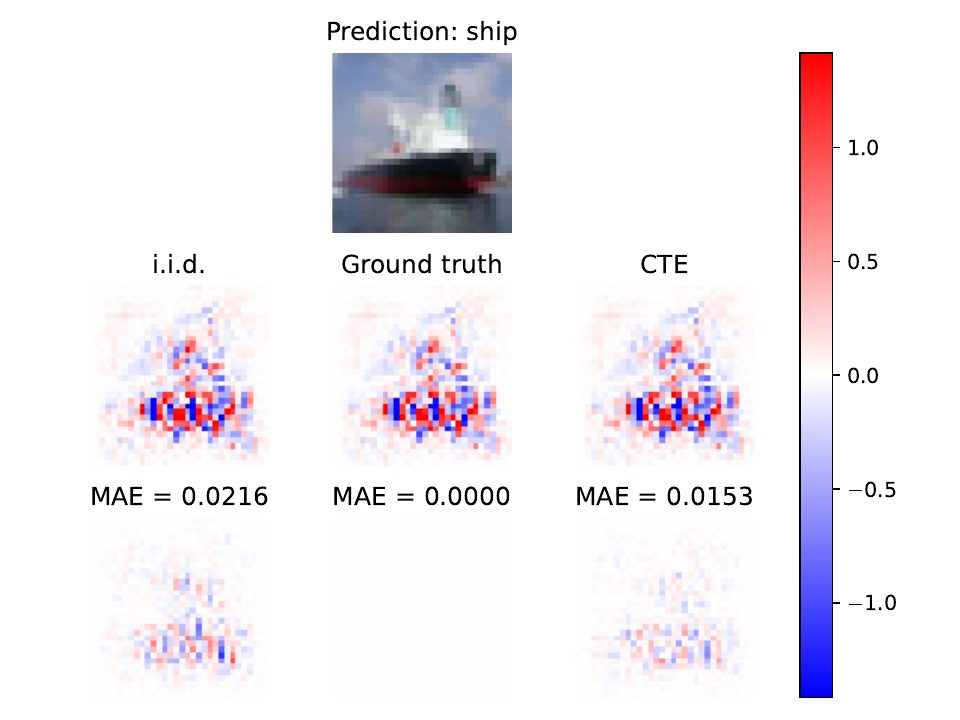}
    \caption{
    Comparison between \expectedgradients explanations estimated on the full (Ground truth), sampled (\iid), and compressed (\cte) subsets of the \texttt{CIFAR\_10} dataset. The bottom rows visualize the differences (MAE $\downarrow$) from the ground truth explanation. All predictions are correct.
    }
    \label{fig:example_expectedgradients_cifar10}
\end{figure}

\begin{figure}
    \centering
 \includegraphics[width=0.49\linewidth]{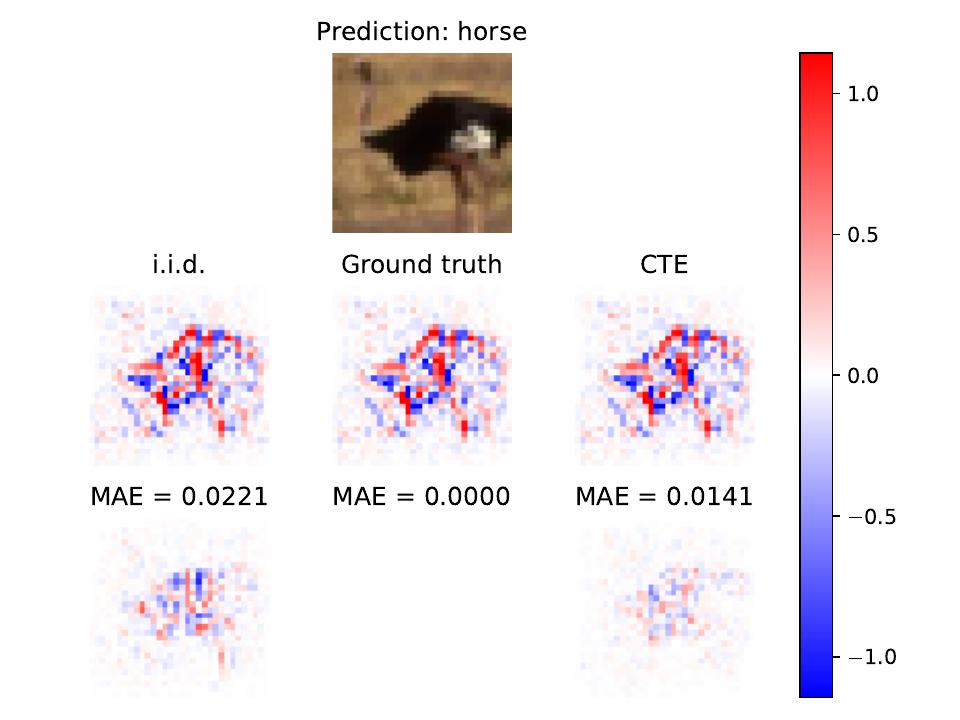}
    \includegraphics[width=0.49\linewidth]{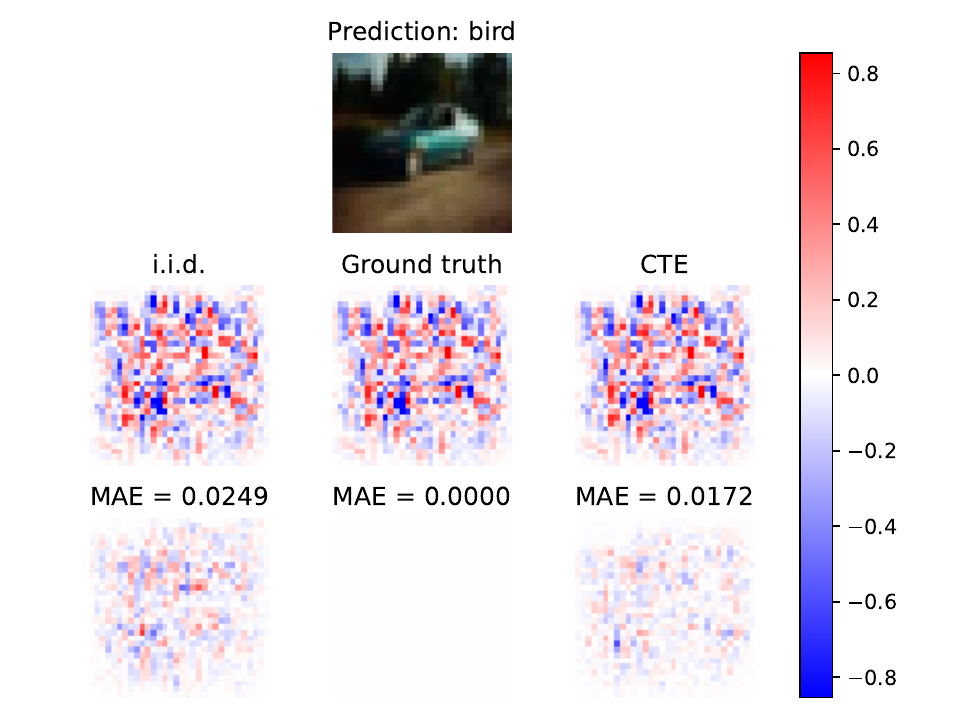}
    \includegraphics[width=0.49\linewidth]{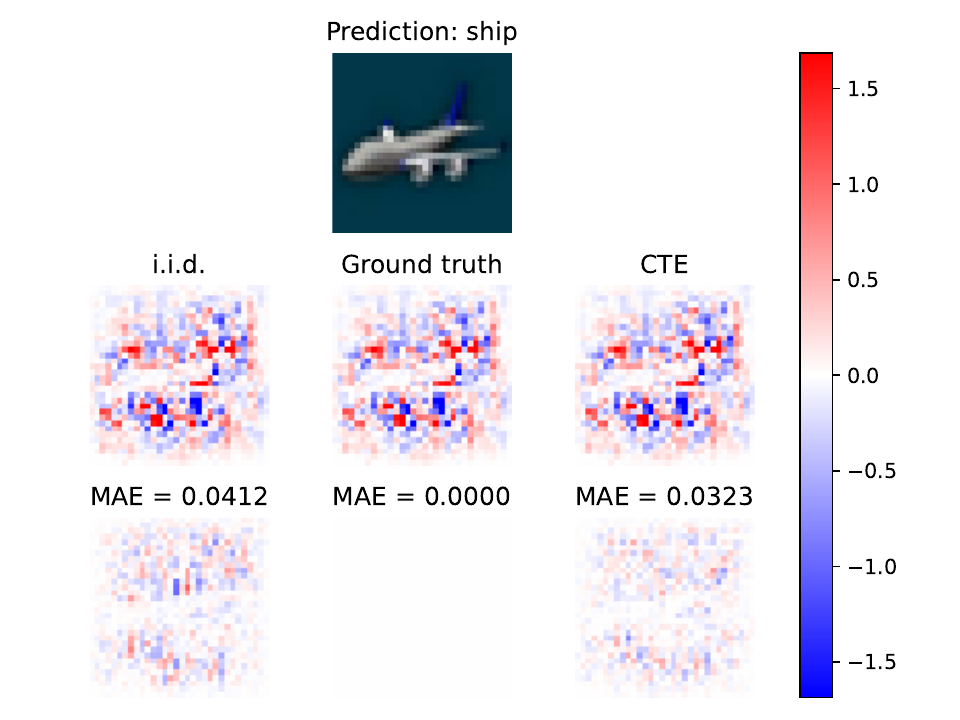}
    \includegraphics[width=0.49\linewidth]{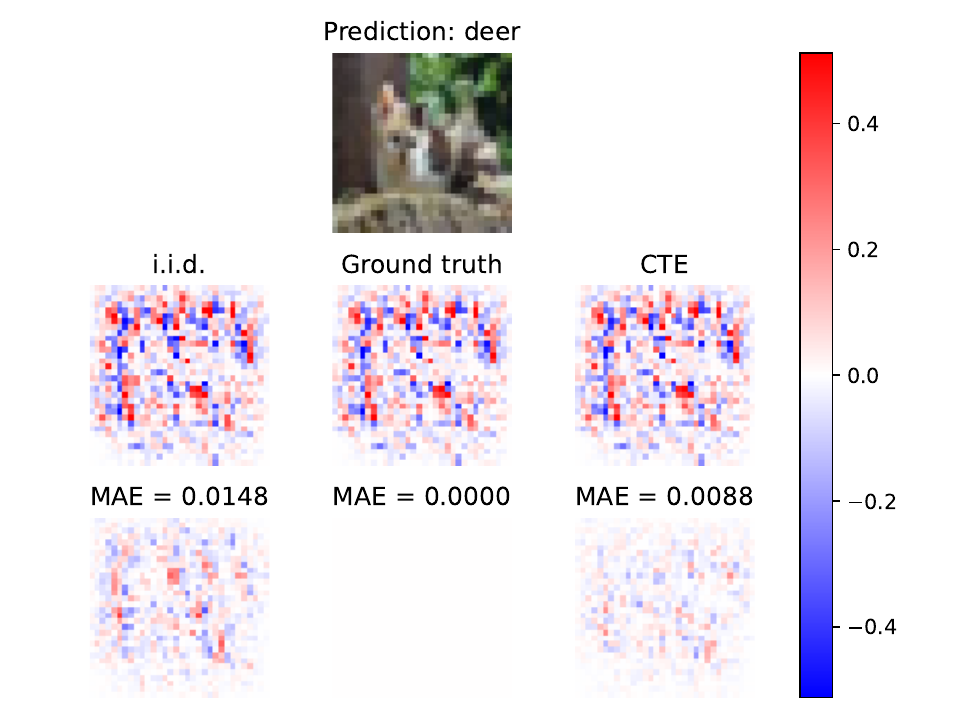}
    \caption{
    Comparison between \expectedgradients explanations estimated on the full (Ground truth), sampled (\iid), and compressed (\cte) subsets of the \texttt{CIFAR\_10} dataset. The bottom rows visualize the differences (MAE $\downarrow$) from the ground truth explanation. All predictions are wrong.
    }
    \label{fig:example_expectedgradients_cifar10_wrong}
\end{figure}

Figure~\ref{fig:example_sage_compas} shows a comparison for \kernsage explaining an XGBoost model trained on the \texttt{compas} dataset. Figure~\ref{fig:example_shap_compas} shows a comparison for \permshap explaining a neural network trained on the \texttt{compas} dataset, and Figure~\ref{fig:example_shap_heloc} shows the same on the \texttt{heloc} dataset. Note that we show all local explanations at once (not to hand-pick a single one), which might falsely look like a good approximation ``on average'' when in fact the attribution values in specific cases differ significantly. Figure~\ref{fig:example_featureffects_grid_stability} shows a comparison for \featureeffects explaining an XGBoost model trained on the \texttt{grid\_stability} dataset, and Figure~\ref{fig:example_featureffects_miami_housing} shows the same on the \texttt{miami\_housing} dataset. Figures~\ref{fig:example_expectedgradients_cifar10}~\&~\ref{fig:example_expectedgradients_cifar10_wrong} show exemplary visual comparisons for \expectedgradients explaining a convolutional neural network trained on the \texttt{CIFAR\_10} dataset.

\end{document}